\documentclass[preprint,12pt]{elsarticle}

\usepackage{amssymb}
\usepackage{amsfonts}
\usepackage{amsthm}
\usepackage{graphics} 
\usepackage{epsfig} 
\usepackage{amsmath} 
\usepackage{subfigure}
\usepackage{multirow}
\usepackage{makecell}
\usepackage{graphicx}
\usepackage{lscape}
\usepackage{multicol} 
\usepackage{graphicx, epstopdf}
\usepackage[table,xcdraw]{xcolor}
\usepackage{caption}
\usepackage{xspace,epsfig,url}
\usepackage{enumerate}
\usepackage{float}
\usepackage{epsf}
\usepackage{psfrag}
\usepackage{verbatim}
\usepackage[all]{xy}
\usepackage{color,soul}
\usepackage{bm}
\usepackage{comment}
\usepackage{cases}
\usepackage{hyperref}
\usepackage{physics}
\usepackage{amssymb}
\usepackage{pifont}
\usepackage{array} 
\usepackage[ruled,vlined,linesnumbered]{algorithm2e}

\DeclareMathOperator{\sign}{sign}

\journal{Engineering Applications of Artificial Intelligence}

\begin{document}

\begin{frontmatter}



\title{Design Optimizer for Planar \\Soft-Growing Robot Manipulators}


\author[inst1]{Fabio Stroppa}

\affiliation[inst1]{organization={Computer Engineering Department, Kadir Has University},
            addressline={Cibali, Kadir Has Cd., Fatih}, 
            city={Istanbul},
            postcode={34083}, 
            country={Turkey}}

\begin{abstract}
Soft-growing robots are innovative devices that feature plant-inspired growth to navigate environments. Thanks to their embodied intelligence of adapting to their surroundings and the latest innovations in actuation and manufacturing, it is possible to employ them for specific manipulation tasks. The applications of these devices include exploration of delicate/dangerous environments, manipulation of items, or assistance in domestic environments.

This work presents a novel approach for design optimization of soft-growing robots, which will be used prior to manufacturing to suggest to engineers -- or robot designer enthusiasts -- the optimal size of the robot to be built for solving a specific task.
The design process is modeled as a multi-objective optimization problem, optimizing the kinematic chain of a soft manipulator to reach targets and avoid unnecessary overuse of material and resources. 
The method exploits the advantages of population-based optimization algorithms, in particular evolutionary algorithms, to transform the problem from multi-objective into single-objective thanks to an efficient mathematical formulation, the novel rank-partitioning algorithm, and obstacle avoidance integrated within the optimizer operators.
The proposed method was tested on different tasks to assess its optimality, which showed significant performance in solving the problem: the retrieved designs are short, smooth, and precise at reaching targets. Finally, comparative experiments show that the proposed method works better than the one existing in the literature in terms of precision ($14\%$ higher), resource consumption ($2\%$ shorter configurations with $4\%$ fewer links), actuation ($85\%$ less wavy and undulated configurations), and run time ($13\%$ faster).
\end{abstract}

\begin{graphicalabstract}
\includegraphics[width=15cm]{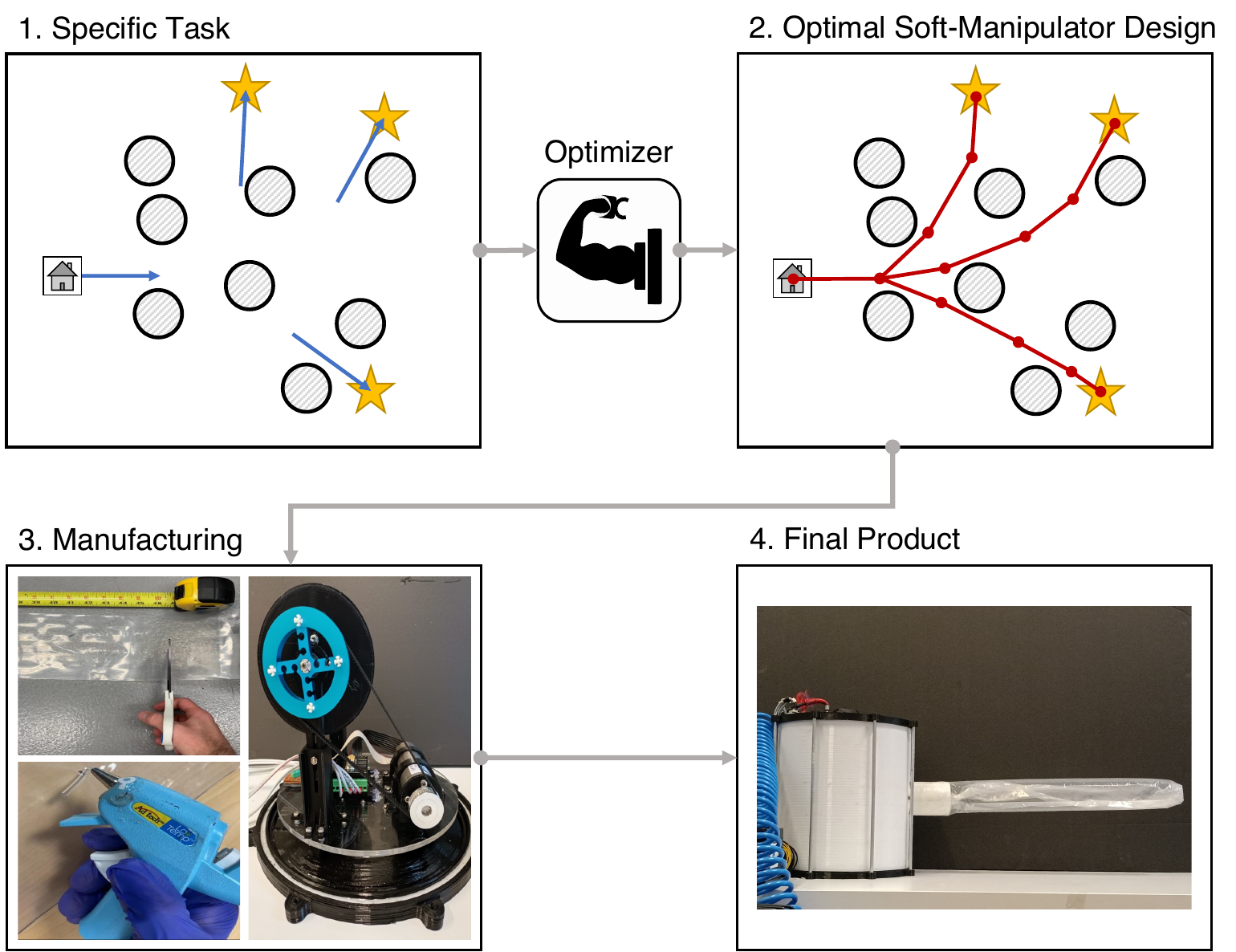}
\end{graphicalabstract}

\begin{highlights}
\item A novel and detailed mathematical formulation to define the design of soft-growing robotic manipulators as a multi-objective optimization problem.
\item A population-based algorithm, namely the \textit{Rank Partitioning} algorithm, which allows the optimizer to transform a multi-objective optimization problem into a single-objective optimization problem without the burden of choosing parametric values to define objective priorities.
\item A method for obstacle avoidance to be integrated with the genetic operators of evolutionary algorithms while generating random individuals (either by variation or initialization).
\end{highlights}


\end{frontmatter}



\section{Introduction}
\label{sec:intro}

\begin{figure}[b!]
    \centering
    \subfigure[\protect\url{}\label{fig:soft_robots_1}]%
    {\includegraphics[height=5cm]{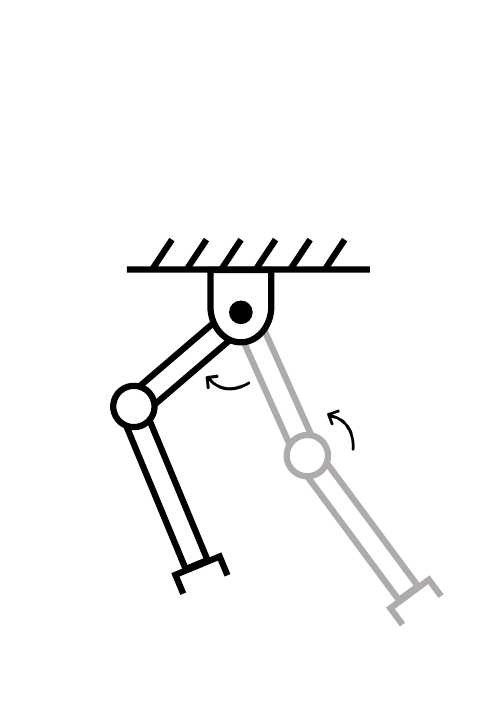}}
    \subfigure[\protect\url{}\label{fig:soft_robots_2}]%
    {\includegraphics[height=5cm]{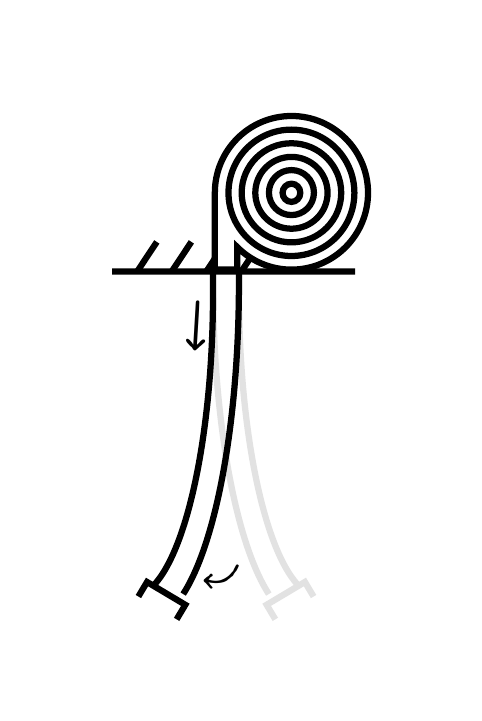}}
    \subfigure[\protect\url{}\label{fig:soft_robots_3}]%
    {\includegraphics[height=5cm]{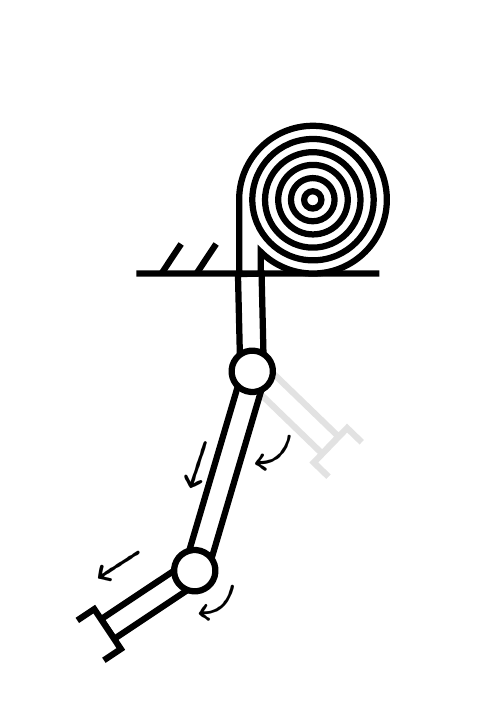}}
    \caption{Different types of robotic manipulators: (a) a traditional rigid manipulator with localized joints, large body workspace, and significant inertia; (b) a traditional soft robot that lacks dexterity and cannot support significant payloads; and (c) a soft growing manipulator with continuum links, variable stiffness, and variable discrete joints~\cite{do2020dynamically}.}
    \label{fig:soft_robots}
\end{figure}

Robots have gone through a dynamic and interesting evolution process with different contributions and aspects. Traditional rigid-industrial robots, such as serial-chain manipulators (Fig.~\ref{fig:soft_robots_1}), can be integrated with control, artificial intelligence, and perception to work alongside humans. However, they also present several aspects that may cause concerns: (i) systems with hard links might have problems interacting with fragile objects or sensitive environments -- e.g., human tissues, glassware, and delicate elements within archaeological sites; (ii) their complexity and material make cost usually high; and (iii) as powerful rigid machines, their safety for human operators is questionable. As a solution, a set of diverse and innovative robot designs has been proposed: soft-growing robots (Fig.~\ref{fig:soft_robots_2}). Inspired by living organisms, the main purpose of these designs is to exploit the inherent compliance of their materials to enable naturalistic physical interactions between them and the environment~\cite{webster2010design, rus2015design, whitesides2018soft, hawkes2021hard}. Indeed, their materials and actuation technologies are less expensive, safer, and easier to manufacture or assemble than traditional rigid robots. 
Furthermore, soft robots own their dexterity to the ability to grow or retract, a process called \textit{eversion}; this degree of freedom (DoF) is directly inspired by growing plants, and these robots are also often called \textit{vine} robots~\cite{coad2019vine, hawkes2017soft, blumenschein2017modeling}.
Studies have shown that they can be used in manipulation tasks requiring payload-handling capabilities~\cite{stroppa2020human}.  
A further innovation was the introduction of continuum links and discrete joints, which can be achieved by including layer jamming to dynamically change the stiffness of the robot where joints are located, as shown in Fig.~\ref{fig:soft_robots_3}~\cite{do2020dynamically}.

Thanks to their low cost, soft robots are accessible to the general public (i.e., non-professional robot designers), and anyone could successfully create their own soft-growing robot to solve specific tasks~\cite{schulz2017interactive, morimoto2018toward, exarchos2022task}. 
However, the process of specifying and combining these components to create a task-specific robot is too difficult for human designers: soft-growing manipulators usually do not feature classical joints (Fig.~\ref{fig:soft_robots_2}); and even those who do have joints still feature eversion (Fig.~\ref{fig:soft_robots_2}), allowing their end effector (i.e., the tip of the robot) to be anywhere on their kinematic chain. 
Therefore, their kinematic formulation and actuation become a considerable challenge. Ideally, a soft-robot designer would only need to provide functional goals and constraints of a specific task, and an optimal design would automatically be generated~\cite{hiller2011automatic}.

Retrieving the design of a soft-growing robot is a non-trivial problem that can be solved through mathematical optimization. A \textit{design} is the set of parameters and characteristics that describe a soft-growing robot and provide instructions for manufacturing: these can include link lengths, joint location and range of motion (if any), material, and shape or type. 
Different optimization methods have been used in literature for soft robot design: gradient-based methods~\cite{marchese2016dynamics, wang2020topology, cheong2021optimal, chen2021enhancing, ghoreishi2021bayesian, bern2021soft, tang2021design}, greedy algorithms~\cite{koehler2020model}, geometric computation~\cite{ma2017computational, fang2020kinematics}, Bayesian optimization~\cite{ghoreishi2021bayesian}, linear programming~\cite{camarillo2008mechanics, rucker2011statics, burgner2013computational, bergeles2015concentric}, non-linear programming~\cite{lloyd2020optimal, doroudchi2021configuration, rosi2022sensing}, quadratic programming~\cite{coevoet2017optimization, adagolodjo2021coupling}, or other informed deterministic algorithms~\cite{anor2011algorithms, de2020topology}.
Evolutionary computation (EC) -- a sub-field of artificial intelligence -- is widely used to optimize the design and kinematics of mechanical robots~\cite{coello1998using,iwasaki2000evolutionary,lim2005inverse,clark2013evolutionary,gao2014performance,wang2015multi,filipiak2015infeasibility,rokbani2022beta, stroppa2023optimizing}, and due to its population-based methodology, it is the best candidate to solve multi-optimization problems (e.g., maximize workspace while minimizing production cost)~\cite{deb1999multi}. 
EC techniques are used in several studies on soft robotics to optimize topology~\cite{hiller2011automatic, bedell2011design, berger2015growing, cheney2015evolving, chen2019optimal, medvet2021biodiversity}, flexibility~\cite{luo2015optimized,chen2019optimal}, dexterity~\cite{bodily2017multi, bodily2017multi,tan2017simultaneous,liu2022simulation,fitzgerald2021evolving}, load capacity~\cite{bodily2017multi}, locomotion~\cite{rieffel2009evolving,tesch2013expensive,methenitis2015novelty,kriegman2017minimal, medvet2021biodiversity, sui2022task, marzougui2022comparative}, actuation~\cite{runge2017design}, stiffness~\cite{chen2019optimal}, morphology~\cite{methenitis2015novelty}, sensor placement~\cite{ferigo2022optimizing}, and control~\cite{nguyen2007genetic, medvet2021biodiversity, pigozzi2022evolving,ghoul2022optimized}.
In these studies, optimizing the robot's design is to identify the least complex configuration (e.g., with a minimal number of sections) that can perform a given task or set of tasks~\cite{anor2011algorithms}. Optimizing the design often involves solving the kinematics of the robot, which for a soft manipulator translates into also optimizing its joint bending angles or curvature at each moment~\cite{chen2020obstacle}.
Furthermore, EC techniques are commonly used for path planning, when robots -- not necessarily soft ones -- have to explore environments with obstacles~\cite{dakev1996evolutionary, castillo2007multiple, chae2009trajectory, ayari2017new,chen2020obstacle}.

In this study, we focus on design optimization for planar soft-growing robots based on the technology introduced by Do et al. for inflatable beams with distributed and reconfigurable stiffness~\cite{do2020dynamically}. The problem we tackle is retrieving the set of link sizes allowing a single robot to reach all targets in a specific environment with obstacles. This involves minimizing the distance to the target(s), aligning the robot with the desired reaching orientation, minimizing the number of links and their size, minimizing obstacle collisions, and so on. To account for all these parameters -- or objectives -- an optimizer should perform multi-objective optimization and retrieve a set of different solutions that favor a specific parameter over another~\cite{deb1999multi}. Since these objectives often conflict with each other (e.g., we cannot find a robot that reaches a distant target without sacrificing the length of its links), some of the solutions retrieved by an optimizer might not be interesting for our purposes (e.g., a very short robot that does not reach the target at all). Therefore, rather than a set of trade-off solutions, we aim at a single solution with different priorities for each objective (e.g., giving more relevance in reaching the target and only then reducing the length).
A first solution to this problem was proposed in our previous work~\cite{exarchos2022task}, in which the optimization method was based on a simple weighted sum~\cite{miettinen2012nonlinear, hwang2012multiple}, and the obstacles were defined only as constraints of the problem. To the best of our knowledge, this is the only study addressing the problem of designing a soft-growing robot for a specific manipulation task. However, the solutions retrieved by this optimizer are wavy, not oriented toward the targets, and not optimized in terms of robot length. Additionally, weighted-sum methods come with the disadvantage of user-defined preference values for each objective without the guarantee of fully exploring non-convex objective spaces.

The method we currently propose exploits the advantages of population-based optimizers such as EC techniques to transform the problem from multi to single-objective with a novel method called \textit{rank partitioning}. This procedure sorts solutions based on the objectives, removing the burden of choosing parametric values to define priorities -- i.e., each objective is ranked based on its priority, such that designers are required only to choose the order of priorities. We present the rank partitioning within a genetic algorithm~\cite{goldberg1990real}, but it can be employed in any population-based optimizer. Additionally, we integrated an obstacle avoidance method within the operators of the optimizer to reduce the size of the search space. 
Note that we developed the code in MATLAB; therefore, every mathematical notation (e.g., genotypes) will have a matrix form.
The contributions of this work can be summarized as follows: (i) a novel mathematical formulation for designing soft-growing robotics manipulator in specific tasks; (ii) a novel methodology (rank partitioning) to solve multi-objective optimization problems when trade-off solutions are limiting the quality of the retrieved solutions; (iii) an alternative to weighted-sum methods that does not require users to define numerical priorities for each objective; (iv) a novel obstacle avoidance algorithm based on geometric computation; and (v) a tool for soft-growing robot design that outperforms the previous work in the state-of-the-art in terms of usability, optimality, and run time.

The rest of the work is divided as follows: Sec.~\ref{sec:background} will provide a brief background on soft-growing robots; Sec.~\ref{sec:optimization_problem} will provide an in-depth description of the optimization problem and a full walk-through of the proposed solution based on EC techniques; Sec.~\ref{sec:experiments} will present the results of the method applied to the design problem; and finally, Sec.~\ref{sec:conclusion} will conclude the work with a summary and a list of on-going future works.

\section{Background on Soft-Growing Robot Manipulators}
\label{sec:background}

Soft-growing robots, particularly vine robots, take inspiration from plants to exploit their ability to grow to navigate environments~\cite{tsukagoshi2011tip, sadeghi2017toward, hawkes2017soft}. 
Soft-growing robots have been used to interact with devices such as valves and/or deliver tools in search and rescue scenarios~\cite{hawkes2017soft, jeong2020tip}, deploy structures such as antennas~\cite{blumenschein2018tip}, and navigate complex environments such as archaeological sites~\cite{coad2019vine}. 
These applications show that eversion is an innovative and advantageous DoF, especially for manipulation and navigation in cluttered, dangerous, or delicate environments \cite{hawkes2017soft, kim2018origami, coad2019vine}. 

The type of robots we are considering in this work have the following properties: (i) they can grow and retract, and (ii) they have discrete joints located in specific points of their body that allow them to steer in the environment (Fig.~\ref{fig:soft_robots_3}). The first property -- growth and retraction (eversion) -- is enabled by the characteristics of their body, which consists of an inflated beam made of thin-walled and complaint material that pneumatically everts: initially completely folded, the application of pressurized air causes the robot to unfold at the tip, drawing new material through itself~\cite{hawkes2017soft, mishima2003development,tsukagoshi2011tip}. 
The second property -- bending at specific locations -- is performed by layer jamming to control the stiffness of different sections of the robot's body, enabling cable-driven steering~\cite{blumenschein2018tip,blumenschein2018helical,gan20203d}. The principle and technology behind these robots are mainly described in the work of Do et al.~\cite{do2020dynamically}.

\begin{figure}[b!]
	\centering
        \subfigure[\protect\url{}\label{fig:maze}]%
	{\includegraphics[height=4cm]{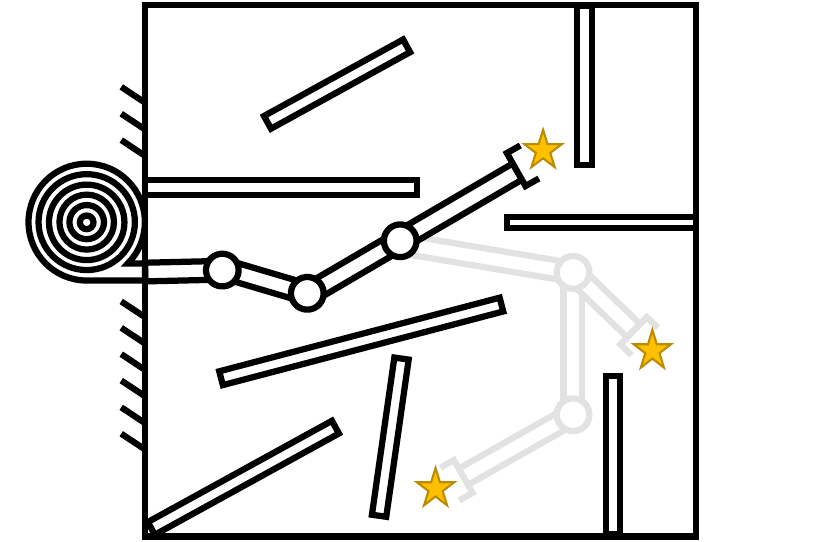}}
        \subfigure[\protect\url{}\label{fig:inflated_robot}]%
        {\includegraphics[height=4cm]{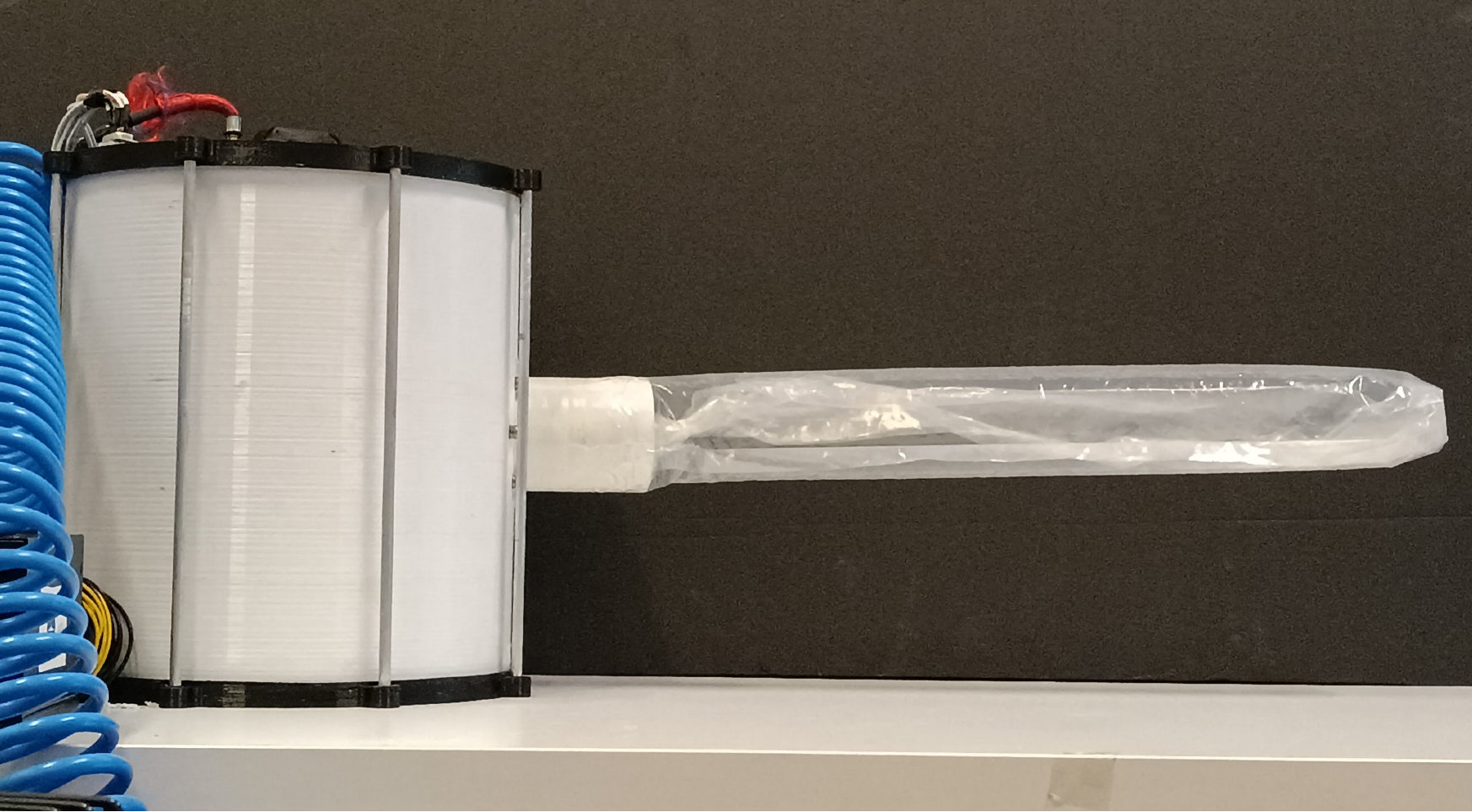}}
	\caption{(a) Sketch of a planar soft-growing robot reaching targets in a maze (top view). The configurations of this solution share the same angle for some of the joints, as the robot progressively grows and retracts to reach the next target. (b) A real prototype of a soft-growing robot. }
	\label{fig:example}
\end{figure}

To allow for manipulative/navigational tasks, the tip of the robot must be equipped with sensors or tools (e.g., a camera or a gripper) that should translate with the robot as it grows. This presents a challenge because the material at the robot's tip continually changes during eversion.
Furthermore, retraction also presents its challenges, especially in terms of controllability as the robot can easily oscillate abruptly while folding its body, causing undesired buckling. 
Recent studies have developed methods to overcome these two problems by proposing a magnetic-based tip mount that successfully remains attached to the robot tip while providing a mounting point for sensors and tools~\cite {jeong2020tip, coad2020retraction}.
Based on the aforementioned technology, several studies have been carried on to demonstrate the capability of soft-growing robots in manipulative tasks (e.g., pick and place tasks). During these tasks, soft-growing robots can be teleoperated with different interfaces (e.g., joystick~\cite{el2018development} and body gesture tracking~\cite{stroppa2020human}).

In the scenario analyzed in this work, we consider soft-growing robots used to reach predefined targets in a complex environment, as schematized in Fig.~\ref{fig:maze}. The environment can be modeled as a maze, with a set of obstacles that must be avoided, and a set of targets that must be reached with a desired orientation. The dexterity of the aforementioned soft robots allows them to explore the maze and manipulate the targets. The problem we face is how to design a robot whose properties ensure the success of the task. Fig.~\ref{fig:inflated_robot} shows a picture of a real manipulator.

\section{Optimization problem and Proposed Solution}
\label{sec:optimization_problem}

\begin{figure}[ht]
    \centering
    \subfigure[\protect\url{}\label{fig:problem_example_a}]%
    {\includegraphics[width=6cm]{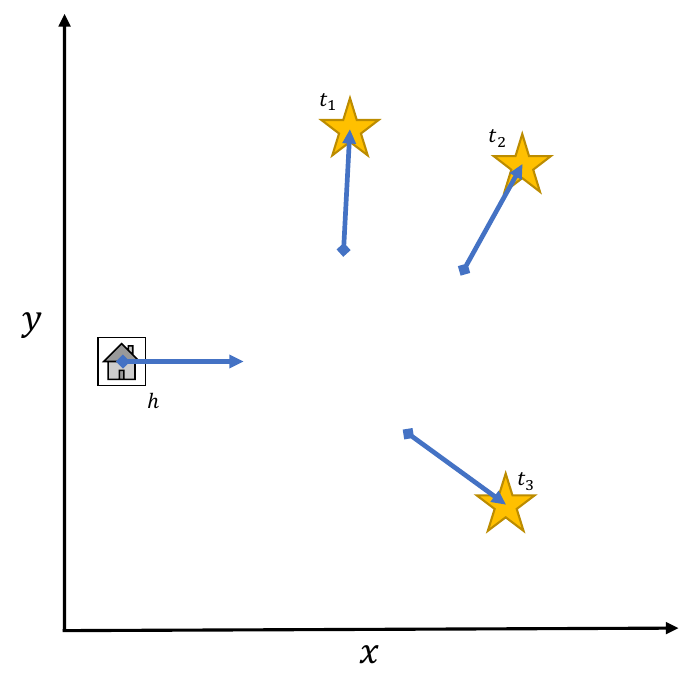}}
    \subfigure[\protect\url{}\label{fig:problem_example_b}]%
    {\includegraphics[width=6cm]{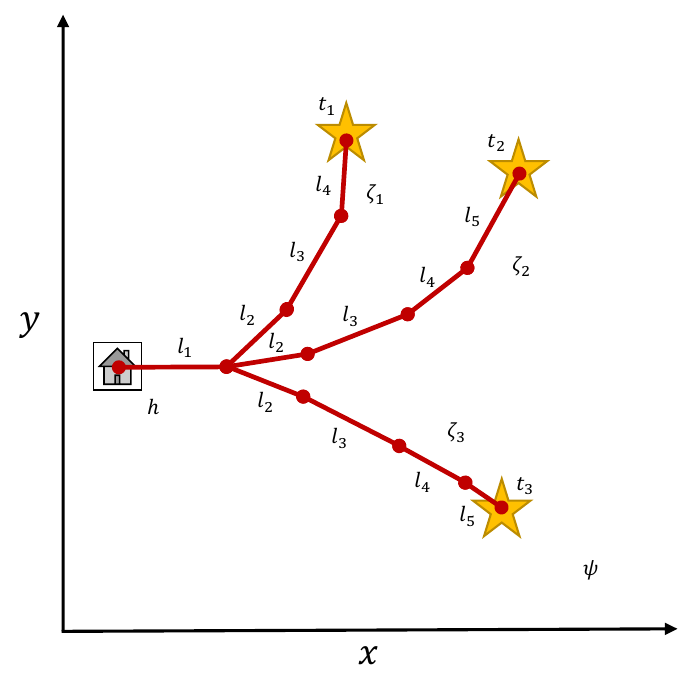}}
    \caption{(a) Example of a task to be performed by a soft-growing robot: the robot should reach the three targets with a given orientation starting from the home base. The task is defined with three targets ($t=3$) and no obstacles ($o=0$). (b) Example of solution $\psi$ provided by the optimizer. The solution $\psi$ is then composed of three configurations $\zeta_{1-3}$ sharing the same design $\delta$ (the lengths of links are shared among the three configurations). The number of joints can vary within each configuration, as the robot will employ only a sufficient number of links to reach a certain target. In this solution, the joints defining the tip of the robots are $\bar{n}_1=4$, $\bar{n}_2=5$, $\bar{n}_3=5$.}
    \label{fig:problem_example}
\end{figure}

A soft-growing robot needs to be designed to reach a set of targets in a workspace, and the design will be retrieved by our optimizer. This problem can be defined in both 2D and 3D based on how gravity affects the system: a robot exploring a tunnel of an archaeological site can be thought of as a 2D robot, but if the site requires the robot to explore different layers (e.g., a multi-story building) then the robot also needs to move in the third dimension. However, in this current work, we only investigated planar robots that act in a two-dimensional space, leaving the third dimension for future works. Fig.~\ref{fig:problem_example} shows a 2D representation of the problem, where the robot explores a plane x-y and gravity effects are neglected\footnote{In this work, gravity is oriented in the negative direction of the z-axis.}. The robot can steer in any direction and the tip does not have gravity constraints on its orientation. As shown in Fig.~\ref{fig:problem_example_a}, the robot is aiming for targets with a specified reaching orientation: this requires the robot to align its last link(s) to a specific straight line. Fig.~\ref{fig:problem_example_b} shows a possible solution of the problem.

The robot starts growing from its container, namely the \textit{home base}. Given the pose (position and orientation) of the home base and the set of targets to be reached, the optimizer retrieves an optimal design of the robot that allows it to reach all targets while satisfying hardware constraints. The design parameters include the number of joints and their location along the robot: these parameters need to be determined before the robot is manufactured and assembled -- i.e., the lengths of the links of the desired robot. However, the number of joints/links that a soft robot can actually use during the task may vary based on the target to be reached. Also, the length of the final link may vary, as it depends on how much the robot has grown to reach a target. In fact, unlike rigid robot links for which the design space can be simply represented by the length of each link, the soft-growing robot's design space has a set of unique characteristics owing to its ability to grow and retract: (i) different targets might require different numbers of links to be reached; (ii) any interior link has a fixed length across different targets while the length of the final link can vary for different targets, though not exceeding its maximum length (which is equal to the length a link would have when acting as an interior link); (iii) a link can be interior or final for different targets; and (iv) the number of links required for a given set of targets is unknown. As shown in Fig.~\ref{fig:problem_example_b}, a single robot with fixed link lengths can be designed to reach all targets with a given pose: each configuration $\zeta_i$ of the robot shares the same link lengths except for the last one, as the specific target $t_i$ might not require further growth to be reached and the rest of the robot would still be rolled inside its body.

This scenario can be modeled as an optimization problem: starting from a random solution (random lengths for each link and random angles for each joint of each configuration), we aim to minimize the distance between the pose of the tip of the robot and the pose of the targets using forward kinematics. The optimizer can be considered an inverse kinematics solver, eventually retrieving the set of angles and link lengths (design) to reach each target.

\subsection{Problem formulation and representation of the decision variables}

The mathematical formulation of the problem uses the following terms:

\begin{itemize}
    \item $n$, the maximum number of links/joints;
    \item $\theta$, joint rotation around z-axis; and
    \item $l$, link length.
\end{itemize}
Additionally, we can define as \textit{node} any endpoint of a link, which also corresponds to a joint except for the last one that will be the tip of the robot (i.e., the end effector); therefore, a robot will have at most $n+1$ nodes.

Bounds on these parameters will be defined by the user depending on the resources available during manufacturing. The number of links $n$ is defined in $\mathbb{N}^+$, and it ultimately depends on how many links a manufacturer is willing to build. 
All angles are bounded in a range $\Delta_\theta \in \mathbb{R}$ that is defined by hardware constraints (e.g., each joint cannot steer more than $\pi/4$); except for the first joint at the base as the robot everts orthogonally from it (i.e., either fixed to zero or range with no bounds while we manually orientate the home base for each target).
Link lengths are bounded in a range $\Delta_l \in \mathbb{R}$ and depend primarily on manufacturing constraints (e.g., if the links are made in polyethylene plastic and are produced with a laser cutter, the maximum length is defined by the size of the laser cutter), but also on the gripping mechanism that is used for grasping (i.e., the size of the link must be at least as long as the gripper).

A specific task is defined with the following:
\begin{itemize}
    \item $t$, a set of targets, each represented by their pose in the space (position and orientation);
    \item $o$, a set of obstacles, each represented by their position in the space and other features that will be defined based on their geometry\footnote{As described in Sec.~\ref{sec:constraints}, we represent all obstacles as circles, therefore the features of the obstacle only include their radius.}; and
    \item $h$, the pose of the home base (position and orientation).
\end{itemize}

The output of the optimization problem is composed of a design, a solution, and a set of configurations (the number of configurations must be equal to the number of targets $t$):

\begin{itemize}
    \item a single configuration $\zeta$ is defined as in (\ref{eq:configuration}), and it is composed by the steering angle at each joint and the length of each link;
    
    \begin{equation}\label{eq:configuration}
        \begin{aligned}
            \zeta = 
            \begin{bmatrix}
                \theta_1 & l_1\\
                \vdots & \vdots\\
                \theta_n & l_n\\
            \end{bmatrix}
        \end{aligned}
    \end{equation}
    
    \item multiple configurations define a solution\footnote{A solution of an optimization problem is a specific combination of values for the decision variables of the problem; also referred to as \textit{individual} in evolutionary computation.} of the problem $\psi$, defined as in (\ref{eq:solution}); and
    
    \begin{equation}\label{eq:solution}
        \begin{aligned}
            \psi = 
            \begin{bmatrix}
                \theta_{1,1} & \cdots & \theta_{1,n}\\
                \vdots & \ddots & \vdots\\
                \theta_{|t|,1} & \cdots & \theta_{|t|,n}\\
                l_1 & \cdots & l_n
            \end{bmatrix}
        \end{aligned}
    \end{equation}
    
    \item the desired design for a specific task is given by the vector $\delta$, defined as in (\ref{eq:design}), which is the values of the link lengths of the robot, as well as the last row of the matrix $\psi$ as all configurations share the same design.
    
    \begin{equation}\label{eq:design}
        \begin{aligned}
            \delta = 
            \begin{bmatrix}
                l_1 & \cdots & l_n
            \end{bmatrix}
        \end{aligned}
    \end{equation}
    
\end{itemize}

\begin{figure}[ht]
    \centering
    \subfigure[\protect\url{}\label{fig:evertedConfiguration_a}]%
    {\includegraphics[width=6.75cm]{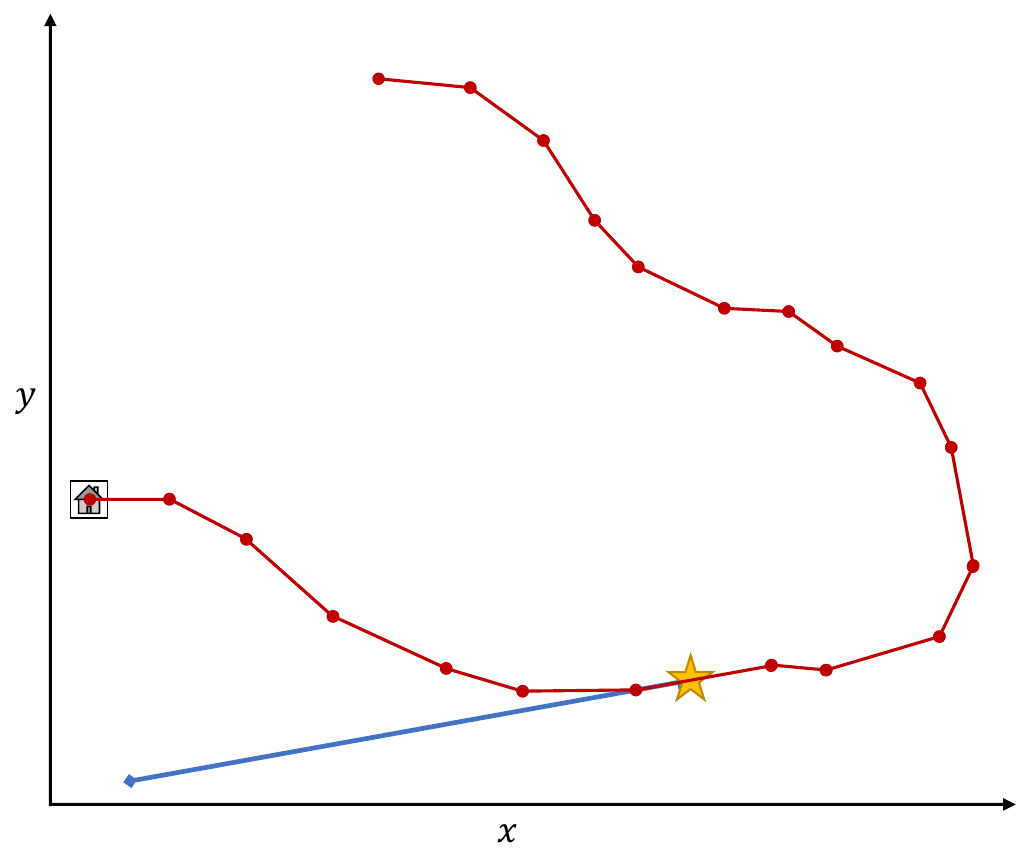}}
    \subfigure[\protect\url{}\label{fig:evertedConfiguration_b}]%
    {\includegraphics[width=6.75cm]{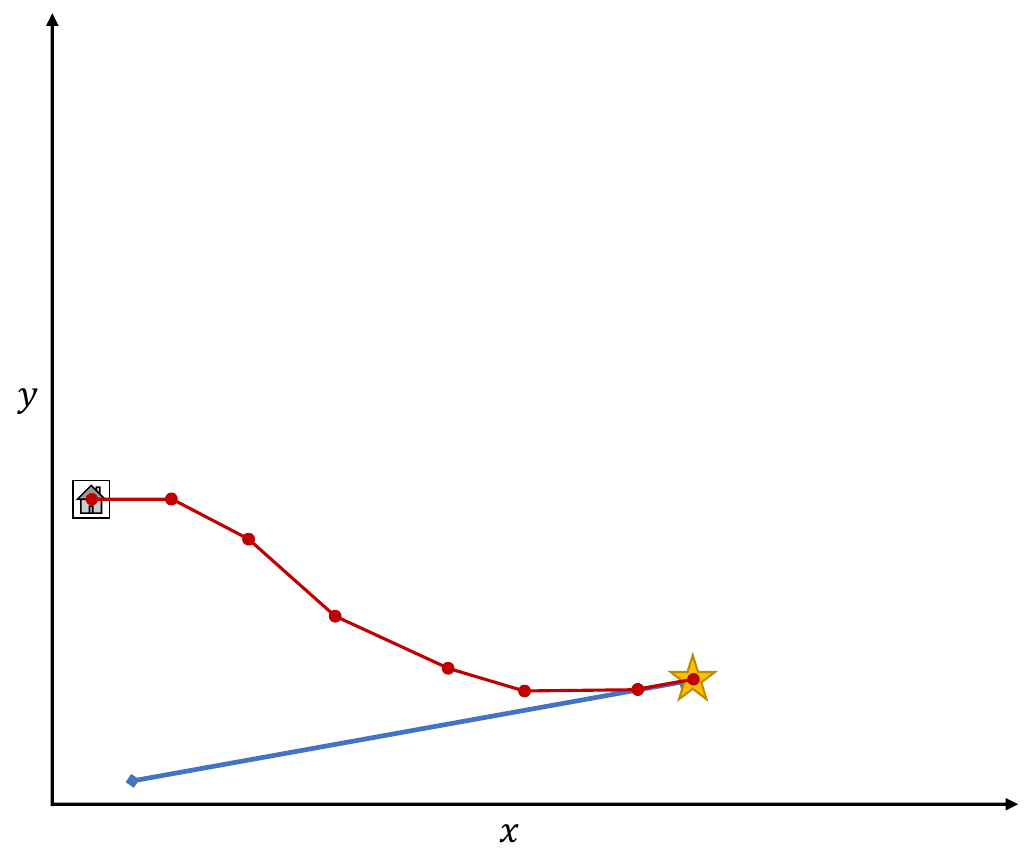}}
    \caption{In (a), the phenotype of a robot generated with the genotype described in (\ref{eq:solution}): in this codification, a random robot is generated with $n=20$ links. However, as shown in (b), only seven of them are actually employed to reach the target correctly ($\bar{n} = 7$). Furthermore, the last link is not fully everted (contrary to the internal links composing the rest of the body): this information is codified by extending the genotype as in (\ref{eq:solution_withExtra1}). In conclusion, the other thirteen links after the end effector do not need to be included in the design $\delta$.}
    \label{fig:evertedConfiguration}
\end{figure}

The tip (end effector) of the robot can be placed anywhere on any link, due to the eversion DoF, as shown in Fig.~\ref{fig:evertedConfiguration}. Indicating with $\bar{n}$ the $i$-th link right before the tip of the robot, the design $\delta$ can have the last $n-\bar{n}-1$ values set to null in case $n \neq \bar{n}$ -- i.e., some of the $n$ links are not used to reach some target and the last one before the tip is only partially everted. As mentioned earlier, an optimized design is created to reach multiple targets, even though the end effector reaching each target (i.e., for each configuration) is not on the last link. For example, a robot designed with five links ($n=5$) might reach one of the targets after the second link ($\bar{n}=2$): in this case, the configuration for that target will fully grow the first two links of $l_1$ and $l_2$, respectively; the third one will not grow of $l_3$, but only until the extent needed to reach the target; whereas the last two links will have their values, $l_4$, and $l_5$ set to null and will be considered to be still rolled in the robot. Fig.~\ref{fig:problem_example_b} shows an example of a solution for a specific given task with three configurations having a different number of employed links.
It is clear that the values of link lengths retrieved in the design $\delta$ are a reference only for the furthest target to the home base, and different configurations can use partial information from $\delta$ depending on how many links are employed. This requirement must be taken into account when codifying the data structure\footnote{In evolutionary computation, this would be the genotype of individuals, or \textit{chromosome}.} of solutions $\psi$, whose elements define the decision variables of the optimization problem.

Therefore, the following additions are made to the structure of the solutions:

\begin{itemize}
    \item $\bar{n}$, the number of employed nodes for each configuration (note that $\bar{n} \leq n$); and
    \item $l_{\bar{n}}$, the length of the last link for each configuration, right before the end effector.
\end{itemize}
The matrix $\psi$ then is defined as in (\ref{eq:solution_withExtra1}), where end effector indices and last link lengths are added at the end of each row, identifying different configurations. To keep a matrix shape, the last two elements of the last row are set to zero -- the last row refers to the link lengths, which are fixed for a single design. As we will see in Sec.~\ref{sec:ik}, this is not the only information we need to add to our solution matrix.

\begin{equation}\label{eq:solution_withExtra1}
    \begin{aligned}
        \psi = 
        \begin{bmatrix}
            \begin{bmatrix}
                \theta_{1,1} & \cdots & \theta_{1,n}\\
                \vdots & \ddots & \vdots\\
                \theta_{|t|,1} & \cdots & \theta_{|t|,n}\\
                l_1 & \cdots & l_n
            \end{bmatrix}
            &
            \begin{bmatrix}
                \bar{n}_1 & l_{\bar{n}_1}\\
                \vdots & \vdots \\
                \bar{n}_{|t|} & l_{\bar{n}_{|t|}}\\
                0 & 0
            \end{bmatrix}
        \end{bmatrix}
    \end{aligned}
\end{equation}


\subsection{Search space}
\label{sec:search_space}

\begin{figure}[ht]
	\centering
	{\includegraphics[width=7cm]{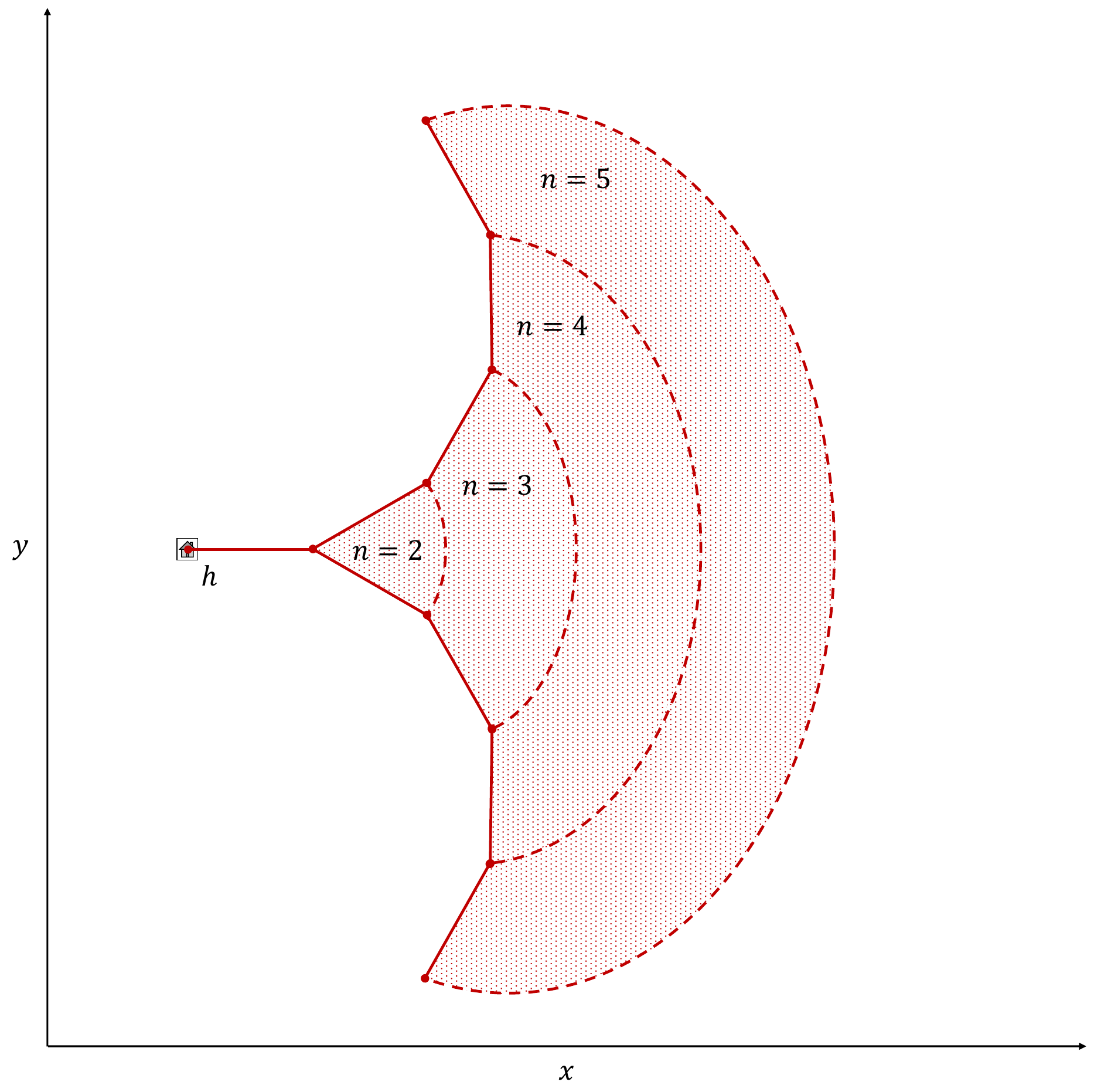}}
	\caption{Workspace of a robot having $2$ to $5$ links of fixed lengths, where each joint can steer in a range of $\Delta_\theta = [+\pi/6,-\pi/6]$. In the figure, the home base has a fixed orientation, meaning that the first joint cannot steer and the robot everts orthogonally from the base. Any point of the workspace can be reached by the end effector of the robot (with a specific orientation) as the robot can grow and shrink.}
	\label{fig:workspace}
\end{figure}

A greater number of links results in greater dexterity and a larger workspace for the robot, as shown in Fig.~\ref{fig:workspace}.
Therefore, the optimization problem must deal with a large search space $\omega$, which linearly increases with the number of targets $|t|$ as it must optimize a unique set of links for all targets. 
The search space $\omega$ is defined as in (\ref{eq:search_space}) when the order of targets is known, or the robot retracts back to base after reaching a target (i.e., the order of target does not matter). 

\begin{equation}\label{eq:search_space}
    \begin{aligned}
        \omega = n^{\Delta_l} + |t| \cdot n^{\Delta_\theta}
    \end{aligned}
\end{equation}

On the other hand, when the order of targets is unknown and it is required to obtain a solution that also minimizes the cost to go from one configuration to the next one, the search space increases factorially, as it should consider each permutation for the order of targets. In this work, we do not consider the order of targets, and we only focus on solving the basic optimization problem. Future works will investigate the best strategy, including path-planning algorithms, to evaluate the more efficient order of reaching targets.

\subsection{Fitness evaluation from the objective functions}
\label{sec:fitness}

The problem aims at optimizing the following objectives:

\begin{enumerate}
    \item $f_1(\psi, t, h, o)$: minimize the distance between the end effector of the robot (which could be anywhere along the body) and each target in the environment -- solve the inverse kinematics;
    \item $f_2(\psi, t, h, o)$: minimize the difference between the orientation of the last link and the desired reaching orientation for each target in the environment -- solve the inverse kinematics; 
    \item $f_3(\psi, t, h, o)$: minimize the robot's components (i.e., its length and number of links), to favor faster and cheaper production.
\end{enumerate}

\begin{figure}[t]
	\centering
	{\includegraphics[width=6.75cm]{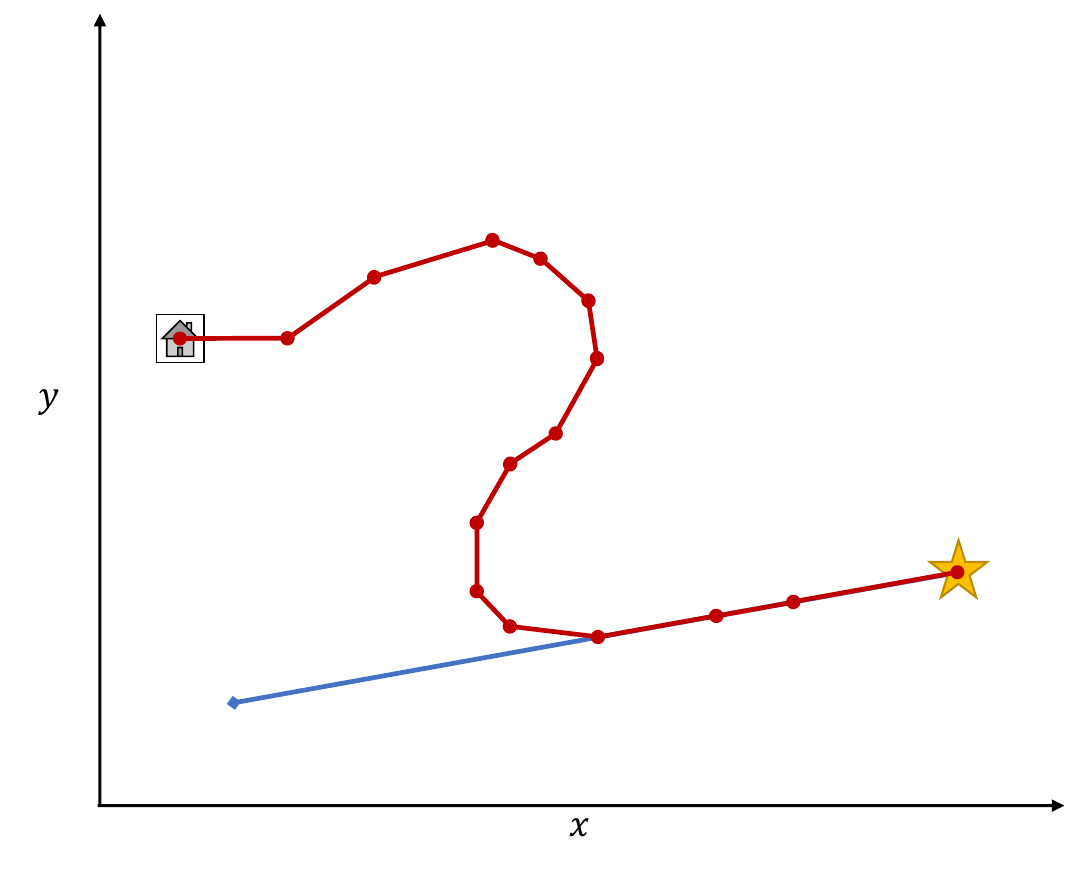}}
	\caption{Instead of directly aiming for the only target in the environment, this robot steers in random directions before finalizing the task, resulting in a ``wavy'' configuration.}
	\label{fig:wavy}
\end{figure}

\begin{figure}[h!]
    \centering
    \subfigure[\protect\url{}\label{fig:tradeoff_nodesOrient_1}]%
    {\includegraphics[width=6.75cm]{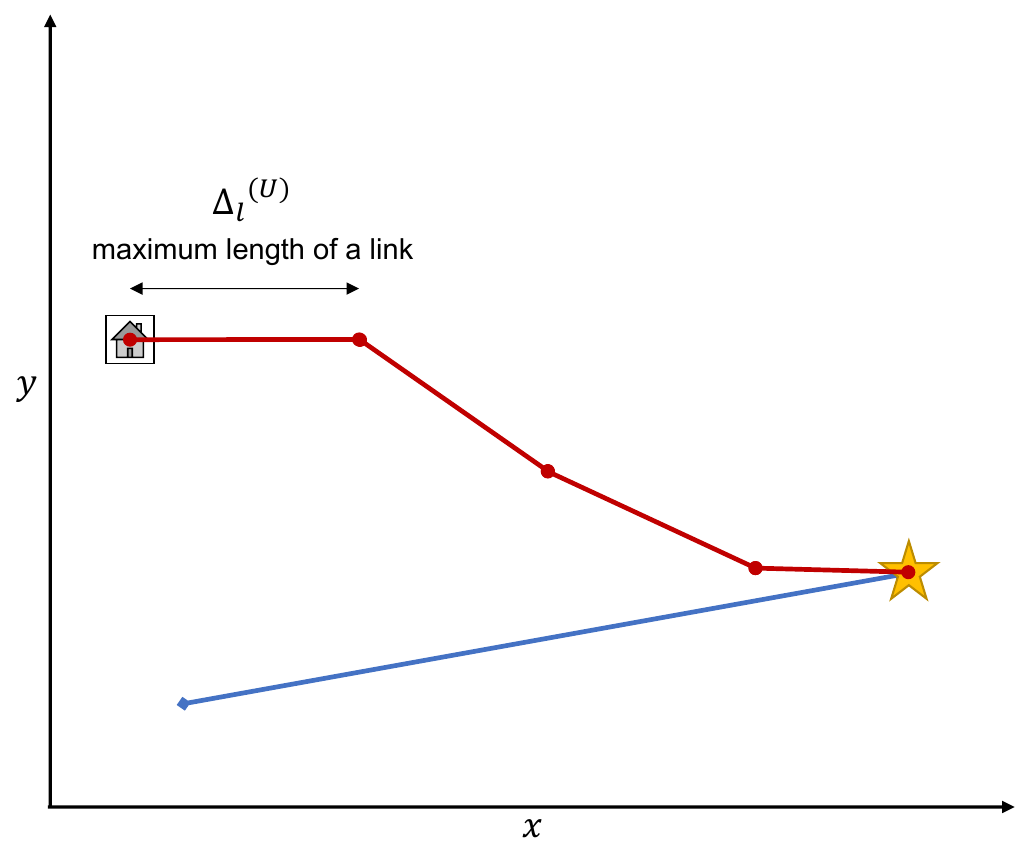}}
    \subfigure[\protect\url{}\label{fig:tradeoff_nodesOrient_2}]%
    {\includegraphics[width=6.75cm]{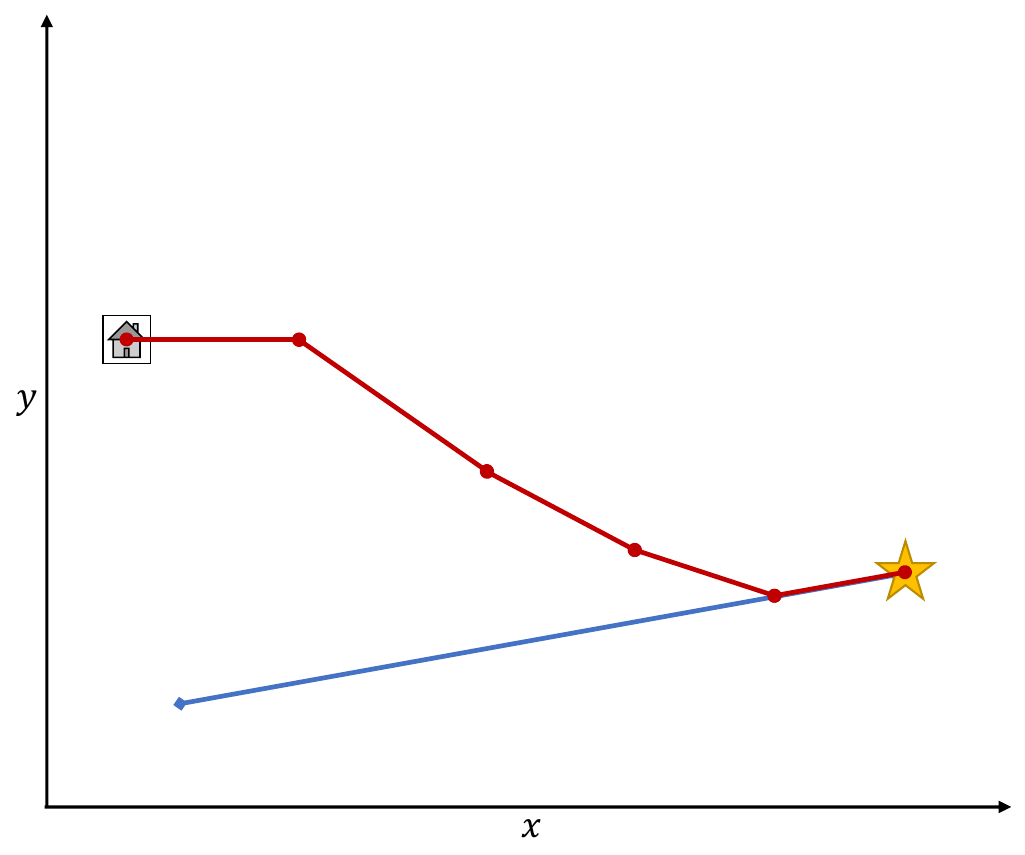}}
    \caption{Trade-off problem between the number of employed links and reaching orientation. In (a), the configuration reaches the target with the minimum amount of links possible (it is assumed that all internal links have maximum length based on their bounds); however, the robot cannot reach the target with the desired orientation. In (b), one more link is employed to ensure the robot reaches the target with the right orientation.}
    \label{fig:tradeoff_nodesOrient}
\end{figure}
When dealing with multi-objective optimization problems (MOOPs), some or all the objectives might conflict with each other, meaning that the optimization of one objective happens to the detriment of another objective. Therefore, instead of retrieving a single solution, we usually retrieve a set of \textit{trade-off} solutions that define the so-called Pareto optimal front. Solutions lying in this front are \textit{non-dominated} by any other solution of the problem, meaning that no other possible solution is better than those in all objectives; however, being trade-off solutions, they might be optimal for one objective but not for others \cite{deb1999multi,miettinen2012nonlinear}. 
In the scenario of our soft robots, treating all objectives as equally important would generate some trade-off solutions that are not interesting for our purposes: for example, it might be impossible to reach the target with the minimum number of links (which would trivially be a single-link robot); however, also having a robot reaching the target with lots of \textit{useless} links (i.e., a ``wavy'' robot, as in Fig.~\ref{fig:wavy}) is definitely not the optimal solution we are looking for. It is obvious that in this problem we are not interested in a robot that does not reach the target, but we might still need a trade-off between the properties that define a robot (e.g., number of links) and reaching orientation. 
For example, in a solution like the one shown in Fig.~\ref{fig:tradeoff_nodesOrient_1}, the minimization of the number of links happens to the detriment of the reaching orientation; whereas in Fig.~\ref{fig:tradeoff_nodesOrient_2}, one more link is employed to fully satisfy the orientation requirement. Sometimes, it might be impossible to reach a target with the exact desired orientation, especially when more than one target composes the task: in that case, a trade-off is inevitable; however, when possible, it is preferable to use more links to accommodate reaching orientation -- hence, $f_2$ has priority over $f_3$, and $f_1$ has overall priority. 

Given the above considerations, the problem becomes \textit{reaching the target's pose by employing the minimum amount of resources}. Therefore, firstly we should ensure that the configuration under investigation reaches the target ($f_1$), and then we minimize its components as well. Note that, even though we need the robot to exactly reach the target, $f_1$ should not be considered as a constraint (i.e., all solutions having tip-target distance larger than zero are infeasible solutions); otherwise, the search algorithm will lose its heuristic, and convergence would be random\footnote{This claim can be debatable based on the method used for constraint handling: it would be true with death or static penalty methods, but it might be interesting to investigate results with dynamic methods or methods based on separation~\cite{coello2002theoretical}, even though the risk of obtaining only infeasible solutions is high.}.

Solving the problem with methods that retrieve Pareto optimal solutions, namely \textit{generating methods} \cite{cohon1985multicriteria}, might not be ideal for our case unless we purposely cut part of the objective space; instead, we want to combine the three objectives into a single function, namely a \textit{fitness} function. Besides the advantage of enforcing priority on the first objective, single-objective optimization methods have lower computational complexities than multi-objective ones.
In a previous work \cite{exarchos2022task}, we used a weighted-sum method to convert multiple objectives into a single one by multiplying each objective with a non-negative weight. 
Albeit being the simplest way to address the problem \cite{miettinen2012nonlinear, hwang2012multiple}, this method is still considering trade-off solutions distributed along the Pareto-optimal front: different values of weights would generate different Pareto-optimal solutions, which does not really solve the problem of enforcing $f_1$ no matter what solution we are generating. Additionally, these values must be chosen a priori, which means assigning an exact value of preference to each objective: although in this scenario we might have a clear idea of what the priorities are, this method requires choosing numerical parameters for each preference (i.e., weight), which is always a challenge. Lastly, this method would fail at retrieving solutions on non-convex Pareto optimal fronts, which is likely our case. In conclusion, the weighted-sum method is not the best option for our purposes.

Therefore, the solution we propose formalizes the optimization problem by merging some objectives into a single function instead of considering them as separate objectives, which is possible due to the geometrical properties of the problem. By prioritizing that single function, we ensure that all of the included objectives are optimized, reducing the need for trade-offs. The remaining objectives will be optimized following a MOOP preference-based method rather than a generic/ideal one, by employing a partitioning method that will partially explore the Pareto optimal front and retrieve only the solution of interest.

\subsubsection{Solving the inverse kinematics}
\label{sec:ik}

The objective function $f_1$ (minimize tip-target error) is the obvious strategy for solving the inverse kinematics of a rigid manipulator. Given a configuration in polar coordinates $\zeta^p$, we can retrieve the corresponding configuration in Cartesian coordinates $\zeta^c$ through forward kinematics as in (\ref{eq:forwardkinematics}). The configuration $\zeta^c$ comprises $n+1$ points -- i.e., all joints plus the end effector. From $\zeta^c$, the function $f_1$ simply becomes the $\ell^2$ norm between the last point and the target $\norm{\zeta^c_{n+1}-t_i}^2$ -- for simplicity, we are implying that only the position x-y of the target $t_i$ is taken from the pose (which also includes orientation as discussed earlier). 

\begin{equation}\label{eq:forwardkinematics}
    \begin{aligned}
        \zeta^p = 
        \begin{bmatrix}
            \theta_1 & l_1\\
            \vdots & \vdots\\
            \theta_n & l_n\\
        \end{bmatrix}
        \quad
        \xrightarrow[kinematics]{forward}
        \quad
        \zeta^c = 
        \begin{bmatrix}
            x_1 & y_1 \\
            \vdots & \vdots\\
            x_{n+1} & y_{n+1} 
        \end{bmatrix}
    \end{aligned}
\end{equation}
Unlike rigid manipulators, in the case of soft-growing robots, we must consider that the end effector could be anywhere along their body; therefore, $f_1$ becomes the minimum among the distances between each node of the robot and the target. Extending for $|t|$ targets and therefore $|t|$ configurations in a single solution $\psi$, the fitness is expressed as in (\ref{eq:f1}), where the $\ell^2$ norm is evaluated for each node of each configuration against each target, and the objective becomes the summation of the minimum distances of each configuration. Note that the index $j+1$ indicates that we did not include the home base in the evaluation because it is a fixed point that does not contribute to the heuristic.

\begin{equation}\label{eq:f1}
    \begin{aligned}
        f_1 = \sum_{i=1}^{|t|} {\min_{\forall j \in n}{\norm{\zeta^c_{(i,j+1)}-t_i}^2}}
    \end{aligned}
\end{equation}

\begin{figure}[ht]
    \centering
    \subfigure[\protect\url{}\label{fig:fit_ikf1}]%
    {\includegraphics[width=6.75cm]{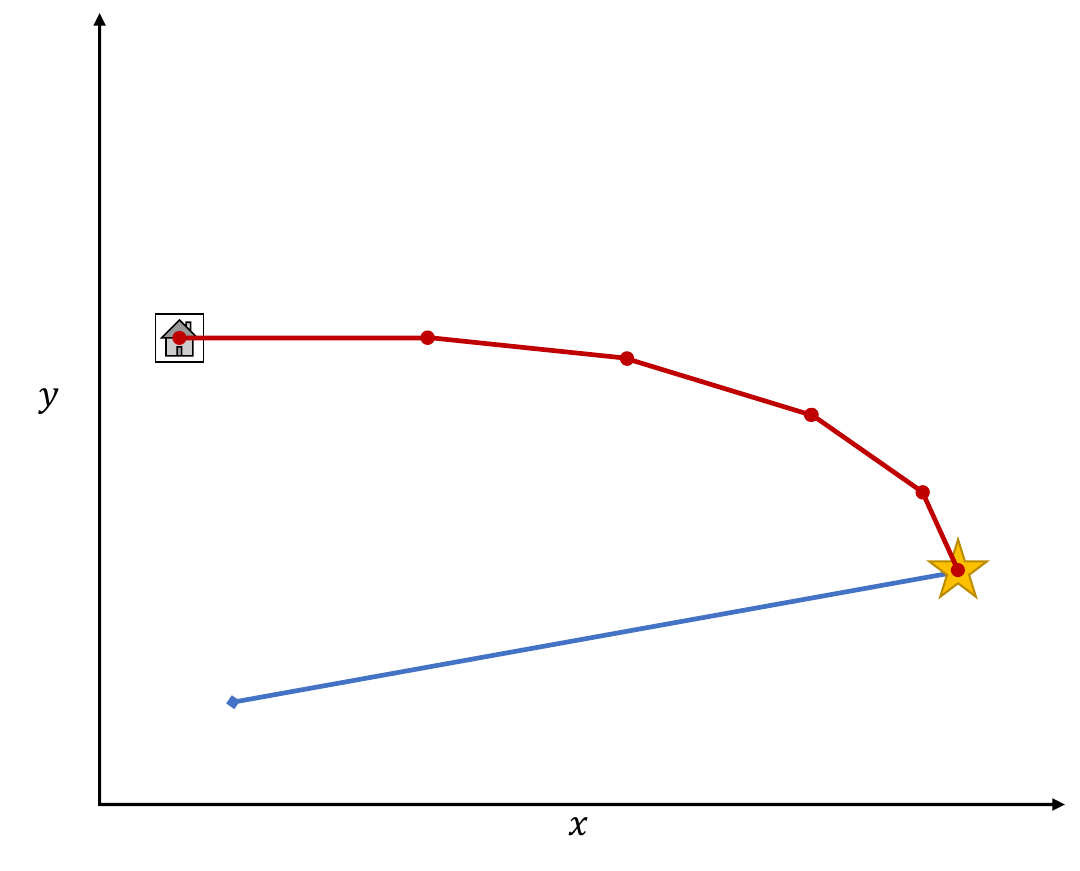}}
    \subfigure[\protect\url{}\label{fig:fit_ikf1f2}]%
    {\includegraphics[width=6.75cm]{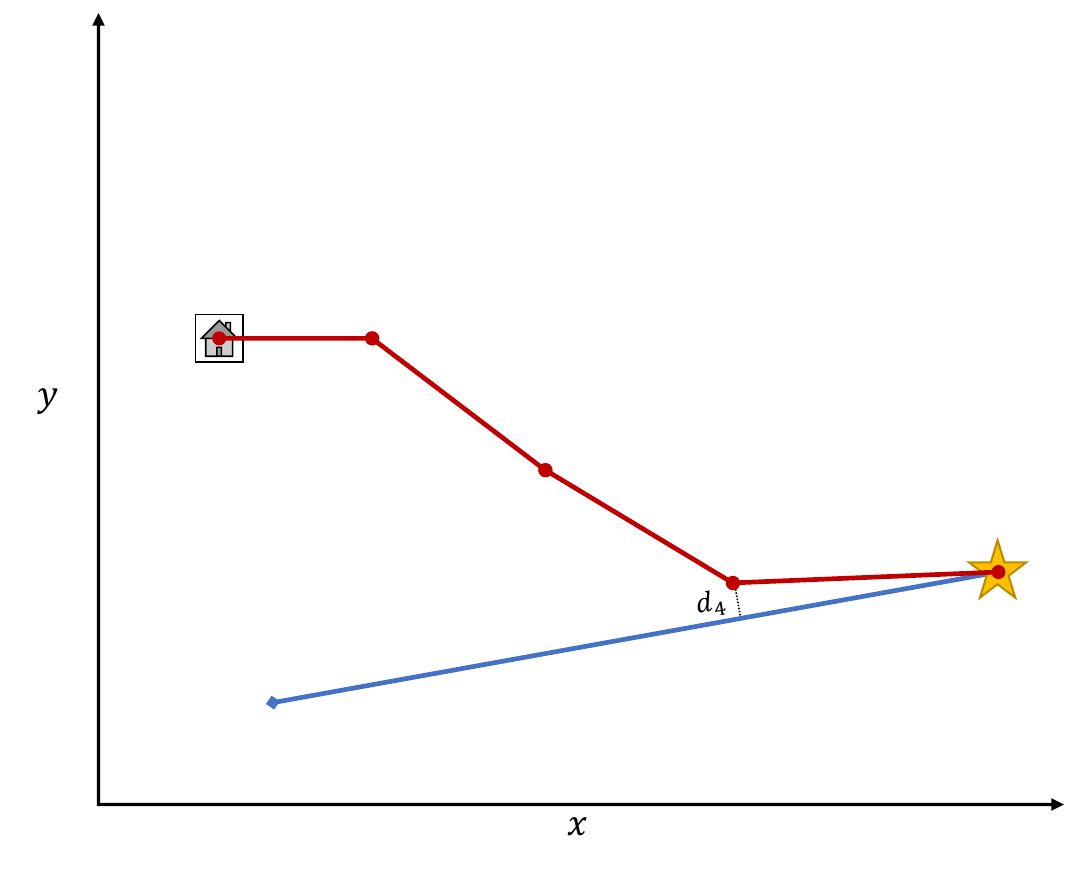}}
    \caption{In (a), a configuration obtained by only minimizing the distance between the end effector and the target (inverse kinematics). Although the robot reaches the target correctly, it does not take into account the desired reaching orientation. In (b), a configuration is obtained by minimizing the distance between the orientation segment and the closest node to it  (the fourth one). By minimizing the latter, the formulation also ensures that the orientation is considered. In both (a) and (b), the rest of the links generated in $\zeta$ are discarded and replaced with segments that directly aim for the target (having $\epsilon = 6$ and $\epsilon = 4$ for (a) and (b), respectively). }
    \label{fig:fit_ikf}
\end{figure}

\subsubsection{Reaching the desired orientation}
\label{sec:reaching_orientation}

Minimizing $f_1$ to solve the robot's inverse kinematics would generate solutions like the one shown in Fig.~\ref{fig:fit_ikf1}, where the robot reaches the target with any orientation. To include also $f_2$ (minimize error in reaching orientation), instead of aiming directly for the target, we aim at the target's orientation straight line as shown in Fig.~\ref{fig:fit_ikf1f2}. Indicating with $s_i$ a segment lying on the target's orientation straight line (having $t_i$ as one endpoint and a length within the workspace $\omega$), the fitness function is expressed as in (\ref{eq:f1f2}), where the distance between a node of the robot and the segment $s_i$ is calculated with the point-segment distance \cite{stroppa2018convex} as reported in Algorithm \ref{alg:pointSegmentDistance}. This calculates the distance from the point to its projection on the segment or to one of its closest edges.

\begin{equation}\label{eq:f1f2}
    \begin{aligned}
        f_{1-2} = \sum_{i=1}^{|t|} {\min_{\forall j \in n}{(\texttt{pointSegmentDistance}(\zeta^c_{(i,j+1)},s_i)})}
    \end{aligned}
\end{equation}

\IncMargin{1em}
\begin{algorithm}
	\SetKwData{P}{p}
	\SetKwData{R}{R}
	\SetKwData{V}{v}
	\SetKwData{Dist}{d}
	\SetKwData{C}{c}	
	\SetKwFunction{Atan}{atan2}	
	\SetKwFunction{Sin}{sin}	
	\SetKwFunction{Cos}{cos}	
	\SetKwInOut{Input}{input}\SetKwInOut{Output}{output}
	
	\Input{The point $\P = \langle x,y \rangle$ to be compared with the segment}
	\Input{The first endpoint $\V_1 = \langle x,y \rangle$ of the segment}
	\Input{The second endpoint $\V_2 = \langle x,y \rangle$ of the segment}
	\Output{The distance \Dist between the point \P and the segment $\overline{\V_1\V_2}$}
	\BlankLine		
	
	\Begin{		
		$\C \leftarrow (\V_1 + \V_2) / 2$\; 
		$\V_1 \leftarrow \V_1 - \C$\; 
		$\V_2 \leftarrow \V_2 - \C$\; 
		$\P \leftarrow \P - \C$\; 
		
		$\beta \leftarrow $ \Atan{$\V^{(y)}_2, \V^{(x)}_2$}\; 
		$\R \leftarrow 
		    \begin{bmatrix}
                \Cos(-\beta) & -\Sin(-\beta)\\
                \Sin(-\beta) & \Cos(-\beta)
            \end{bmatrix}$\; 
		$\V_1 \leftarrow  \R \cdot {\V_1}'$\; 
		$\V_2 \leftarrow  \R \cdot {\V_2}'$\; 
		$\P \leftarrow  \R \cdot \P'$\; 
						
		\uIf{$\P^{(x)} < \V^{(x)}_1$}{	
				$\Dist \leftarrow \norm{\P - \V_1}^2$\;	
		}
		\uElseIf{$(\V^{(x)}_1 \leq \P^{(x)}) \wedge (\P^{(x)} \leq \V^{(x)}_2)$}{
			$\Dist \leftarrow |\P^{(y)}|$\;		
		}
		\Else{
			$\Dist \leftarrow \norm{\P - \V_2}^2$\;	
		}		
		\KwRet{$\Dist$};
	}
	
	\caption{Point-Segment Distance~\cite{stroppa2018convex}}\label{alg:pointSegmentDistance}
\end{algorithm}\DecMargin{1em}

When using the fitness in (\ref{eq:f1f2}), the distance to be minimized is not identified by the end effector but by the closest node to the target's orientation segment. With reference to the example in Fig.~\ref{fig:fit_ikf1f2}, the closest node to the segment is the fourth one; however, the robot needs to grow further to reach the target, leading the end effector to be the fifth node. Once the target's orientation segment is reached by any robot node, the rest of the links can grow in the direction of the target until the reaching is achieved. Therefore, the configuration data following the closest node to the target's orientation segment must be discarded and considered as not everted and still rolled in the robot. We indicate with: 

\begin{itemize}
    \item $\epsilon$, the index of the closest node to the target's orientation segment, namely the \textit{last employed node} of a configuration; and
    \item $\theta_{\epsilon}$, the angle for the $\epsilon$-th link of the robot to be aligned to the target's orientation segment.
\end{itemize}
When the distance defined in (\ref{eq:f1f2}) is optimal, then $\theta_{>\epsilon}=0$, and the robot simply grows in the direction of the target's orientation segment. The notation for a solution $\psi$ reported in (\ref{eq:solution_withExtra1}) is then modified as in (\ref{eq:solution_withExtra2}) to include this extra information. Note that the sub-matrix on the left-hand side -- as defined in (\ref{eq:solution}) -- is not reduced in size to keep the dimension of the chromosomes consistent within the population and throughout all generations; however, the actual length of each configuration may change based on the pose of the respective target also, note that the elements in the last row after $l_n$ are zeros to maintain a matrix form\footnote{We remind the readers that this code was developed in MATLAB; readers who want to implement their own code in a different language can create a more efficient data structure without using further resources and/or increasing the efficiency of the algorithm by including fitness in the data structure and use priority queues instead of sorting algorithms.}.

\begin{equation}\label{eq:solution_withExtra2}
    \begin{aligned}
        \psi = 
        \begin{bmatrix}
            \begin{bmatrix}
                \theta_{1,1} & \cdots & \theta_{1,n}\\
                \vdots & \ddots & \vdots\\
                \theta_{|t|,1} & \cdots & \theta_{|t|,n}\\
                l_1 & \cdots & l_n
            \end{bmatrix}
            &
            \begin{bmatrix}
                \epsilon_1 & \theta_{\epsilon,1} & \bar{n}_1 & l_{\bar{n},1}\\
                \vdots & \vdots & \vdots & \vdots \\
                \epsilon_{|t|} & \theta_{\epsilon,|t|} & \bar{n}_t & l_{\bar{n},|t|}\\
                0 & 0 & 0 & 0
            \end{bmatrix}
        \end{bmatrix}
    \end{aligned}
\end{equation}

\begin{figure}[b!]
    \centering
    \subfigure[\protect\url{}\label{fig:f3_bad}]%
    {\includegraphics[width=6.75cm]{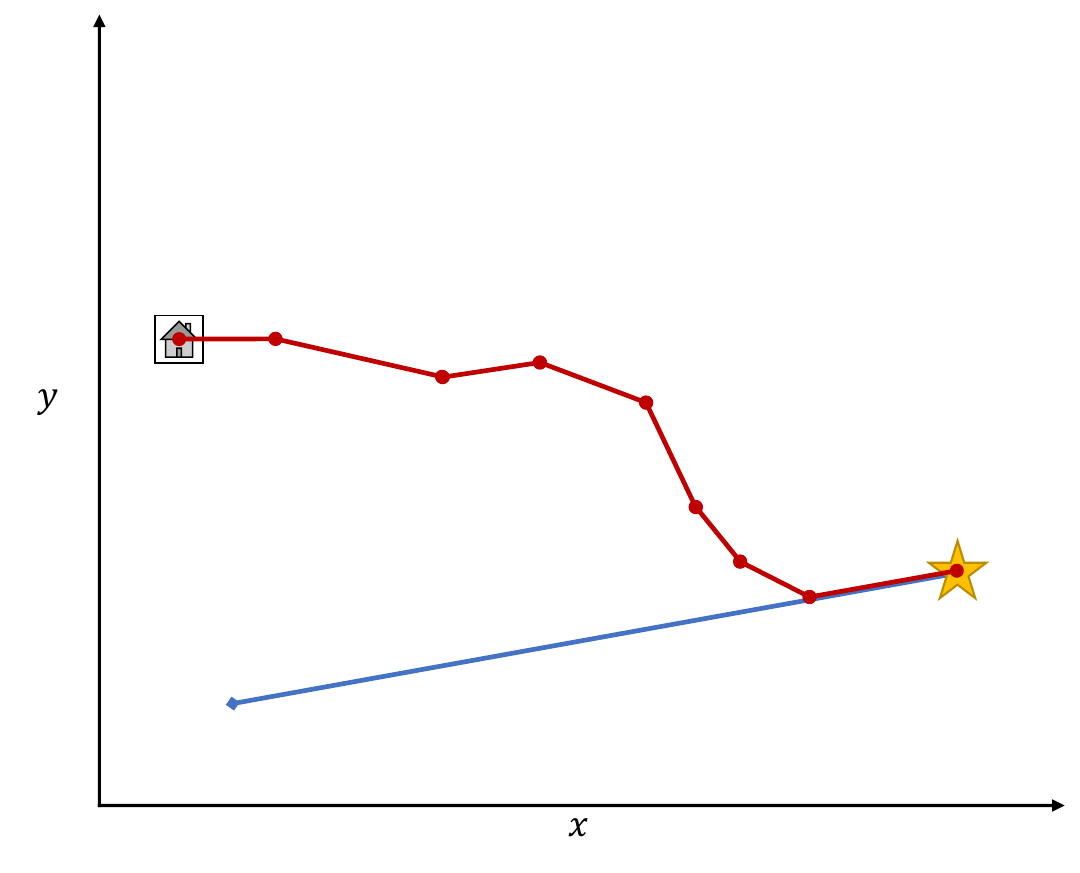}}
    \subfigure[\protect\url{}\label{fig:f3_best}]%
    {\includegraphics[width=6.75cm]{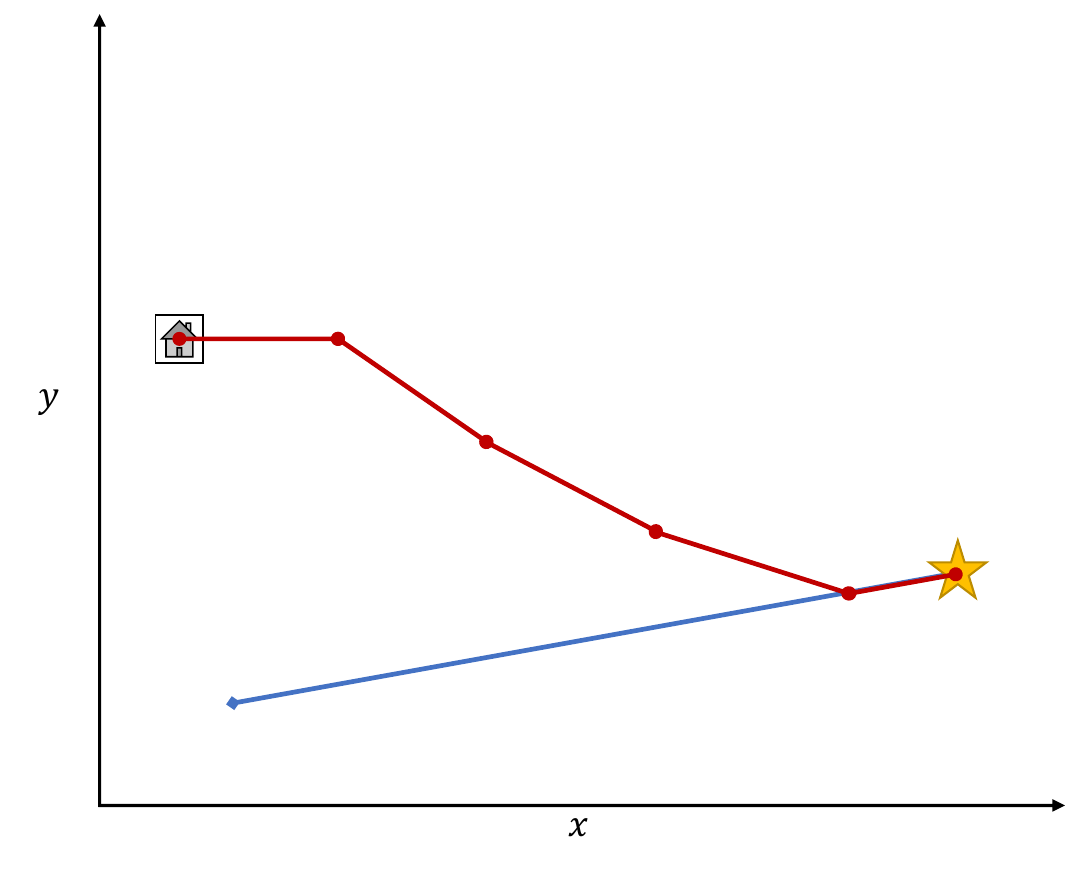}}
    \caption{If the optimization problem only minimized the distance between the orientation segment and the closest node to it, both solutions in (a) and (b) would be considered equally good in terms of fitness. However, the robot in (a) shows a curvature at the first five nodes that makes the design uselessly long, wasting resources in building more links than required. On the other hand, the robot shown in (b) immediately starts steering toward the target and must be identified as a better solution by the optimizer.}
    \label{fig:f3}
\end{figure}

\subsubsection{Optimizing the kinematic components of the robot}
\label{sec:optimizing_components}

The objective functions $f_1$ and $f_2$ (inverse kinematics) have priority over $f_3$ (optimal components); once the first two have been incorporated in a single fitness function $f_{1-2}$, it is time to penalize long solutions with lots of nodes. Ideally, we would prefer a robot that gets progressively close to the target's orientation segment forming a gentle curve. The robot shown in Fig.~\ref{fig:f3_best} is a better solution than the one shown in Fig.~\ref{fig:f3_bad}: in addition to having all nodes closer to the orientation segment, it also employs fewer nodes to reach the target. 

Objective $f_3$ aims at optimizing the components of the robot to reduce manufacturing time and resources, and cannot be described by a single property. Therefore, we consider it as a macro objective that includes the following:

\begin{itemize}
    \item $f_{3.1}$: minimize the number of joints/links of each configuration to reduce redundant DoFs, as in (\ref{eq:f_31});

    \begin{equation}\label{eq:f_31}
        \begin{aligned}
            f_{3.1} = \sum_{i=1}^{|t|} \bar{n}_i 
        \end{aligned}
    \end{equation}
    
    \item $f_{3.2}$: minimize the overall length of the robot to favorite shorter configurations, as in (\ref{eq:f_32}); and

    \begin{equation}\label{eq:f_32}
        \begin{aligned}
            f_{3.2} = {\max_{\forall i \in |t|}{\left( \left( \sum_{j=1}^{\bar{n}_i - 1}{l_j} \right) + l_{\bar{n}_i} \right)}}
        \end{aligned}
    \end{equation}
    
    \item $f_{3.3}$: minimize the change in direction between links in all configurations, or \textit{undulation}, to favorite non-intersecting and non-``wavy'' trajectories, calculated in percentage over the length of all configurations as in (\ref{eq:f_33}). 

    \begin{equation}\label{eq:f_33}
        \begin{aligned}
            f_{3.3} = 
            \frac{\sum_{i=1}^{|t|}{\left( \frac{\sum_{j=1}^{\epsilon_i - 1}{ \left[ \sign{(\theta_{i,j})} \neq \sign{(\theta_{i,j+1})} \; \wedge \; \theta_{i,j} \neq 0 \right]}}{\epsilon_i} \right)}}{|t|} \cdot 100
        \end{aligned}
    \end{equation}
    
\end{itemize}
Priorities among these three sub-objectives can be discussed to properly tune our preference-based method, ultimately without the need for weight parameters. 
The number of links and the overall length of the robot ($f_{3.1}$ and $f_{3.2}$, respectively) might seem to have a linear relationship (i.e., the more links, the longer the robot), but that is actually not the case: a robot with lots of small links can be shorter than a robot with a few very long ones. We prefer a robot with few links because the time to manufacture a link does not notably depend on its length ($f_{3.1}$ has priority over $f_{3.2}$). 
On the other hand, the number of links and their directions ($f_{3.1}$ and $f_{3.3}$, respectively) have different implications on the shape of the robot: clearly, a robot with lots of links has more chances to be wavy, as shown in Fig.~\ref{fig:f3}; but the question on whether high-DoFs non-wavy robots are better than low-DoFs wavy ones might not be trivial, and it might depend on the type of task (e.g., while solving a maze, the robot might be forced to steer in different directions). Ultimately, we concluded that building fewer links is always better for manufacturing ($f_{3.1}$ has priority over $f_{3.3}$), but other manufacturers might have a different opinion and therefore swap this order.  
Finally, the undulation of a robot and its length ($f_{3.3}$ and $f_{3.2}$, respectively) are highly correlated, as a robot that directly aims at the target results to be shorter than a robot rambling around the environment before reaching its target, as previously shown in Fig.~\ref{fig:wavy}; therefore, we find more important to build non-wavy robots, which will eventually result in shorter configurations ($f_{3.3}$ has priority over $f_{3.2}$).

Additionally, we observed that to improve the quality of retrieved solutions, the objective $f_{3.1}$ should actually be further divided into two sub-objectives:
\begin{itemize}
    \item $f_{3.1a}$: minimize the number of joints/links to reach the target's orientation segment, as in (\ref{eq:f_31a}); and

    \begin{equation}\label{eq:f_31a}
        \begin{aligned}
            f_{3.1a} = \left( \sum_{i=1}^{|t|} \epsilon_i - 1 \right)
        \end{aligned}
    \end{equation}
    
    \item $f_{3.1b}$: minimize the number of joints/links lying on the target's orientation segment, as in (\ref{eq:f_31b}).

    \begin{equation}\label{eq:f_31b}
        \begin{aligned}
            f_{3.1b} = \left( \sum_{i=1}^{|t|} \bar{n}_i - \left(\epsilon_i - 1 \right) \right)
        \end{aligned}
    \end{equation}
\end{itemize}
Thanks to this division, instead of minimizing the overall number of links of the various configurations, we specifically indicate how the configuration must be organized by prioritizing the minimization of the number of links to reach the target's orientation segment; once the segment has been reached, we separately minimize the number of links to finally reach the target. Note that this last point can be simply achieved by manually maximizing the lengths of those links, and although the optimization algorithm should implicitly learn this heuristic, in practice, we might require some artificial fix-ups due to the large size of the search space.

\subsection{Separation of the objectives with Rank Partitioning}
\label{sec:separation_partitioning}

\IncMargin{1em}
\begin{algorithm}
	\SetKwData{K}{k}
	\SetKwData{N}{N}
	\SetKwData{F}{f}
        \SetKwData{Rank}{rank}
        \SetKwData{Ref}{ref}
	\SetKwData{Diff}{d}
	\SetKwData{Sum}{partLimit}
	\SetKwData{Max}{max}
	\SetKwData{Step}{binSize}
	\SetKwData{Str}{start}
	\SetKwData{Stp}{end}
	\SetKwFunction{Sort}{sortrows}
	\SetKwFunction{Fix}{floor}
        \SetKwFunction{Mod}{mod}
        \SetKwFunction{Abs}{abs}
	\SetKwInOut{Input}{input}\SetKwInOut{Output}{output}
	
	\Input{$\N$, the number of individuals in the population.}
	\Input{$\F$, a matrix $\N  \times 9$ having the following information for each individual: fitness for each objective (integer values first followed by real values), fitness after partitioning ($\Rank$), reference to the individual in the population ($\Ref$). A single row of $\F$ is: 
   $[(1)$: $f_{1-2\mathbb{N}}$, $(2)$: $f_{3.1a}$, $(3)$: $f_{3.3}$, $(4)$: $f_{3.1b}$, $(5)$: $f_{3.2\mathbb{N}}$, $(6)$: $f_{1-2\mathbb{R}}$, $(7)$: $f_{3.2\mathbb{R}}$, $(8)$: $\Rank$, $(9)$: $\Ref]_i$ $\forall i \in [1,\N]$. }
	\Input{$\Step$, distance between partitions for each real-coded objective (i.e., one for $f_{1-2}$ and one for $f_{3.2}$.)}
	\Output{The updated matrix $\F$ with the value of $\Rank$ after partitioning.}
	\BlankLine		
	
	\Begin{	
            \For{$i \in [1, \N]$}
            {
                $\F_{(i,1)} \leftarrow \F_{(i,6)} - \Mod(\F_{(i,6)},\Step_{1-2})$\;
                $\F_{(i,5)} \leftarrow \F_{(i,7)} - \Mod(\F_{(i,7)},\Step_{3.2})$\;
            }
            
		\Sort{$\F,>$} on column 1-5 for all rows\;
            $\Diff \leftarrow$ initialize a $\N \times 5$ matrix with zeros\;
            \For{$i \in  [2, \N]$}
            {
                \For{$j \in  [1, 5]$}
                {
                    $\Diff_{(i,j)} \leftarrow \Abs(\F_{(i,j)} - \F_{(i-1,j)})$\;
                }
            }
            
            $\Str \leftarrow 1$\;
            \For{$i \in  [1, \N]$}
            {
                \For{$j \in  [1, 5]$}
                {
                    $\Sum \leftarrow \Sum + \Diff_{(i,j)}$\;
                }
                \uIf{$ \Sum > 0$}
                {
                    $\Stp \leftarrow i$\;
                    \Sort{$\F,>$} on column 6-7 from row \Str to \Stp-1\; 
                    $\Str \leftarrow \Stp$\;
                }
            }
            
            \For{$i \in [1, \N]$}
            {
                $\F_{(i,8)} \leftarrow i$\;
            }
    	\KwRet{$\F$};
        }
	
	\caption{Rank Partitioning}\label{alg:rank_partitioning}
\end{algorithm}\DecMargin{1em}

Given these observations, we propose a strategy that rewards solutions having less and well-organized components only if they are \textit{sufficiently} close to the target's orientation segment. This strategy exploits the concept of population in EC techniques as it compares different solutions/individuals within a set. Population-based algorithms, such as evolutionary algorithms, are the perfect candidates as optimization tools for our purposes. The idea is to divide the population into $k$ partitions with fixed steps based on their kinematic fitness $f_{1-2}$ -- individuals with lower (better) fitness will populate the first partitions, whereas individuals with higher (worse) fitness will be placed in the last partitions. Individuals within the same partition will be considered equals in terms of distance to the target, which means that the best among them are the ones with optimal components. 
Therefore, the procedure will sort individuals within the same partition based on the priorities discussed in Sec.~\ref{sec:optimizing_components}: when a tie occurs, the individuals will be ordered based on the next objective with higher priority, creating nested sub-partitions for each objective. In particular, the procedure will follow this order for each partition:

\begin{enumerate}
    \item distance from the target, or inverse kinematics ($f_{1-2}$);
    \item number of links to reach the segments ($f_{3.1a}$);
    \item amount of undulation among configurations ($f_{3.3}$);
    \item number of links on the segments ($f_{3.1b}$); and
    \item length of the robot or design ($f_{3.2}$).
\end{enumerate}
This operation only works if ties occur within partitions, producing nested sub-partitions following the priority order of objectives. This poses a challenge when dealing with objectives having continuous domains, as we are not considering single values but bins with fixed ranges (e.g., like in a histogram). To solve the problem, all real-coded objectives (i.e., $f_{1-2}$ and $f_{3.2}$) are associated\footnote{We used the verb \textit{associated} rather than \textit{replaced} because we want to keep the original real values in case a tie occurs over the objective with the lowest priority.} with the lower-bound value of their bin. Integer objectives can be used as they are (i.e., $f_{3.1} \in [1,n\cdot |t|]$ and $f_{3.3} \in [0,100]$ both in $\mathbb{N}^+$). In this way, the partitioning operation can be solved with a simple sorting algorithm with ties; when a tie occurs over the objective with the lowest priority, individuals are sorted based on their original real-coded values of $f_{1-2}$ and $f_{3.2}$, which should be stored rather than replaced by their integer bin value.
Lastly, we can update the fitness value by assigning a rank to each individual in the population that linearly increases with their order in the partitions (i.e., progressively enumerating each individual after partitioning). 

The procedure is detailed in Algorithm~\ref{alg:rank_partitioning}, which runs in $\mathcal{O}(N \texttt{log} N)$ where $N$ is the number of individuals in the population. 
It takes as input a matrix $N \times 9$ containing, for each individual, 
(i) its kinematic fitness $f_{1-2}$, 
(ii) the four sub-objectives describing the components $f_3$,
(iii) the original values of the continuous objectives that were evaluated by their lower-bound bin value,
(iv) the updated rank value that will be calculated by the procedure and used by the selection operators for the reproduction and survival of individuals, 
and (v) a reference to the individual in the population\footnote{The reference is used to link fitness in the matrix and the corresponding individual in the population, as only the fitness matrix will be sorted to save computational time.}.
Lines 2-4 transform real-coded objectives into an integer value (depicted in the algorithm with the subscripts $\mathbb{R}$ and $\mathbb{N}$, respectively) based on the specified size of their bin. Line 5 sorts each row of the fitness matrix based on the order of the five integer objectives. Lines 6-9 evaluate the boundaries within partitions by calculating differences between each sorted column of the fitness matrix: if two consecutive elements of the same column have the same value, they belong to the same partition. By summing the differences to each row (lines 12-13), the algorithm identifies the boundaries within partitions. Lines 11-17 sort each row of the fitness matrix based on the original values of real-coded objectives, but only within partitions to avoid disrupting the order computed at line 5. Finally, lines 18-20 assign the updated rank to each individual.

In terms of multi-objective optimization, this operation can be seen as a restriction in the objective space, where we only analyze the section of the Pareto optimal front having minimum values for one of the objectives: the first partition corresponds to the optimal solution for $f_{1-2}$, so we can select the individual that minimizes $f_3$ as the most fitting. The advantages of such a method are: (i) enforcing priority on a specific objective (e.g., the inverse kinematic in our case), (ii) removing the burden of choosing weight values as in the common preference-based methods, and (iii) allowing future studies to easily include further sub-objectives that were not considered in this work but that might be relevant for the problem.

\subsection{Constraints}
\label{sec:constraints}

The configurations represented in the solutions are subjected to some kinematics constraints in addition to the bounds on the decision variables (i.e., $\Delta_{\theta}$ and $\Delta_l$, which are user parameters defined before manufacturing).

\begin{figure}[h]
    \centering
    \subfigure[\protect\url{}\label{fig:constraint_12}]%
    {\includegraphics[width=5.5cm]{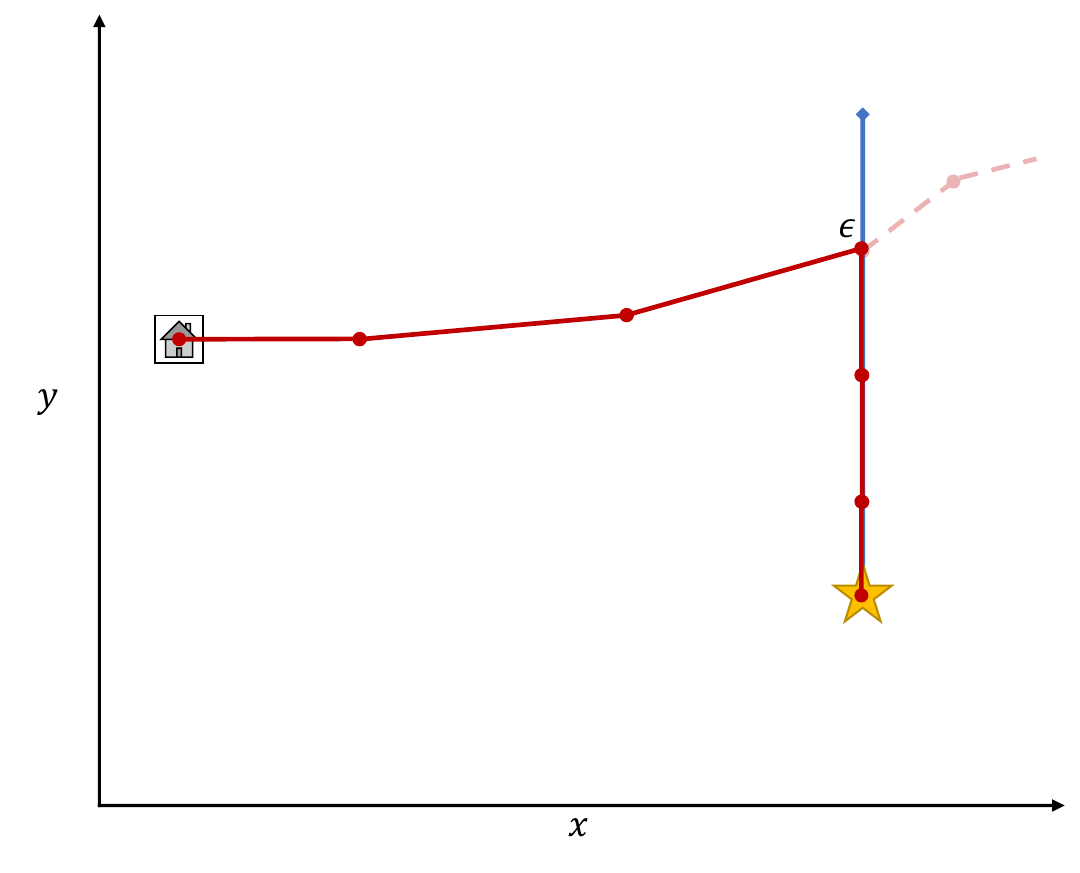}}
    \subfigure[\protect\url{}\label{fig:constraint_3}]%
    {\includegraphics[width=5.5cm]{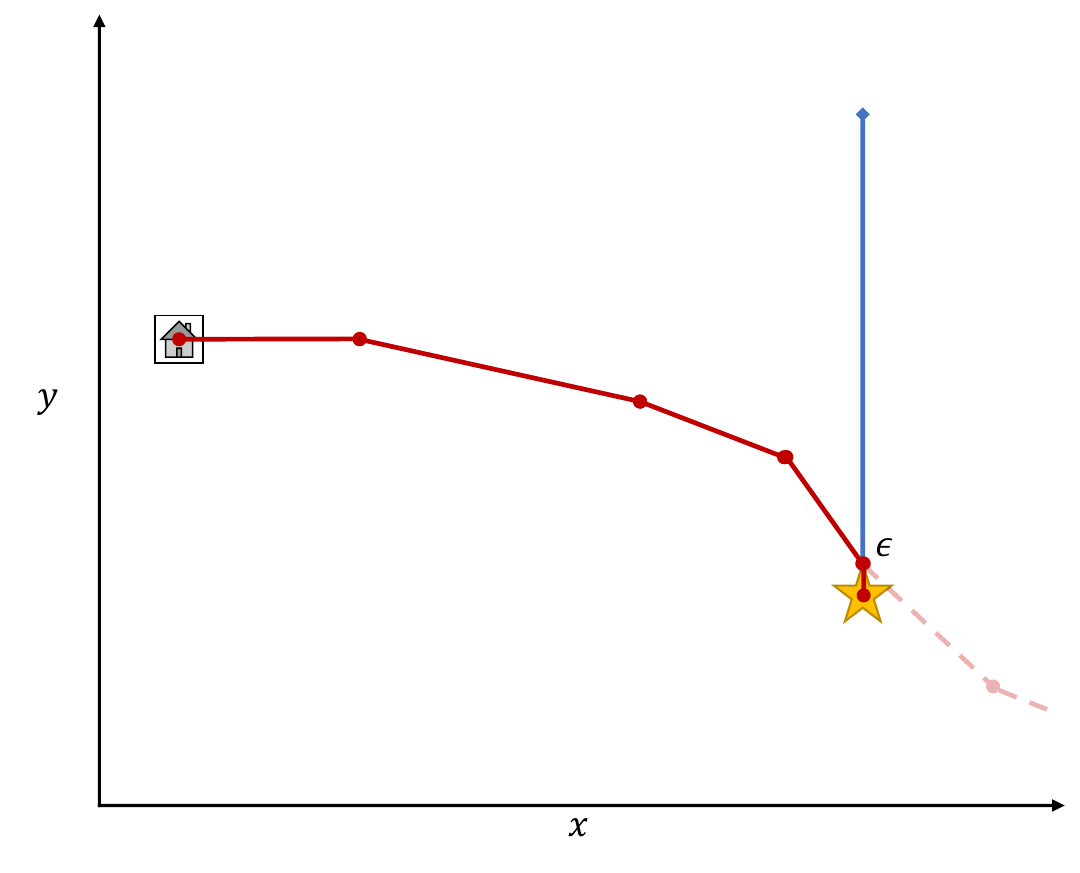}}
    \subfigure[\protect\url{}\label{fig:constraint_4}]%
    {\includegraphics[width=5.5cm]{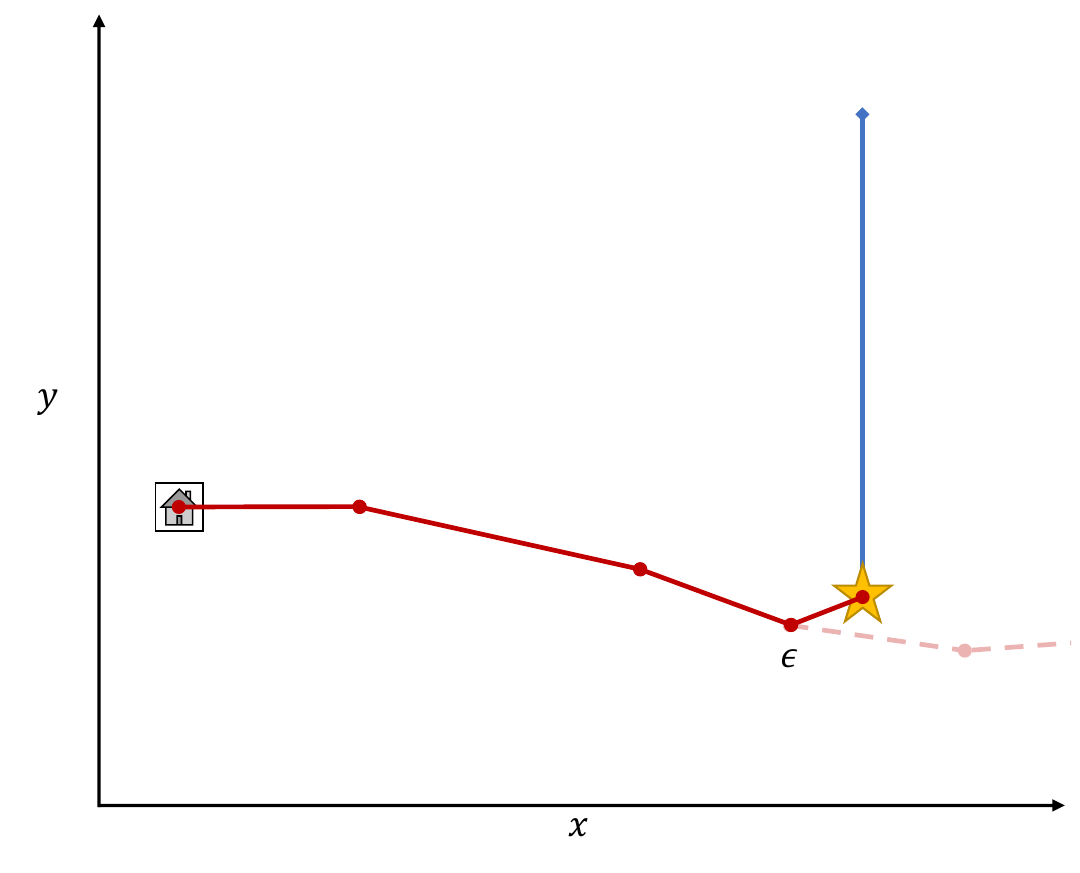}}
    \subfigure[\protect\url{}\label{fig:constraint_5}]%
    {\includegraphics[width=5.5cm]{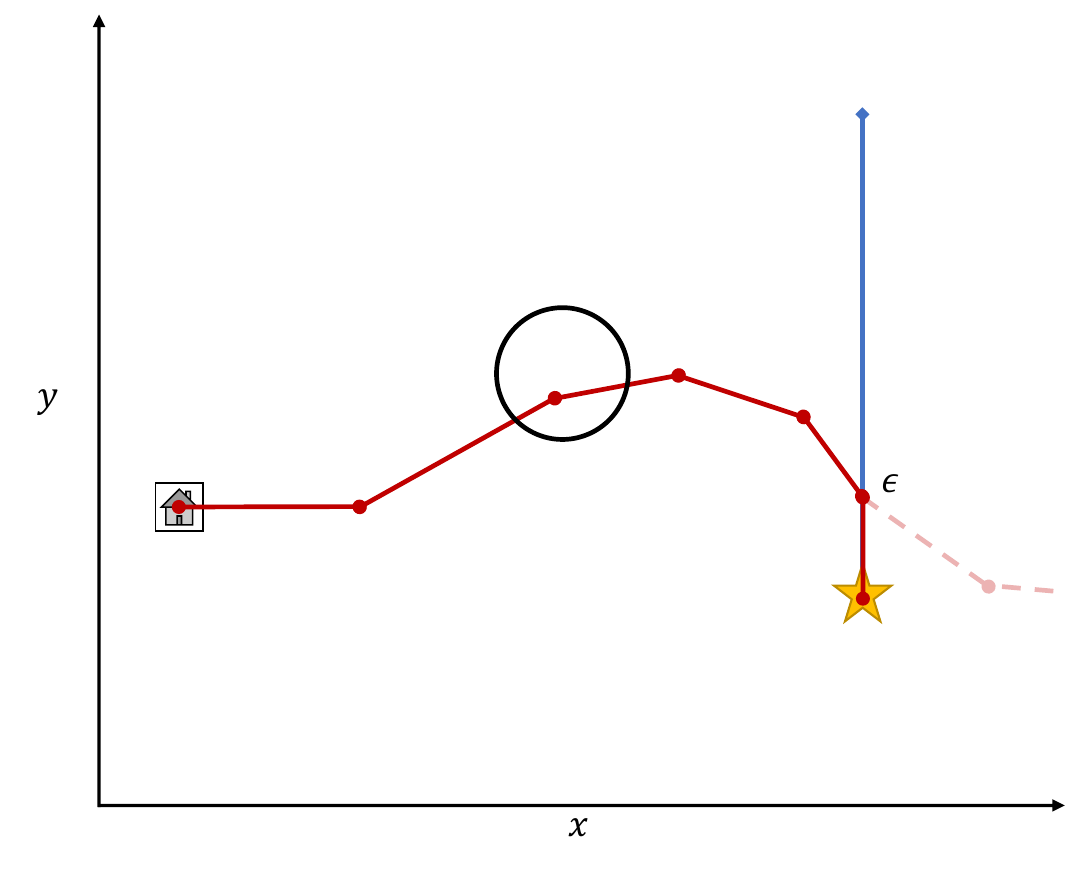}}
    \caption{Some examples of infeasible solutions. In (a), a robot that steers excessively to reach the target; in (b), a robot that has a very small link at the end right after steering that cannot accommodate a gripper device; in (c), a robot that does not reach the target with the desired orientation; and in (d), a robot that intersects with an obstacle. }
    \label{fig:constraints_34}
\end{figure}

Since the last links (i.e., from the $\epsilon$-th to the $\bar{n}$-th node) are automatically defined to reach the target as discussed in Sec.~\ref{sec:reaching_orientation}, the orientation angle $\theta_{\epsilon}$ between the last employed link and the target's orientation segment depends on the kinematics chained produced by the genetic operators and therefore might be generated out of bounds $\Delta_{\theta}$, as shown in Fig.~\ref{fig:constraint_12}. We defined two inequality constraints as in (\ref{eq:constraint_12}) to consider infeasible those solutions having configurations with $\theta_{\epsilon}$ exceeding these limits.

\begin{equation}\label{eq:constraint_12}
    \begin{aligned}
        \theta_{\epsilon} \geq {\Delta_\theta}^{(L)} \wedge \theta_{\epsilon} \leq {\Delta_\theta}^{(U)}
    \end{aligned}
\end{equation}

Similarly, the length of each configuration depends on how far the $i$-th target is from the $\bar{n}_i$-th node, which might generate configurations having the length of the last link $l_{\bar{n}}$ shorter than the limit on its lower bound. In practice, this is a problem only if the last link is required to steer to be aligned to the target's orientation segment: assuming that these manipulators are equipped with some sort of gripping mechanism at their tip, we must take into account the space required to accommodate the gripper on the target's orientation segment for correct grasping. We formulated the bounds ${\Delta_l}^{(L)}$ based on this constraint; therefore, we consider as infeasible those solutions having configurations with a single link on the target's orientation segment that is not sufficiently long, as shown in Fig.~\ref{fig:constraint_3}. This defines a third inequality constraint as in (\ref{eq:constraint_3}).

\begin{equation}\label{eq:constraint_3}
    \begin{aligned}
        \text{if} \; f_{3.1b} = 1 \text{,} \quad  l_{\bar{n}} \geq {\Delta_l}^{(L)}
    \end{aligned}
\end{equation}

The node $\epsilon$ is defined as the closest joint to the target's orientation segment and is evaluated with the point-segment distance~\cite{stroppa2018convex} described in Algorithm~\ref{alg:pointSegmentDistance}. 
Since this procedure can indicate one of the edges of the segment as the closest point to $\epsilon$ -- rather than its projection on the segment -- in case that edge turned out to be the target itself, the solution would look like the one shown in Fig.~\ref{fig:constraint_4}.
In particular, we refer to the case at line 16 of Algorithm~\ref{alg:pointSegmentDistance}. To prevent such an inconvenience, we defined a further inequality constraint as in (\ref{eq:constraint_4}): if the difference in orientation of the end effector and the desired orientation to reach the $j$-th target (indicated as $t^{(3)}_j$) is greater than a certain threshold (e.g., $\pi/18$ radians or 10 deg), then the solution is infeasible. Note that the orientation of the end effector is calculated as the summation of the angles at all joints until the node $\epsilon$, as after that the robot simply goes on a straight line.

\begin{equation}\label{eq:constraint_4}
    \begin{aligned}
        \bigg | \left( \sum_{j=1}^{\epsilon_i - 1}{\theta_j} \right) + \theta_{\epsilon_i} - t^{(3)}_j \bigg | \geq \frac{\pi}{18}
    \end{aligned}
\end{equation}

Lastly, we have a constraint on obstacle collision: if the environment is composed of elements that are not targets, then they have to be considered obstacles. Solutions colliding with obstacles are infeasible, as shown in Fig.~\ref{fig:constraint_5}. In our optimization problem, we consider obstacle collision an equality constraint, expressed as in (\ref{eq:constraint_5}).

\begin{equation}\label{eq:constraint_5}
    \begin{aligned}
        \texttt{intersections}(\psi,o) = 0
    \end{aligned}
\end{equation}

The function in (\ref{eq:constraint_5}) returns the number of intersections between each obstacle in the set $o$ and every link of each configuration in $\psi$. This procedure runs in $\mathcal{O}(|t| \cdot n \cdot |o|)$. For the sake of simplicity and to reduce computational complexity, we represent every obstacle as a circle \cite{dong2015obstacle}. During problem definition, the users can choose a radius slightly larger than the one of the circle that would circumscribe the obstacle to avoid any contact between the robot and the obstacle -- as shown later in Fig.~\ref{fig:optimal_solutions}. 

\begin{figure}[ht]
    \centering
    \subfigure[\protect\url{}\label{fig:obstacleAvoidance_graph}]%
    {\includegraphics[height=5.7cm]{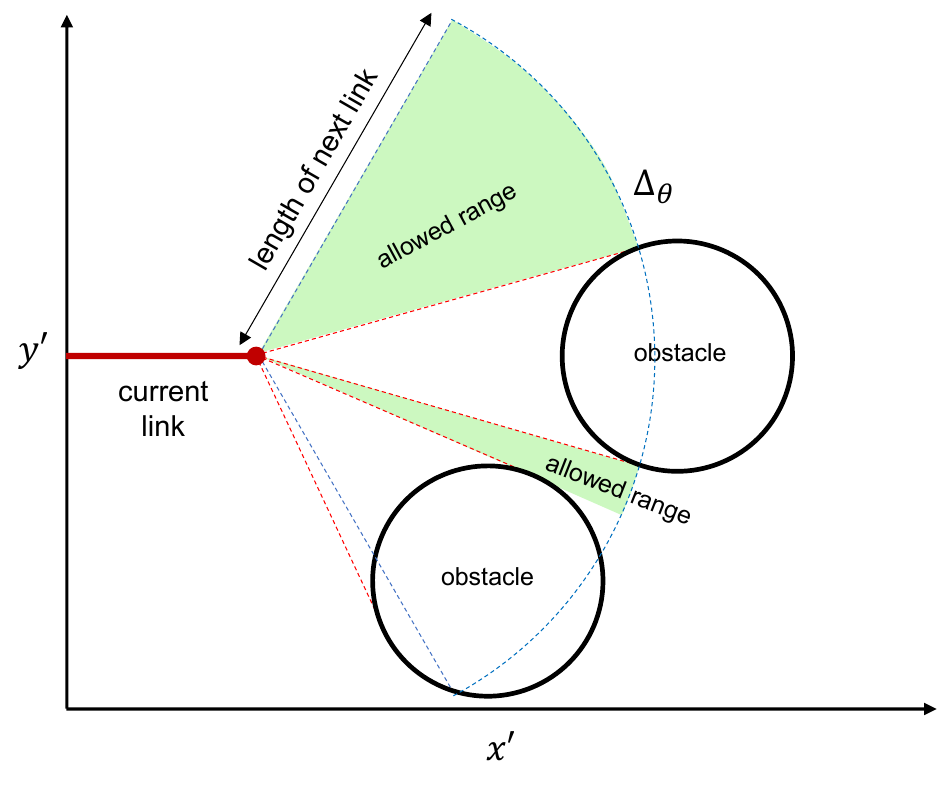}}
    \subfigure[\protect\url{}\label{fig:obstacleAvoidance_angles}]%
    {\includegraphics[height=5.7cm]{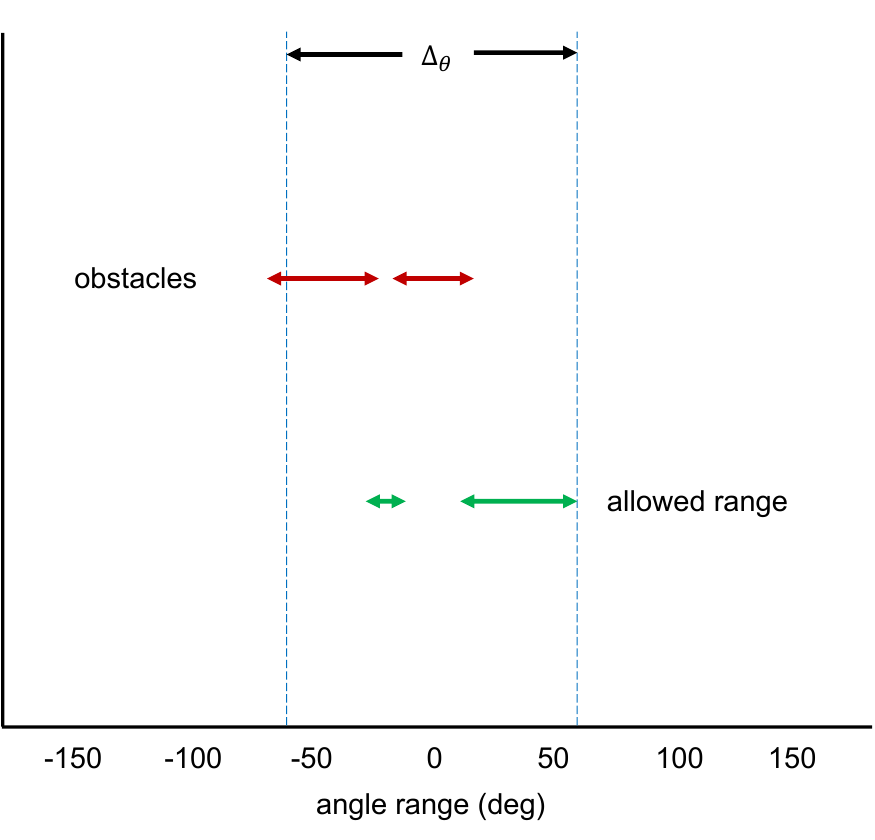}}
    \caption{Example of obstacle avoidance during the generation of a configuration. The steering angle for the next link will be generated based on the current pose of the robot’s end effector and its proximity to obstacles. Tangent lines between obstacles and the current end effector are used to estimate angle ranges that should be avoided, which will be subtracted from the fixed angle bounds $\Delta_\theta$ such that only non-intersecting links are generated. The plot in (a) shows the allowed range for angle generation, whereas the plot in (b) defines the allowed range as the difference between $\Delta_\theta$ and the obstacles’ range.}
    \label{fig:obstacleAvoidance}
\end{figure}

\subsubsection{Obstacle avoidance}
\label{sec:obstacle_avoidance}

In addition to the obstacle-collision constraint described in (\ref{eq:constraint_5}), we can speed up the optimizer convergence by generating solutions that avoid obstacles during the random initialization of the population. 
For each node of each configuration in a solution $\psi$, the random value of steering angle $\theta$ will be generated such that the link is not oriented towards a close obstacle. As shown in Fig.~\ref{fig:obstacleAvoidance}, by retrieving the lines starting from the node and tangent to any obstacle in the range, we can define portions of the angle bounds $\Delta_{\theta}$ that will generate a collision and remove them from $\Delta_{\theta}$ for that node. Part of $\Delta_{\theta}$ will be removed for some nodes, decreasing the size of the search space $\omega$ -- defined in (\ref{eq:search_space}) -- and preventing the generation of infeasible solutions.
The procedure is described in Algorithm~\ref{alg:obstacle_avoidance}, and it is executed for each node of a configuration $\zeta$ to retrieve the orientation of the two tangent lines to each nearby obstacle -- therefore, only to the obstacles who are in a range within the length of the next link from the node, as shown in Fig.~\ref{fig:obstacleAvoidance_graph}. The orientation is calculated geometrically at line 7 from the points on the tangent lines intersecting the obstacles; these points are calculated at line 6 by the intersection of two circles: (i) the obstacle and (ii) the circle with origin at the node and radius equal to the $\ell^2$ norm between the node and the origin of the obstacle (line 5). Note that to retrieve angles with respect to the current link's orientation, the reference system must be transformed to align the link with the horizontal axis (lines 2 and 4 through Rodriguez Rotation \cite{cheng1989historical}). The allowed range, summarized in Fig.~\ref{fig:obstacleAvoidance_angles}, is composed of the complement of the bound $\Delta_{\theta}$ and the collision range with each obstacle (line 8). 
The procedure runs linearithmically with the number of nearby obstacles, as the complement operation requires the sets to be sorted\footnote{We assume that the MATLAB function \texttt{circcirc}, which analytically solves a system of equations, runs in constant time.}.

\IncMargin{1em}
\begin{algorithm}
    \SetKwData{Node}{node}
	\SetKwData{P}{p}
	\SetKwData{U}{$\vec{u}$}
	\SetKwData{O}{o}
	\SetKwData{B}{$\Delta$}
	\SetKwData{R}{R}
	\SetKwData{Dist}{d}
	\SetKwFunction{Rod}{rodriguezRotation}
	\SetKwFunction{Circcirc}{circcirc}
	\SetKwFunction{Atan}{atan2}
	\SetKwInOut{Input}{input}
	\SetKwInOut{Output}{output}
	
	\Input{$\Node = \langle x,y \rangle$, coordinates of the current node.}
	\Input{$\U$, unit vector of the current link's orientation. }
	\Input{$\O$, set of nearby obstacles, each of them with coordinates and radius $\langle x,y,r \rangle$.}
	\Input{$\B_{\theta}$, angle bounds.}
	\Output{The updated $\B_{\theta}$, angle bounds without collision ranges.}
	\BlankLine		
	
	\Begin{	

		\R $\leftarrow$ \Rod{$\U$,$\langle 1,0 \rangle$}\;
        \For{$i \in [1, |\O|]$}
        {
            $\O^{(x,y)}_i \leftarrow (\R \cdot (\O^{(x,y)}_i - \Node)')'$\;
            $\Dist \leftarrow \norm{\langle 1,0 \rangle-\O^{(x,y)}_i}^2$ \;
            $[\P_1, \P_2] \leftarrow$ \Circcirc{$0,0,\Dist,\O^{(x)}_i,\O^{(y)}_i,\O^{(r)}_i$}\;
            $\B^{\O_i}_{\theta} \leftarrow [$\Atan{$\P^{(y)}_1,\P^{(x)}_1$}$,$\Atan{$\P^{(y)}_2,\P^{(x)}_2$}]\;
            $\B_{\theta} \leftarrow \B_{\theta} \setminus \B^{\O_i}_{\theta}$\;
            
        }
        
		\KwRet{$\B_{\theta}$};
	}
	
	\caption{Angle Range with Obstacle Avoidance}\label{alg:obstacle_avoidance}
\end{algorithm}\DecMargin{1em}

The obstacle avoidance method is used in both random initialization and variation operations: the blx-$\alpha$ operator \cite{eshelman1993real} offers the best opportunity to include obstacle avoidance while performing crossover because it generates random values of the decision variables -- angles $\theta$, in our case -- with uniform probability within a specific range; this range can easily be reduced by removing angles that would generate a collision. Similarly, with random mutation \cite{michalewicz1992genetic} we can exploit the obstacle avoidance method by simply re-generating a random individual.

\begin{figure}[ht]
	\centering
	{\includegraphics[width=8cm]{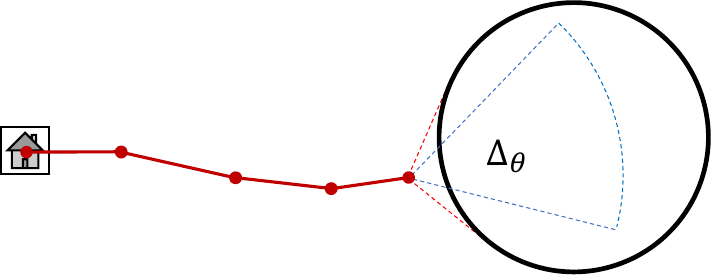}}
	\caption{While randomly generating the angle for the current joint within the bounds $\Delta_\theta$, if the joint is located in the very proximity of an obstacle, the next link will always be generated within the obstacle causing a collision.}
	\label{fig:wrongGeneration}
\end{figure}

It is still required to keep the constraint specified in (\ref{eq:constraint_5}) to check whether any obstacle collision occurred: when a new individual is generated (either through random initialization or variation), one of its nodes can be placed near an obstacle. When that happens, the range of angles for the next link might not be sufficiently wide to avoid a collision (Fig.~\ref{fig:wrongGeneration}). This issue might be avoided by exploiting deterministic search algorithms with backtracking (such as Depth First Search), which would discard this node and go back to re-generate the previous node until a free range is found. However, due to the stochastic nature of the whole procedure, the generation of the following node will not follow a specific order as required by deterministic search algorithms, leading to a random repetition of the node generation until a free range is found. The time complexity of such a method would be high with a deterministic procedure\footnote{With Depth First Search, the complexity would be $\mathcal{O}(b^n)$, where $b = \Delta_\theta$ is the branching factor of the search tree represented by all the angles in the range of the angle bounds -- which are real values -- and $n$ is the number of nodes of the individual.}, whereas it becomes hard to predict with stochastic methods. Therefore, instead of regenerating the previous node when collisions cannot be avoided, we simply allow collisions to happen and then penalize them afterward.

\subsection{Optimization method}
\label{sec:genetic_algorithm}

Due to the nonlinearity, multi-modality, and complexity of the problem, the advantages of EC techniques over other numerical optimization methods made them the perfect candidates for our case. In particular, we used a real-coded single-objective genetic algorithm implementing the rank-partitioning method described in Sec.~\ref{sec:separation_partitioning} to solve a MOOP. The pseudo-code of the procedure is illustrated in Algorithm~\ref{alg:genetic}: a set of $N$ individuals is randomly generated and evaluated based on their fitness, pairs of individuals are selected to create a mating pool and combined with variation operators (crossover and mutation) to produce a set of offspring, and finally, the best individuals among the old population and the offspring are selected to populate the next generation of individuals (survival stage). 

\IncMargin{1em}
\begin{algorithm}
        \SetKwData{P}{pop}
        \SetKwData{X}{mat}
        \SetKwData{Q}{off}
        \SetKwData{G}{G}
        \SetKwData{F}{f}
        \SetKwData{N}{N}
	\SetKwFunction{Init}{randomInitialization}
        \SetKwFunction{Eval}{evaluation}
        \SetKwFunction{Rank}{rankPartitioning}
        \SetKwFunction{Sel}{selection}
        \SetKwFunction{Var}{variation}
        \SetKwFunction{Sur}{survival}
	\SetKwInOut{Input}{input}
	\SetKwInOut{Output}{output}
	\Input{The optimization problem described by $t$, $h$, $o$, $n$, $\Delta_{\theta}$, $\Delta_{l}$}
        \Input{The GA parameter such as number of individuals $\N$ and number of generations $\G$}
	\Output{Either the entire population $\P$ or just the most fitting individual $\P(1)$}
	\BlankLine		
	
	\Begin{	

    	$\P \leftarrow \Init(\N)$\;
            $\P \leftarrow \Eval(\P)$\;
            \For{$i \in [1, \G]$}
            {
                $\X \leftarrow \Sel(\P)$\;
                $\Q \leftarrow \Var(\X)$\;
                $\Q \leftarrow \Eval(\Q)$\;
                $\P \leftarrow \Sur(\P,\Q)$\;
                
            }
        
		\KwRet{$\P$};
	}
	
	\caption{Genetic Algorithm}\label{alg:genetic}
\end{algorithm}\DecMargin{1em}

The algorithm implements standard genetic operators such as binary tournament selection \cite{goldberg1991comparative}, blend crossover (blx-$\alpha$) for adaptive search \cite{eshelman1993real}, 
random mutation for exploration \cite{michalewicz1992genetic}, 
and an elitist survival method ($\mu+\lambda$ scheme \cite{beyer2002evolution}). 

The evaluation of individuals performs the following operations:

\begin{enumerate}
    \item calculates the fitness of each individual $\psi$ based on the methods described in Sec.~\ref{sec:fitness}, which also modifies chromosomes by updating the values of extra genes for each configuration (i.e., $\epsilon$, $\theta_{\epsilon}$, $\bar{n}$, and $l_{\bar{n}}$);
    \item evaluates the constraints described in Sec.~\ref{sec:constraints} with static penalty method \cite{coello2002theoretical} and updates the value of fitness by degrading it (i.e., increasing its value when a violation occurs) -- note that the penalty is added only on $f_{1-2}$, not on $f_3$; and
    \item ranks each individual by using the rank-partitioning procedure described in Algorithm~\ref{alg:rank_partitioning}. 
\end{enumerate}
Since the rank assigned by the partitioning method is the final score that defines the multi-objective optimality of an individual within a population, the binary tournament operator will perform selection over this value rather than on the fitness $f_{1-2}$. 

The rank-partitioning procedure is also used during survival operations to select which individuals will be carried on in the next generation. Since the ranking is performed with respect to the individuals composing the population, the partitioning must be re-performed when combining the two populations of parents and offspring ($\mu+\lambda$ schema). This inevitably burdens the complexity of the algorithm but ensures the preservation of elites. 

Finally, as mentioned in Sec.~\ref{sec:obstacle_avoidance}, the procedure for obstacle avoidance described in Algorithm~\ref{alg:obstacle_avoidance} is used any time a new individual is generated: this includes the random initialization at line 2 and the variation operations at line 6. In particular, we included it in the blx-$\alpha$ crossover and in the random mutation operator by removing collision ranges from the uniformly distributed domains while generating joint angles. 

\section{Experimental Results and Discussion}
\label{sec:experiments}

This section illustrates the experiments we ran with the optimizer to assess its optimality (Sec.~\ref{sec:exp_optimality}) and to compare it with our previous method (Sec.~\ref{sec:comparing_methods}). Lastly, we will discuss the impact of problem dimensionality in Sec.~\ref{sec:dimensionality}.

\subsection{Optimality}
\label{sec:exp_optimality}

\begin{figure}[t!]
    \centering
    \subfigure[\protect\url{}\label{fig:best_sol_t1}Task 1]%
    {\includegraphics[height=5.5cm]{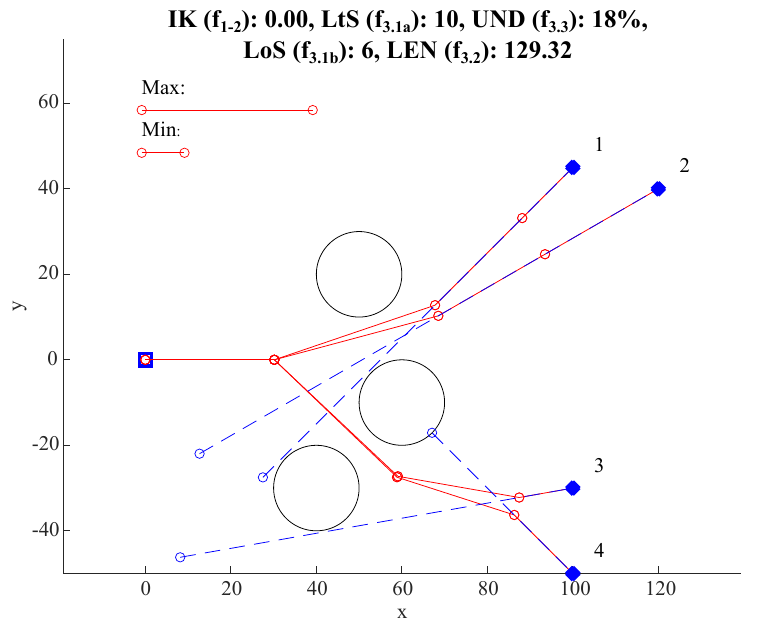}}
    \subfigure[\protect\url{}\label{fig:best_sol_t2}Task 2]%
    {\includegraphics[height=5.5cm]{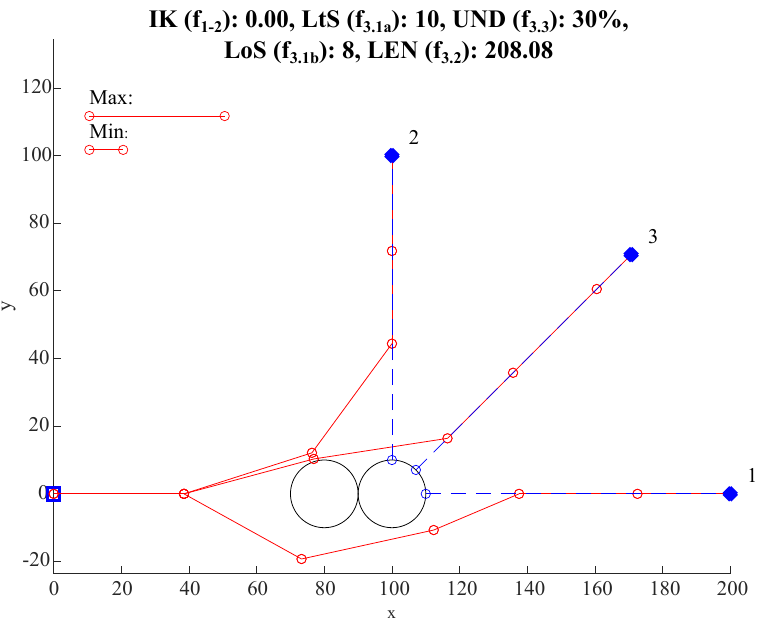}}
    \subfigure[\protect\url{}\label{fig:best_sol_t3}Task 3]%
    {\includegraphics[height=5.5cm]{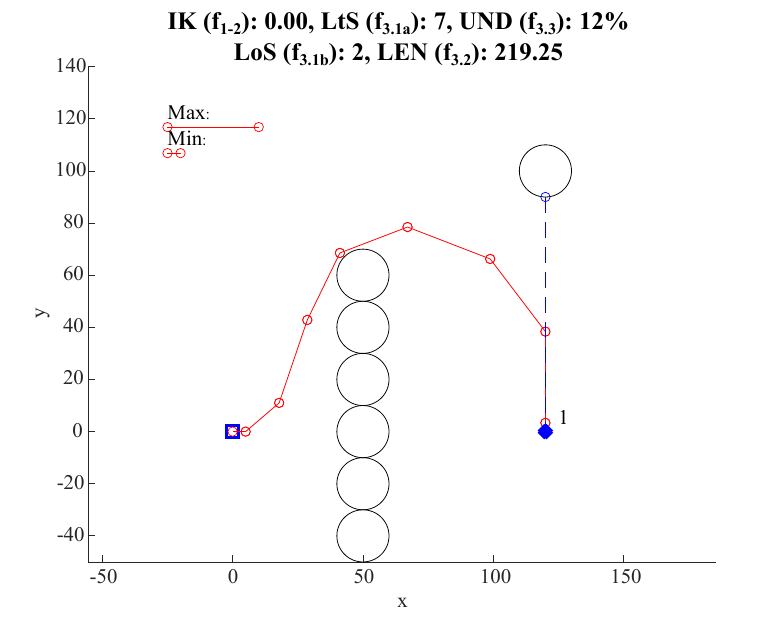}}
    \subfigure[\protect\url{}\label{fig:best_sol_t4}Task 4]%
    {\includegraphics[height=5.5cm]{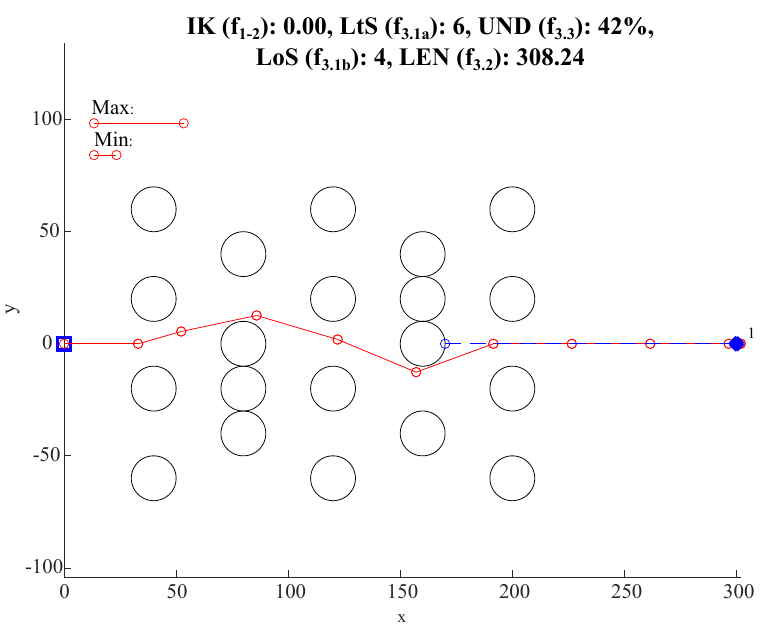}}
    \caption{Optimal solutions retrieved by the optimizer for each task, reporting the values of all the objectives as well as a graphical representation of the bounds on link lengths ($\Delta_l$). Note that configurations 3(b), 1(c), and 1(d) are passing very close to the obstacles without touching them (they are feasible solutions).}
    \label{fig:optimal_solutions}
\end{figure}

We ran a series of tests to check the optimality\footnote{Within the limitations of non-exact methods such as evolutionary algorithms in a complex task such as soft-robot design.} of the algorithm. Picking a task that can be optimally solved within the bounds given to the robot is challenging, as it is hard to estimate whether or not the robot can actually reach all targets in specific conditions -- and, in the end, that is the main motivation behind this work. Therefore, We chose tasks that were somewhat non-trivial but for which I could estimate an optimal solution. We selected the following tasks:

\begin{itemize}
    \item \textbf{Task 1}: multiple targets and scattered obstacles as shown in Fig.~\ref{fig:best_sol_t1}, to check if the algorithm converges to a (sub)optimal solution;
    \item \textbf{Task 2}: multiple targets and a brick obstacle as shown in Fig.~\ref{fig:best_sol_t2}, to check if the algorithm converges to a (sub)optimal solution;
    \item \textbf{Task 3}: single target and a wall as shown in Fig.~\ref{fig:best_sol_t3}, to pose a challenge to the obstacle-avoidance method; and
    \item \textbf{Task 4}: single target and a maze as shown in Fig.~\ref{fig:best_sol_t4}, to simulate a real scenario.
\end{itemize}
Each task was executed with a maximum number of nodes $n=20$. Bounds on joint rotation were set to $\pi/6$, a limited value to emulate possible hardware constraints; whereas bounds on link lengths were chosen to fit the tasks properly (Fig.~\ref{fig:f3_best} graphically shows the values of $\Delta_{l}$). Each task was executed twenty times for each condition to test all the combinations of parameters discussed in the next Sec.~\ref{sec:parameters}.

\begin{table}[h!]
\centering
\caption{Experimental conditions and parameters}
\label{tab:param}
\vspace{0.2cm}
\begin{tabular}{|l|c|}
\hline
\rowcolor[HTML]{C0C0C0} 
\multicolumn{1}{|c|}{\cellcolor[HTML]{C0C0C0} \textbf{PARAMETER}} & \textbf{VALUE}                       \\ \hline
\rowcolor[HTML]{FFFFFF} 
Num. of Generations                                              & $150$                                \\
\rowcolor[HTML]{EFEFEF}  
Population Size                                                  & $500$                                \\
\rowcolor[HTML]{FFFFFF} 
Selection Type                                                   & Binary Tournament                    \\
\rowcolor[HTML]{EFEFEF}
Crossover Type                                                   & blx-$\alpha$ ($\alpha = 0.5$)        \\
\rowcolor[HTML]{FFFFFF} 
Crossover Probability                                            & $0.9$                                \\
\rowcolor[HTML]{EFEFEF} 
Mutation Type                                                    & Random                                \\
\rowcolor[HTML]{FFFFFF} 
Mutation Probability                                             & $0.4$                                 \\
\rowcolor[HTML]{EFEFEF} 
Survival Type                                                    & Elitist ($\mu+\lambda$ scheme)       \\
\rowcolor[HTML]{FFFFFF} 
Constraint Handling                                              & \makecell{Static Penalty \\ $R=[10,10,10,10,100]$} \\ \hline
\rowcolor[HTML]{EFEFEF} 
Bin size for $f_{1-2}$                            & $0.10$, $0.55$, $1.00$\\ 
\rowcolor[HTML]{FFFFFF} 
Bin size for $f_{3.2}$                      & $5.0$  \\
\rowcolor[HTML]{EFEFEF} 
Obstacle Avoidance                                               & Active, Non Active                   \\ \hline
\end{tabular}
\end{table}

\subsubsection{Conditions and parameters}
\label{sec:parameters}
The algorithm was executed with the parameters shown in Tab.~\ref{tab:param}; we fixed all those related to the optimization method to focus the analysis on the novel methods we propose: rank partitioning (Sec.~\ref{sec:separation_partitioning}) and obstacle avoidance (Sec.~\ref{sec:obstacle_avoidance}). Therefore, the test analyzed two factors: 
\begin{enumerate}
    \item the \textit{bin size} of $f_{1-2}$, chosen in a range of values in order to observe its significance in the retrieval of optimal links, undulation, and length; and
    \item the obstacle avoidance to be active or not active, to understand whether this method does, in fact, improve the search or we could simply rely on an obstacle-collision-based constraint as in (\ref{eq:constraint_5}).
\end{enumerate}
Since the overall length of the robot ($f_{3.2}$) was chosen to be the objective with the lower priority (Sec.~\ref{sec:optimizing_components}), the impact of its bin size in the rank partitioning was fixed to a value that complied to the sensitivity of the task space (i.e., a two robot whose difference in length is within the value of this bin size are considered equally optimal).

About the GA, we selected suitable values based on pilot studies executed during implementation.
The number of generations $150$ was empirically chosen based on convergence observations. The population size is related to the complexity of the problem and must be adequately large to represent a sufficient number of solutions for each building block of the problem \cite{jh1975adaptation}; however, due to the high multi-modality of these tasks, it is hard to estimate a correct value. We found that the value of 500 allowed us to retrieve good solutions in all four tasks, but it might be possible to achieve better results with larger populations. The crossover probability was empirically set at a fixed value of $0.9$ to ensure exploitation; whereas the mutation probability was empirically set to 0.4 to have a sufficient margin of exploration since we deal with a wide search space -- however, we did not observe a significant contribution of exploration during pilot executions. Constraints were handled with a static penalty (the penalty-factor vector $R$ in the table follows the order of the constraints presented in Sec.~\ref{sec:constraints}, with a higher factor for obstacle collisions); we also tried separation methods~\cite{coello2002theoretical} but we found that the rank-partitioning algorithm wasn't performing as good, so we did not include them in the analysis. 

\subsubsection{Results and discussion}
\label{sec:results}

\begin{figure}[h!]
    \centering
    \subfigure[\protect\url{}\label{fig:plot1_ik}]%
    {\includegraphics[height=4.8cm]{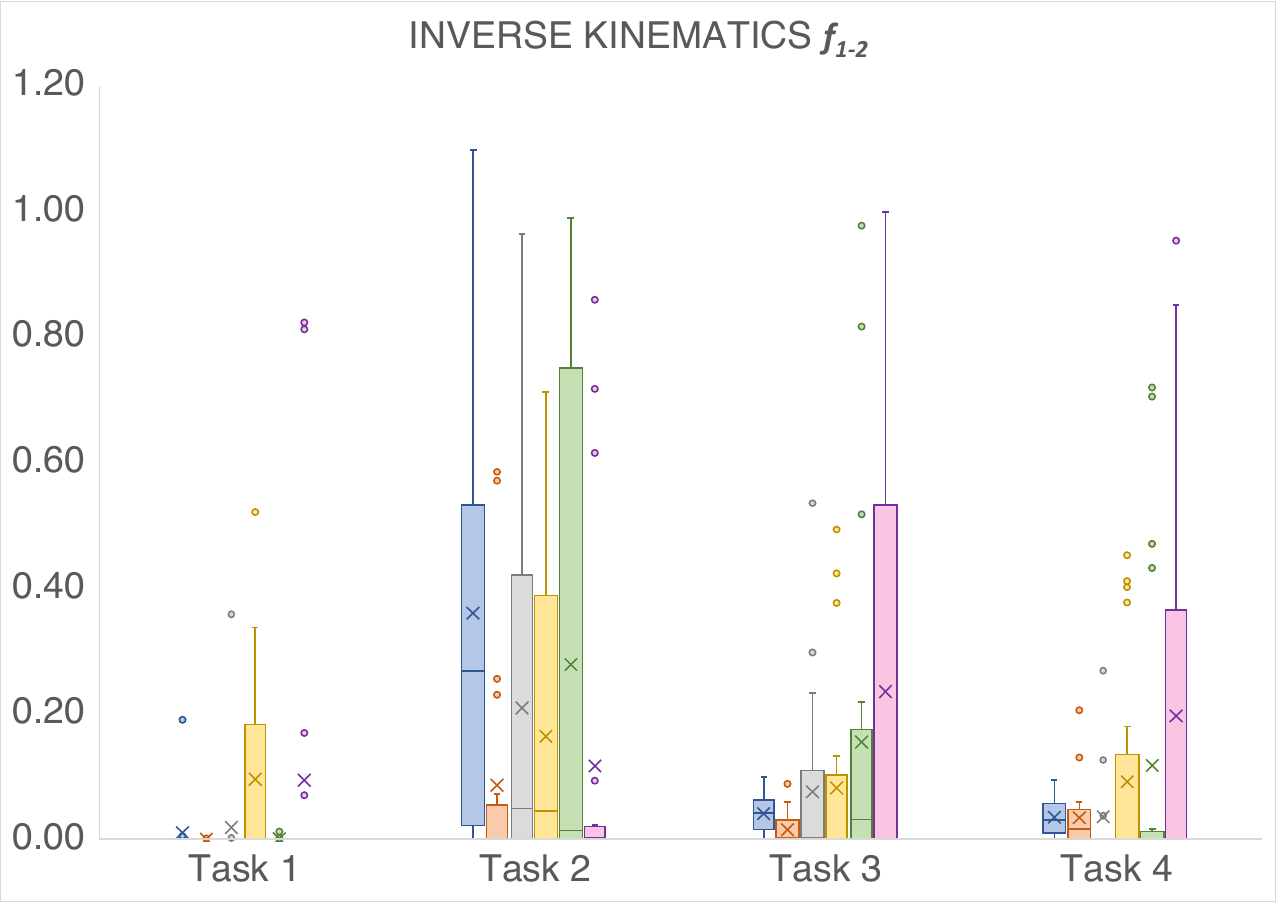}}
    \subfigure[\protect\url{}\label{fig:plot2_links2seg}]%
    {\includegraphics[height=4.8cm]{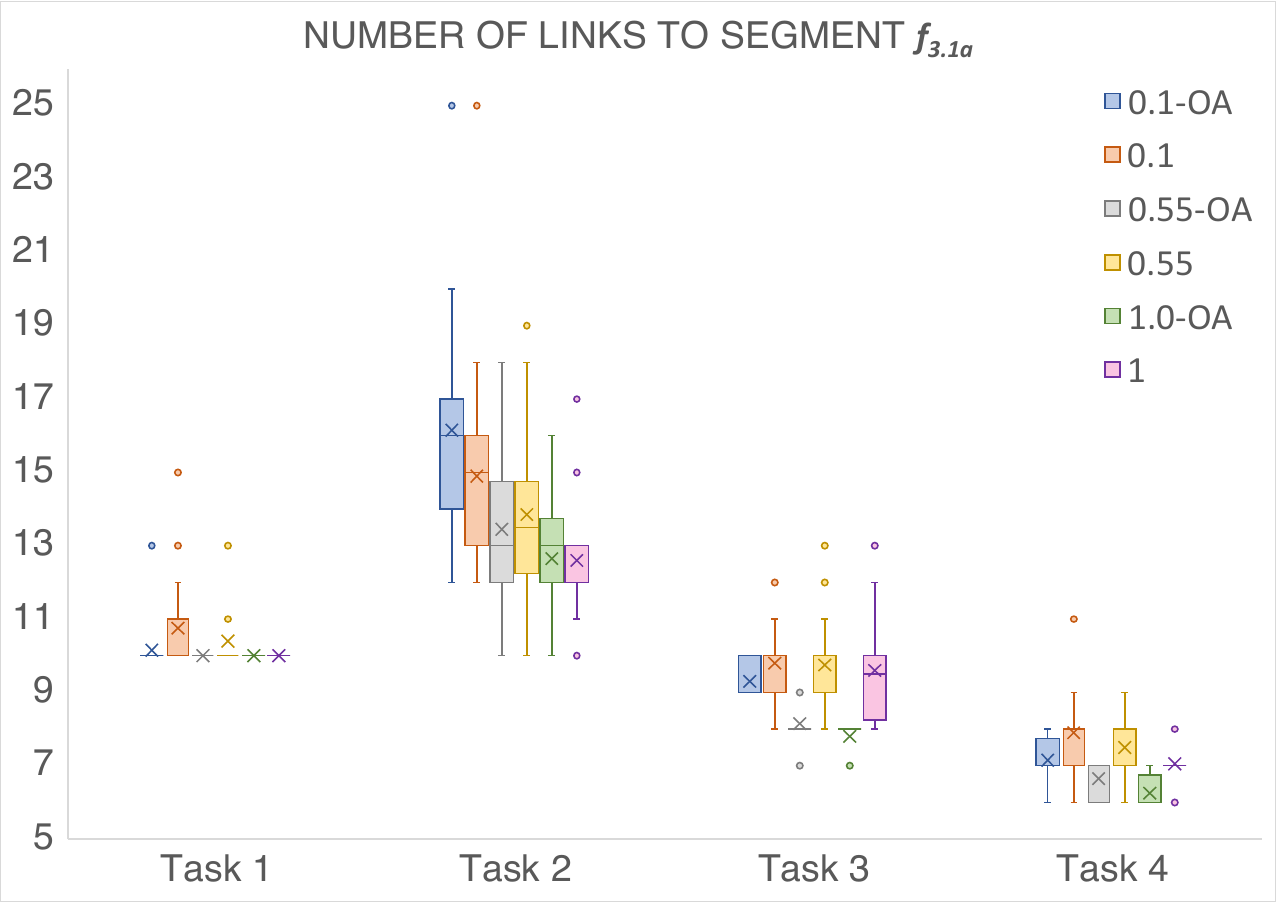}}
    \subfigure[\protect\url{}\label{fig:plot3_und}]%
    {\includegraphics[height=4.8cm]{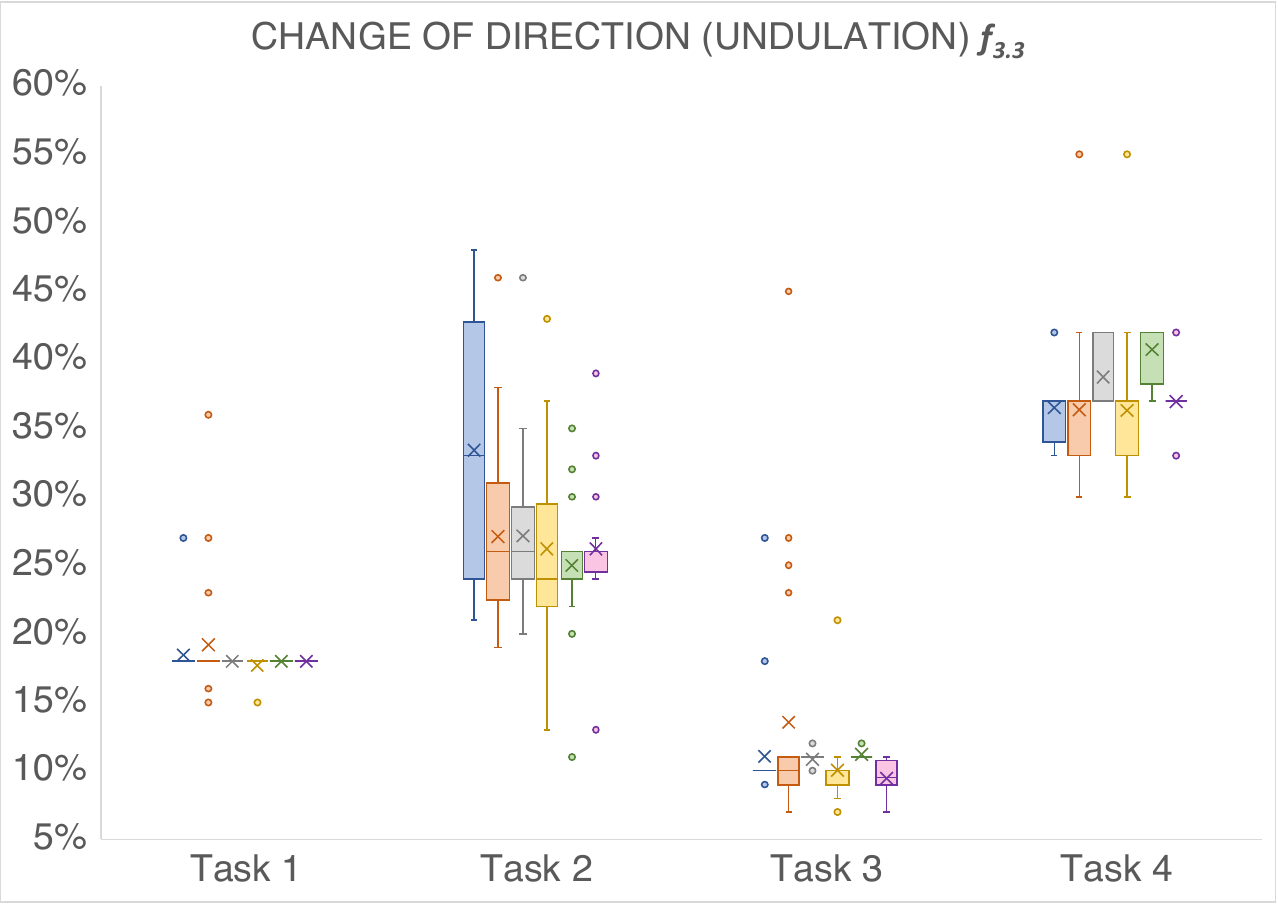}}
    \subfigure[\protect\url{}\label{fig:plot4_linksoseg}]%
    {\includegraphics[height=4.8cm]{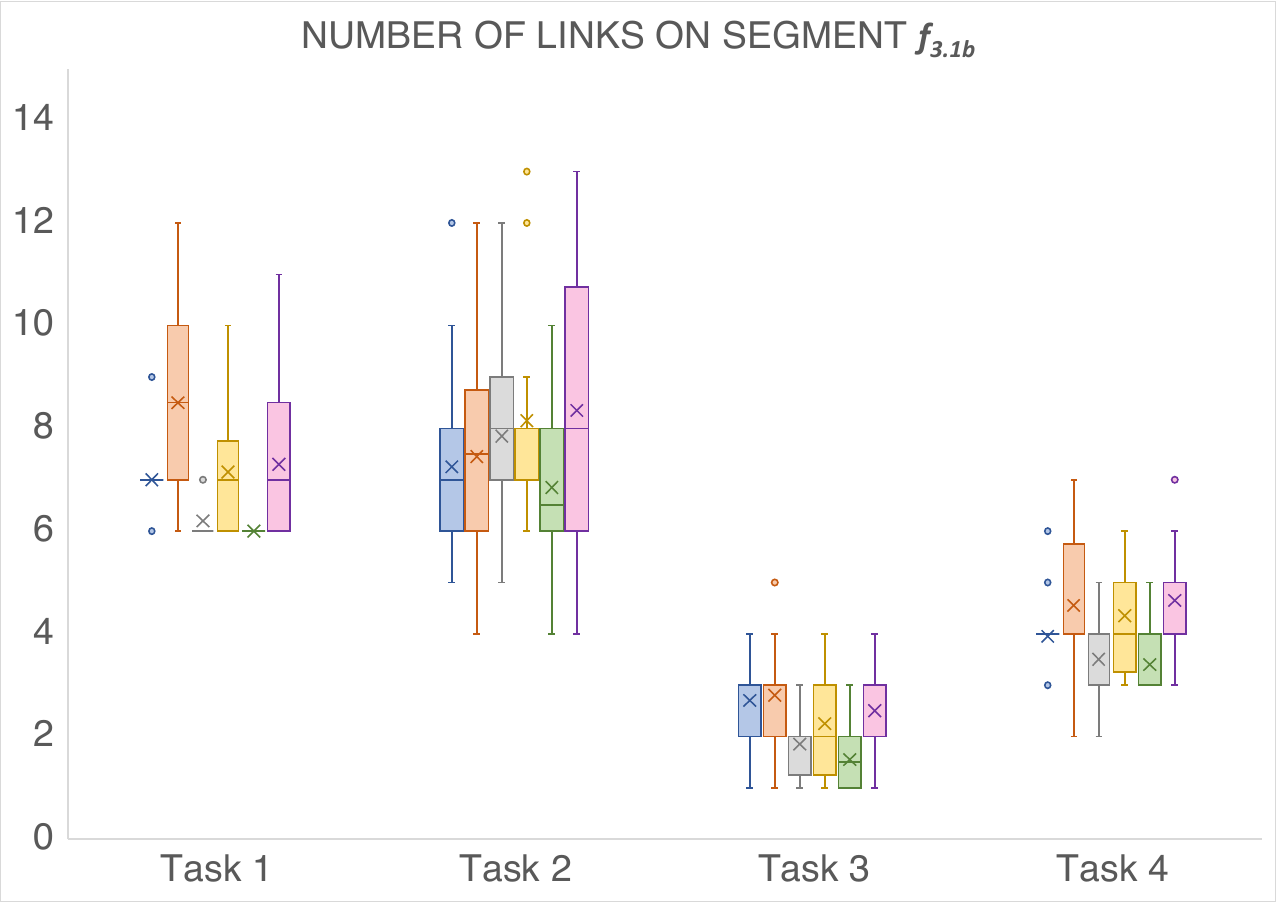}}
    \subfigure[\protect\url{}\label{fig:plot5_length}]%
    {\includegraphics[height=4.8cm]{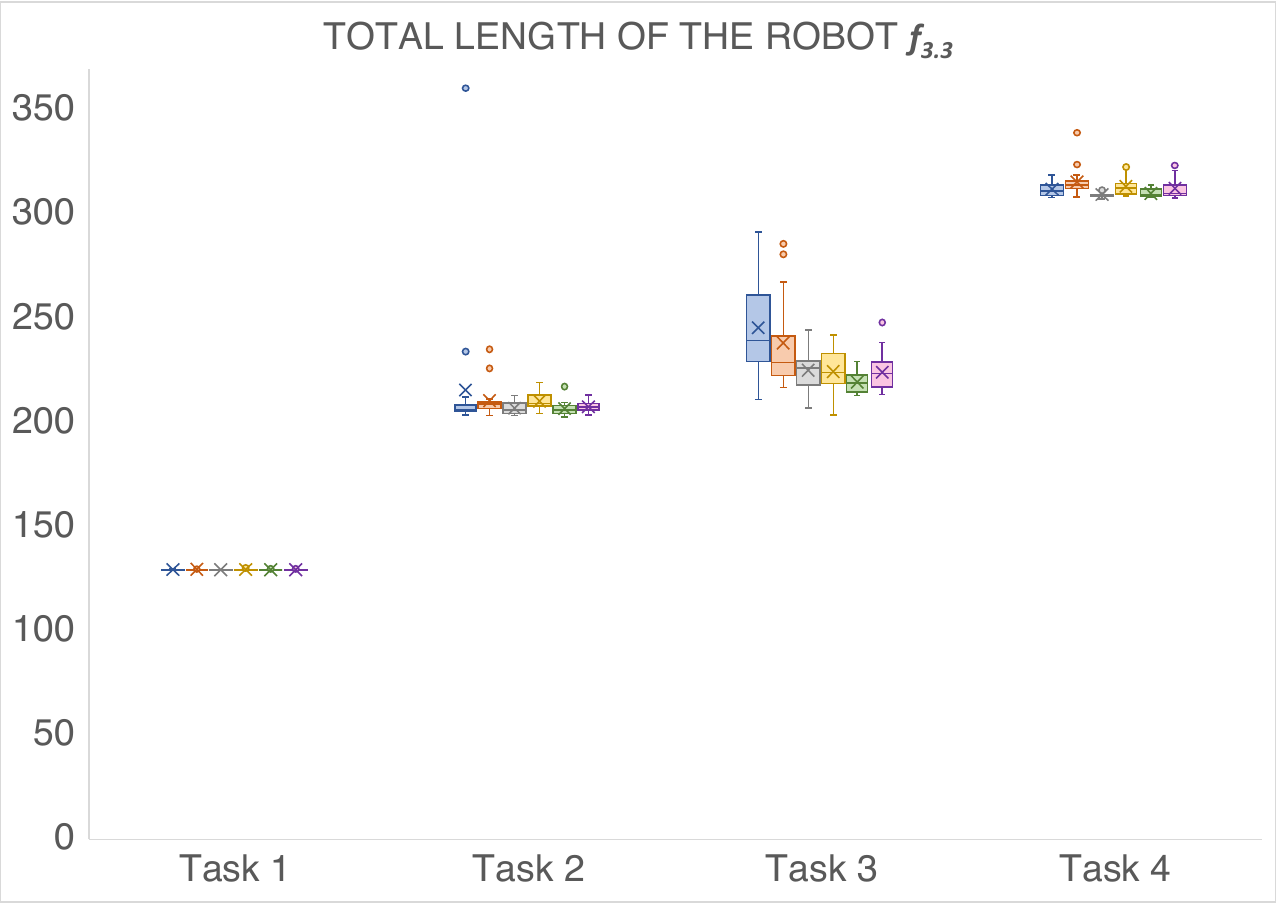}}
    \subfigure[\protect\url{}\label{fig:plot6_coll}]%
    {\includegraphics[height=4.8cm]{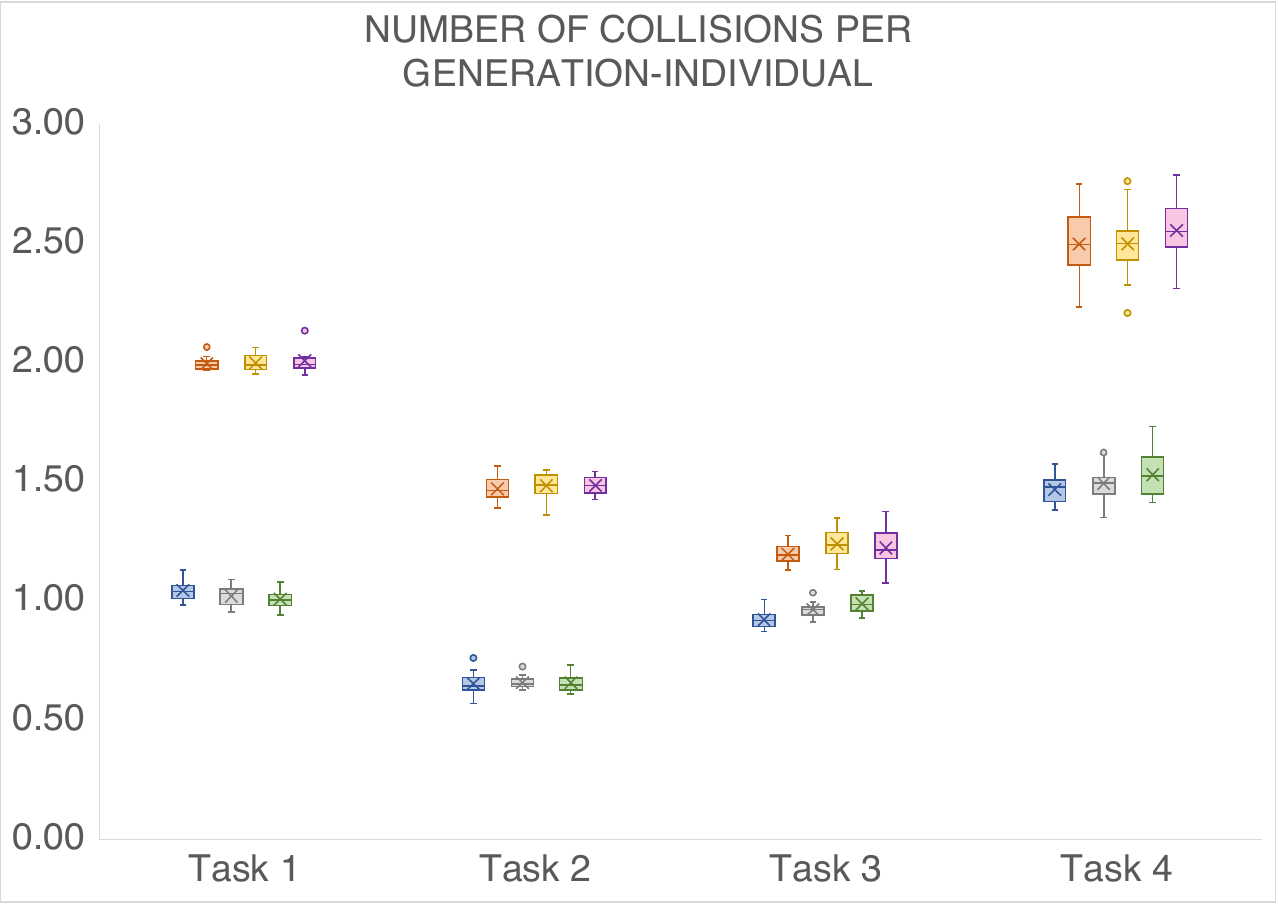}}
    \caption{Results of the optimizer over four tasks. The legend specifies six different conditions with value bin size for $f_{1-2}$ and whether the obstacle avoidance is active.}
    \label{fig:plots}
\end{figure}

Results are presented in Fig.~\ref{fig:plots}, showing average, median, interquartile, range with outliers, and max/min for each individual task. The plots include data from all conditions as specified by the legend of the figure.
In particular, we analyzed the error in solving the inverse kinematics $f_{1-2}$ (Fig.~\ref{fig:plot1_ik}\footnote{The unit of measurement depends on the workspace of the task one intends to solve. In these examples, we can assume its centimeters.}), the number of links to reach the target's orientation segment $f_{3.1a}$ (Fig.~\ref{fig:plot2_links2seg}), the change of direction of the configurations or undulation $f_{3.3}$ (Fig.~\ref{fig:plot3_und}), the number of links on the target's orientation segment $f_{3.1b}$ (Fig.~\ref{fig:plot4_linksoseg}), the overall length of the robot $f_{3.2}$ (Fig.~\ref{fig:plot5_length}), and the number of collisions with obstacles per generated individual over the number of generations (Fig.~\ref{fig:plot6_coll}).

\begin{table}[h!]
\captionsetup{justification=centering}
\caption{ANOVA Analysis Results \\ $\oplus$ indicates data that were log-transformed \\ $\otimes$ indicates data that were ART-transformed}
\label{tab:results_ANOVA}
\vspace{0.1cm}
\resizebox{\columnwidth}{!}{%
\begin{tabular}{cc|c|c|c|c|c|c|}
\cline{3-8}
\multicolumn{1}{l}{}                                   &                               & \makecell{\textbf{Inverse Kinematics} \\ $f_{1-2}(\psi)$}           & \makecell{\textbf{Links to Target} \\ $f_{3.1a}(\psi)$}   & \makecell{\textbf{Undulation} \\ $f_{3.3}(\psi)$}    & \makecell{\textbf{Links on Target} \\ $f_{3.1b}(\psi)$}  & \makecell{\textbf{Length} \\ $f_{3.2}(\psi)$}   & \textbf{Num. of Collisions} \\ \hline
\multicolumn{1}{|c|}{\multirow{2}{*}{\rotatebox[origin=c]{90}{\textbf{Task 1\color[HTML]{FFFFFF}--}}}} & \makecell{$f_{1-2}(\psi)$\\\textbf{Bin Size}} & \makecell{$\oplus$ $F(2, 38) = 0.466$ \\ $p = 0.631$ \\ $\mu^2=0.024$} & \makecell{$\otimes$ $F(1, 31) = 7.898$ \\ \bm{$p < 0.01$} \\ $\mu^2=0.294$} & \makecell{$\otimes$ $F(2, 29) = 4.538$ \\ \bm{$p = 0.027$} \\ $\mu^2=0.193$} & \makecell{$F(2, 38) = 11.077$ \\ \bm{$p < 0.01$} \\ $\mu^2=0.368$} & \makecell{$\otimes$ $F(2, 38) = 1.249$ \\ $p = 0.298$ \\ $\mu^2=0.062$} & \makecell{$\oplus$ $F(2, 38) = 1.940$ \\ $p = 0.158$ \\ $\mu^2=0.093$} \\ \cline{2-8} 
\multicolumn{1}{|c|}{}                                 & \makecell{\textbf{Obstacle}\\\textbf{Avoidance}}   & \makecell{$\oplus$ $F(1, 19) = 5.472$ \\ $\bm{p = 0.030}$ \\ $\mu^2=0.224$} & \makecell{$\otimes$ $F(1, 19) = 4.501$ \\ \bm{$p = 0.047$} \\ $\mu^2=0.192$} & \makecell{$\otimes$ $F(1, 19) = 23.094$ \\ \bm{$p < 0.01$} \\ $\mu^2=0.549$} & \makecell{$F(1, 19) = 57.000$ \\ \bm{$p < 0.01$} \\ $\mu^2=0.750$} & \makecell{$\otimes$ $F(1, 19) = 4.181$ \\ $p = 0.055$ \\ $\mu^2=0.180$} & \makecell{$\oplus$ $F(1, 19) = 16239.458$ \\ \bm{$p < 0.01$} \\ $\mu^2=0.999$} \\ \hline \hline
\multicolumn{1}{|c|}{\multirow{2}{*}{\rotatebox[origin=c]{90}{\textbf{Task 2\color[HTML]{FFFFFF}--}}}} & \makecell{$f_{1-2}(\psi)$\\\textbf{Bin Size}} & \makecell{$F(2, 38) = 0.163$ \\ $p = 0.851$ \\ $\mu^2=0.008$} & \makecell{$\oplus$ $F(2, 38) = 16.464$ \\ \bm{$p < 0.01$} \\ $\mu^2=0.464$} & \makecell{$F(2, 38) = 6.784$ \\ \bm{$p < 0.01$} \\ $\mu^2=0.263$} & \makecell{$F(2, 30) = 1.150$ \\ $p = 0.327$ \\ $\mu^2=0.057$} & \makecell{$\otimes$ $F(2, 38) = 1.823$ \\ $p = 0.175$ \\ $\mu^2=0.088$} & \makecell{$F(2, 38) = 0.610$ \\ $p = 0.549$ \\ $\mu^2=0.031$} \\ \cline{2-8} 
\multicolumn{1}{|c|}{}                                 & \makecell{\textbf{Obstacle}\\\textbf{Avoidance}}   & \makecell{$F(1, 19) = 6.072$ \\ \bm{$p = 0.023$} \\ $\mu^2=0.242$} & \makecell{$\oplus$ $F(1, 19) = 0.297$ \\ $p = 0.592$ \\ $\mu^2=0.015$} & \makecell{$F(1, 19) = 3.213$ \\ $p = 0.089$ \\ $\mu^2=0.145$} & \makecell{$F(1, 19) = 3.470$ \\ $p = 0.078$ \\ $\mu^2=0.154$} & \makecell{$\otimes$ $F(1, 19) = 4.559$ \\ \bm{$p = 0.046$} \\ $\mu^2=0.194$} & \makecell{$F(1, 19) = 10459.665$ \\ \bm{$p < 0.01$} \\ $\mu^2=0.998$} \\ \hline \hline
\multicolumn{1}{|c|}{\multirow{2}{*}{\rotatebox[origin=c]{90}{\textbf{Task 3\color[HTML]{FFFFFF}--}}}} & \makecell{$f_{1-2}(\psi)$\\\textbf{Bin Size}} & \makecell{$\oplus$ $F(2, 38) = 12.331$ \\ \bm{$p < 0.01$} \\ $\mu^2=0.394$} & \makecell{$F(2, 38) = 6.895$ \\ \bm{$p < 0.01$} \\ $\mu^2=.0266$} & \makecell{$\otimes$ $F(2, 38) = 0.159$ \\ $p = 0.853$ \\ $\mu^2=0.008$} & \makecell{$\otimes$ $F(2, 38) = 11.300$ \\ \bm{$p < 0.01$} \\ $\mu^2=0.373$} & \makecell{$\otimes$ $F(2, 38) = 24.535$ \\ \bm{$p < 0.01$} \\ $\mu^2=0.564$} & \makecell{$F(2, 38) = 11.714$ \\ \bm{$p < 0.01$} \\ $\mu^2=0.381$} \\ \cline{2-8} 
\multicolumn{1}{|c|}{}                                 & \makecell{\textbf{Obstacle}\\\textbf{Avoidance}}   & \makecell{$\oplus$ $F(1, 19) = 8.114$ \\ $\bm{p = 0.010}$ \\ $\mu^2=0.299$} & \makecell{$F(1, 19) = 53.616$ \\ \bm{$p < 0.01$} \\ $\mu^2=0.738$} & \makecell{$\otimes$ $F(1, 19) = 0.055$ \\ $p = 0.055$ \\ $\mu^2=0.180$} & \makecell{$F(1, 19) = 13.108$ \\ \bm{$p < 0.01$} \\ $\mu^2=0.408$} & \makecell{$\otimes$ $F(1, 19) = 0.712$ \\ $p = 0.409$ \\ $\mu^2=0.036$} & \makecell{$F(1, 19) = 1118.194$ \\ \bm{$p < 0.01$} \\ $\mu^2=0.983$} \\ \hline \hline
\multicolumn{1}{|c|}{\multirow{2}{*}{\rotatebox[origin=c]{90}{\textbf{Task 4\color[HTML]{FFFFFF}--}}}} & \makecell{$f_{1-2}(\psi)$\\\textbf{Bin Size}} & \makecell{$\oplus$ $F(2, 38) = 14.339$ \\ \bm{$p < 0.01$} \\ $\mu^2=0.430$} & \makecell{$\oplus$ $F(2, 38) = 17.710$ \\ \bm{$p < 0.01$} \\ $\mu^2=0.482$} & \makecell{$\otimes$ $F(2, 38) = 9.648$ \\ \bm{$p < 0.01$} \\ $\mu^2=0.337$} & \makecell{$F(2, 38) = 1.430$ \\ $p = 0.252$ \\ $\mu^2=0.070$} & \makecell{$\otimes$ $F(2, 38) = 5.199$ \\ \bm{$p < 0.01$} \\ $\mu^2=0.215$} & \makecell{$F(2, 38) = 2.856$ \\ $p = 0.070$ \\ $\mu^2=0.131$} \\ \cline{2-8} 
\multicolumn{1}{|c|}{}                                 & \makecell{\textbf{Obstacle}\\\textbf{Avoidance}}   & \makecell{$\oplus$ $F(1, 19) = 4.805$ \\ $\bm{p = 0.41}$ \\ $\mu^2=0.202$} & \makecell{$\oplus$ $F(1, 19) = 22.317$ \\ \bm{$p < 0.01$} \\ $\mu^2=0.540$} & \makecell{$\otimes$ $F(1, 19) = 17.990$ \\ \bm{$p < 0.01$} \\ $\mu^2=0.486$} & \makecell{$F(1, 19) = 25.603$ \\ \bm{$p < 0.01$} \\ $\mu^2=0.574$} & \makecell{$\otimes$ $F(1, 19) = 36.292$ \\ \bm{$p < 0.01$} \\ $\mu^2=0.656$} & \makecell{$F(1, 19) = 3171.856$ \\ \bm{$p < 0.01$} \\ $\mu^2=0.994$} \\ \hline
\end{tabular}%
}
\end{table}

We ran a two-way ANOVA to determine the statistical difference between bin size of ranking partitioning and obstacle avoidance, whose results are in Tab.~\ref{tab:results_ANOVA}. We normalized data after a log transfer or ART, used the Green-House-Geisser for correcting the assumption on the sphericity condition, and conducted the pairwise comparisons with the LSD correction method to overcome the issue of protection from multiple univariate tests in case of a significant main effect of focused ANOVA. Significant values were set for $p < 0.05$.   

By comparing the results of the analysis with the data in the plots, we can make the following observations over every task:
\begin{itemize}
    \item the error in solving the inverse kinematics ($f_{1-2}$) decreases significantly when the obstacle avoidance is active;
    \item the number of links decreases significantly with a higher bin size (especially for $f_{3.1a}$) and partially when the obstacle avoidance is active;
    \item undulation ($f_{3.3}$) and overall length ($f_{3.2}$) are mostly not significantly affected by any of the factors, as expected, being objectives with low priority; and
    \item the number of collisions with obstacles during execution significantly decreases with the obstacle avoidance algorithm, as expected.
\end{itemize}
Based on these observations, the bin size can be set to a (reasonable) high value to leave room for the optimization of the other objectives; this will still provide good results in terms of inverse kinematics because the rank partitioning will eventually sort the population ensuring the best solutions to be considered as elites. 
The obstacle-avoidance method shows good results by organically generating feasible solutions that do not collide with obstacles, reducing this type of infeasible solutions only to the case depicted in Fig~\ref{fig:wrongGeneration}. As shown by the analysis, this feature also allows for more precision in solving the inverse kinematics and in reducing the number of links of the robot; however, as the computational complexity of the algorithm increases, the running time is consequently slowed down.

Fig.~\ref{fig:optimal_solutions} shows the optimal solutions retrieved during the experiments for each task; these were selected from the pool of all solutions in the experiment ($20 \cdot 6 = 120$ for each task) with the rank-partitioning algorithm, which yield the same result for every bin size. 
These optimal solutions were retrieved with the condition bin size $ = 1.0$ and \textit{active} obstacle avoidance, which complies with the results of the ANOVA. Note that the robots in Fig.~\ref{fig:best_sol_t2},~\ref{fig:best_sol_t3} and~\ref{fig:best_sol_t4} are not colliding nor touching any obstacle, but since they get very close to it, it is preferable to set the radius for the obstacles to be bigger than what they actually are.
Finally, we report that all solutions retrieved by the optimizer during the experiments were feasible.




\subsection{Comparison with previous work}
\label{sec:comparing_methods}

\begin{figure}[t!]
    \centering
    \subfigure[\protect\url{}\label{fig:best_method_prev}Previous]%
    {\includegraphics[height=5.5cm]{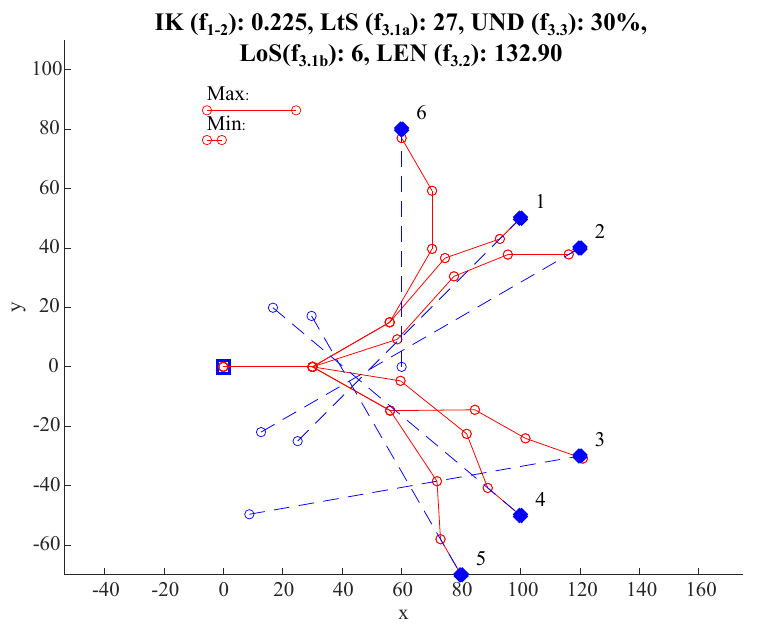}}
    \subfigure[\protect\url{}\label{fig:best_method_prop}Proposed]%
    {\includegraphics[height=5.5cm]{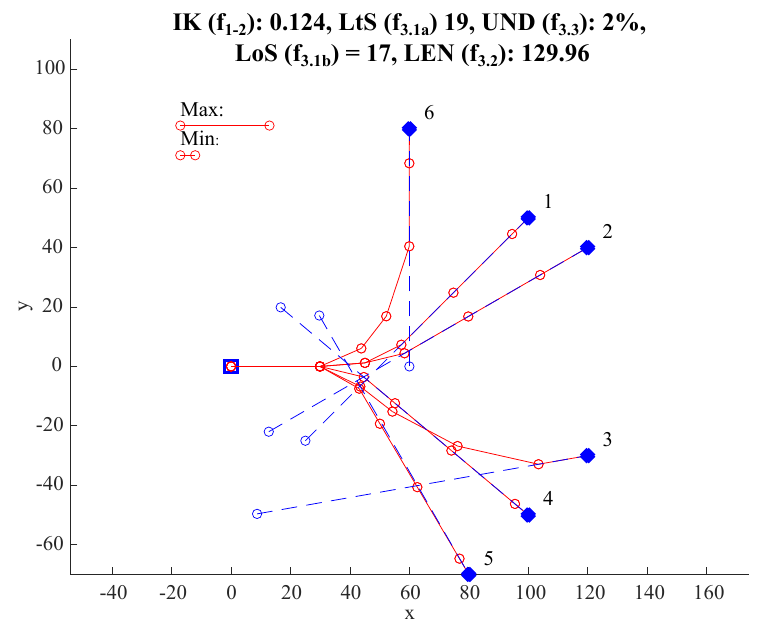}}
    \caption{Comparison of solutions retrieved by: (a) the previous method~\cite{exarchos2022task}, and (b) the one proposed in this work. The solution in (b) outperforms the solution in (a) under all objectives -- although the overall number of links in (b) is higher than in (a), the configurations in (b) immediately aim for the targets without rambling in the environment. Furthermore, the solution in (a) is unfeasible as its configuration 3 does not reach the target with the right orientation, and configurations 2 and 6 have a last-link length smaller than its lower bound $\Delta^{(L)}_{\theta}$.}
    \label{fig:comparing_methods}
\end{figure}

\begin{table}[]
\caption{Comparative results}
\label{tab:comparison}
\resizebox{\textwidth}{!}{%
\begin{tabular}{l|c|c|c|c|c|c|}


\cline{2-7}
                            & \textbf{IK ($f_{1-2}$)}           & \textbf{LtS ($f_{3.1a}$)}            & \textbf{UND ($f_{3.3}$) [$\%$]}            & \textbf{LoS ($f_{3.1b}$)}            & \textbf{LEN ($f_{3.2}$)}            & \textbf{Runtime {[}s{]}} \\ \hline
\multicolumn{1}{|l|}{Previous} & $0.42\pm0.19$ & $31.50\pm4.50$ & $29.00\pm1.00$ & $7.50\pm1.50$  & $132.39\pm0.51$ & $98.48\pm2.98$ \\
\multicolumn{1}{|l|}{Proposed} & $0.36\pm0.28$ & $21.20\pm1.63$ & $4.45\pm3.44$  & $16.55\pm3.41$ & $130.51\pm1.09$ & $84.92\pm2.80$   \\ \hline
\end{tabular}%
}
\end{table}

The prior work on soft-growing-robot optimization design~\cite{exarchos2022task} presented a preliminary preference-based strategy with a different formulation for the objective functions. We ran a brief experiment to compare the previous method to the one presented in this paper. Since we were mostly interested in evaluating precision, we used a task with many targets and no obstacles. Both algorithms were executed twenty times on the same task. 
The specific parameters for each optimization method, Cross-Entropy for the previous and Genetic Algorithm for the proposed, are reported in the prior work~\cite{exarchos2022task} and in Sec.~\ref{sec:parameters}, respectively; whereas similar parameters to both methods were fixed to the same values:
$500$ population size and $150$ generations. 

Numerical results are summarized in Tab.~\ref{tab:comparison}, whereas Fig.~\ref{fig:comparing_methods} shows the best solution retrieved by each method, (a) for the previous and (b) for the proposed. The values of minimized objectives are reported in the figures for each method, showing smaller values in almost every metric for the proposed method. In fact, the configurations retrieved by the proposed method are shorter, more compact, and more precise in solving the inverse kinematics; whereas the previous method retrieved longer, wavy, and more dispersed configurations. The proposed method does, however, employ more links overall, but this is to the benefit of configuration distribution.
Furthermore, based on the formulation described in Sec.~\ref{sec:optimization_problem}, the solution retrieved by the previous method is unfeasible -- with the configuration reaching for target 3 not being aligned with the target's orientation segment and even overshooting it, and configurations reaching for targets 2 and 6 having the length of their last link smaller than the specified lower bound $\Delta^{(L)}_{\theta}$.

Quantitatively, the proposed method features a $14\%$ higher precision in solving the inverse kinematics with respect to the previous method, with configurations that are $2\%$ shorter, employing $33\%$ fewer nodes to reach the right orientation, and $85\%$ less undulated -- the latter being particularly important because the robot will require less actuation to reach the targets. The proposed method was also, on average, $13\%$ faster than the previous one -- for a fair comparison, the experiments were executed on the same machine with equal population size and number of generations. In light of these results, we can report that the proposed method based on the novel fitness formulation and rank partitioning outperforms the previous one, resulting in robots that are easier and cheaper to build due to their shorter design.  


\subsection{Considerations on problem dimensionality}
\label{sec:dimensionality}

The tasks proposed in Sec.~\ref{sec:exp_optimality} and~\ref{sec:comparing_methods} have high variable dimensionality, which can be evaluated by considering the fixed value of the maximum link number $n=20$ and the number of targets $|t|$ in the task. Each link accounts for a decision variable per joint angle, multiplied for each target to be reached (i.e., one robot configuration for each target in the task); whereas the link lengths are shared among all configurations, hence adding a constant term. This results in a dimensionality of $n \cdot |t| + n$. We can then evaluate the following dimensionalities:
\begin{itemize}
    \item Task 1 (Sec.~\ref{sec:exp_optimality}): $20 \cdot 4 + 20 = 100$ decision variables;
    \item Task 2 (Sec.~\ref{sec:exp_optimality}): $20 \cdot 3 + 20 = 80$ decision variables;
    \item Task 3 (Sec.~\ref{sec:exp_optimality}): $20 \cdot 1 + 20 = 40$ decision variables;
    \item Task 4 (Sec.~\ref{sec:exp_optimality}): $20 \cdot 1 + 20 = 40$ decision variables; and
    \item Comparative Task (Sec.~\ref{sec:comparing_methods}): $20 \cdot 6 + 20 = 140$ decision variables.
\end{itemize}
	
As shown by the results reported in Sec.~\ref{sec:exp_optimality} and~\ref{sec:comparing_methods}, the proposed method works well with different (high) dimensionalities. It is reasonable to claim that the limits depend on the computational cost of solving a specific task.

\section{Conclusion}
\label{sec:conclusion}
In this work, we presented an innovative and efficient tool to optimize the design of planar soft-growing robots for solving specific tasks. We introduced an innovative method to solve a multi-objective optimization problem as a single-objective, namely the \textit{rank-partitioning} algorithm, which exploits the population-based methodology of evolutionary computation techniques. Furthermore, we proposed an obstacle-avoidance algorithm that can be integrated directly into the generation of solutions (either randomly or by variation) to reduce the search space of the problem. 
Results show that the proposed method features an effective formulation for the optimization problem, retrieves solutions with higher precision compared to the previous method~\cite{exarchos2022task}, has significant effects on the performance,  and implements an obstacle-avoidance algorithm that improves both convergence and quality of the optimizer.

In the future, we will work on other robust and admissible heuristics to include in the objective functions to speed up convergence, as well as different methodologies for optimization. When approaching three-dimensional problems, the computational toll of evolutionary algorithms might pose a challenge for more complex problems; therefore, since the rank partitioning method can be employed in any population-based method, we are planning to investigate the performance of alternative and faster optimizers such as Big Bang-Big Crunch \cite{erol2006new} and compare them with the current work.  
Furthermore, more properties can be taken into account while defining the problem, such as the order of targets or the characterization of the joint limits in terms of steering ($\Delta_\theta$): the shorter the link (either due to the proximity to a joint or the tip), the harder it is to achieve larger angles; whereas the longer the link, the easier. Tests on real robots will be needed to properly set the limits on steering to quantify the relationship between the distance of a joint from the tip and the force required to steer for a given desired angle. More constraints can be defined by engineers as they investigate innovative technologies for the manufacturing and actuation of soft-growing robots.



\section*{Acknowledgments}

This work is funded by TUBİTAK within the scope of the 2232-B International Fellowship for Early Stage Researchers Program number 121C145.

I initiated this work during my post-doctorate activities at Stanford University; therefore, I would like to thank his advisor Prof. Allison M. Okamura for the support shown during that period. Furthermore, would I like to thank my undergraduate students who are currently working on the topics mentioned in Sec.~\ref{sec:conclusion}: Kuzey Arar, Reha Oguz Sayin, Kadir Kaan Atalay, and Kemal Erdem Yenin.

Readers interested in manufacturing these soft-growing robots can find more information at \url{https://www.vinerobots.org/}.




 \bibliographystyle{elsarticle-num} 
 \bibliography{optimizer}

\begin{thebibliography}{10}
\expandafter\ifx\csname url\endcsname\relax
  \def\url#1{\texttt{#1}}\fi
\expandafter\ifx\csname urlprefix\endcsname\relax\def\urlprefix{URL }\fi
\expandafter\ifx\csname href\endcsname\relax
  \def\href#1#2{#2} \def\path#1{#1}\fi

\bibitem{webster2010design}
R.~J. Webster~III, B.~A. Jones, Design and kinematic modeling of constant
  curvature continuum robots: A review, SAGE International Journal of Robotics
  Research 29~(13) (2010) 1661--1683.

\bibitem{rus2015design}
D.~Rus, M.~T. Tolley, Design, fabrication and control of soft robots, Nature
  521~(7553) (2015) 467--475.

\bibitem{whitesides2018soft}
G.~M. Whitesides, Soft robotics, Wiley Angewandte Chemie International Edition
  57~(16) (2018) 4258--4273.

\bibitem{hawkes2021hard}
E.~W. Hawkes, C.~Majidi, M.~T. Tolley, Hard questions for soft robotics,
  Science Robotics 6~(53) (2021) eabg6049.

\bibitem{do2020dynamically}
B.~H. Do, V.~Banashek, A.~M. Okamura, Dynamically reconfigurable discrete
  distributed stiffness for inflated beam robots, in: International Conference
  on Robotics and Automation, IEEE, 2020, pp. 9050--9056.

\bibitem{coad2019vine}
M.~M. Coad, L.~H. Blumenschein, S.~Cutler, J.~A.~R. Zepeda, N.~D. Naclerio,
  H.~El-Hussieny, U.~Mehmood, J.-H. Ryu, E.~W. Hawkes, A.~M. Okamura, Vine
  robots: Design, teleoperation, and deployment for navigation and exploration,
  IEEE Robotics and Automation Magazine 27~(3) (2019) 120--132.

\bibitem{hawkes2017soft}
E.~W. Hawkes, L.~H. Blumenschein, J.~D. Greer, A.~M. Okamura, A soft robot that
  navigates its environment through growth, Science Robotics 2~(8) (2017)
  eaan3028.

\bibitem{blumenschein2017modeling}
L.~H. Blumenschein, A.~M. Okamura, E.~W. Hawkes, Modeling of bioinspired apical
  extension in a soft robot, in: Conference on Biomimetic and Biohybrid
  Systems, Springer, 2017, pp. 522--531.

\bibitem{stroppa2020human}
F.~Stroppa, M.~Luo, K.~Yoshida, M.~M. Coad, L.~H. Blumenschein, A.~M. Okamura,
  Human interface for teleoperated object manipulation with a soft growing
  robot, in: International Conference on Robotics and Automation, IEEE, 2020,
  pp. 726--732.

\bibitem{schulz2017interactive}
A.~Schulz, C.~Sung, A.~Spielberg, W.~Zhao, R.~Cheng, E.~Grinspun, D.~Rus,
  W.~Matusik, Interactive robogami: An end-to-end system for design of robots
  with ground locomotion, SAGE International Journal of Robotics Research
  36~(10) (2017) 1131--1147.

\bibitem{morimoto2018toward}
T.~K. Morimoto, J.~D. Greer, E.~W. Hawkes, M.~H. Hsieh, A.~M. Okamura, Toward
  the design of personalized continuum surgical robots, Springer Annals of
  Biomedical Engineering 46~(10) (2018) 1522--1533.

\bibitem{exarchos2022task}
I.~Exarchos, K.~Wang, B.~Do, F.~Stroppa, M.~Coad, A.~Okamura, K.~Liu,
  Task-specific design optimization and fabrication for inflated-beam soft
  robots with growable discrete joints, in: International Conference on
  Robotics and Automation, IEEE, 2022, pp. 7145--7151.

\bibitem{hiller2011automatic}
J.~Hiller, H.~Lipson, Automatic design and manufacture of soft robots, IEEE
  Transactions on Robotics 28~(2) (2011) 457--466.

\bibitem{marchese2016dynamics}
A.~Marchese, R.~Tedrake, D.~Rus, Dynamics and trajectory optimization for a
  soft spatial fluidic elastomer manipulator, SAGE International Journal of
  Robotics Research 35~(8) (2016) 1000--1019.

\bibitem{wang2020topology}
R.~Wang, X.~Zhang, B.~Zhu, H.~Zhang, B.~Chen, H.~Wang, Topology optimization of
  a cable-driven soft robotic gripper, Springer Structural and
  Multidisciplinary Optimization 62~(5) (2020) 2749--2763.

\bibitem{cheong2021optimal}
H.~Cheong, M.~Ebrahimi, T.~Duggan, Optimal design of continuum robots with
  reachability constraints, IEEE Robotics and Automation Letters 6~(2) (2021)
  3902--3909.

\bibitem{chen2021enhancing}
S.~Chen, Y.~Wang, D.~Li, F.~Chen, X.~Zhu, Enhancing interaction performance of
  soft pneumatic-networks grippers by skeleton topology optimization, Springer
  Science China Technological Sciences 64~(12) (2021) 2709--2717.

\bibitem{ghoreishi2021bayesian}
S.~Ghoreishi, R.~Sochol, D.~Gandhi, A.~Krieger, M.~Fuge, Bayesian optimization
  for design of multi-actuator soft catheter robots, IEEE Transactions on
  Medical Robotics and Bionics 3~(3) (2021) 725--737.

\bibitem{bern2021soft}
J.~Bern, D.~Rus, Soft ik with stiffness control, in: 2021 IEEE 4th
  International Conference on Soft Robotics, IEEE, 2021, pp. 465--471.

\bibitem{tang2021design}
C.~Tang, H.~Huang, B.~Li, Design and control of a magnetic driven worm-like
  micro-robot, in: International Conference on Robotics and Biomimetics, IEEE,
  2021, pp. 1304--1308.

\bibitem{koehler2020model}
M.~Koehler, N.~Usevitch, A.~Okamura, Model-based design of a soft 3-d haptic
  shape display, IEEE Transactions on Robotics 36~(3) (2020) 613--628.

\bibitem{ma2017computational}
L.-K. Ma, Y.~Zhang, Y.~Liu, K.~Zhou, X.~Tong, Computational design and
  fabrication of soft pneumatic objects with desired deformations, ACM
  Transactions on Graphics 36~(6) (2017) 1--12.

\bibitem{fang2020kinematics}
G.~Fang, C.-D. Matte, R.~Scharff, T.-H. Kwok, C.~Wang, Kinematics of soft
  robots by geometric computing, IEEE Transactions on Robotics 36~(4) (2020)
  1272--1286.

\bibitem{camarillo2008mechanics}
D.~Camarillo, C.~Milne, C.~Carlson, M.~Zinn, K.~Salisbury, Mechanics modeling
  of tendon-driven continuum manipulators, IEEE Transactions on Robotics 24~(6)
  (2008) 1262--1273.

\bibitem{rucker2011statics}
C.~Rucker, R.~III, Statics and dynamics of continuum robots with general tendon
  routing and external loading, IEEE Transactions on Robotics 27~(6) (2011)
  1033--1044.

\bibitem{burgner2013computational}
J.~Burgner, H.~Gilbert, R.~Webster, On the computational design of concentric
  tube robots: Incorporating volume-based objectives, in: International
  Conference on Robotics and Automation, IEEE, 2013, pp. 1193--1198.

\bibitem{bergeles2015concentric}
C.~Bergeles, A.~Gosline, N.~Vasilyev, P.~Codd, J.~Pedro, P.~Dupont, Concentric
  tube robot design and optimization based on task and anatomical constraints,
  IEEE Transactions on Robotics 31~(1) (2015) 67--84.

\bibitem{lloyd2020optimal}
P.~Lloyd, G.~Pittiglio, J.~Chandler, P.~Valdastri, Optimal design of soft
  continuum magnetic robots under follow-the-leader shape forming actuation,
  in: International Symposium on Medical Robotics, IEEE, 2020, pp. 111--117.

\bibitem{doroudchi2021configuration}
A.~Doroudchi, S.~Berman, Configuration tracking for soft continuum robotic arms
  using inverse dynamic control of a cosserat rod model, in: International
  Conference on Soft Robotics, IEEE, 2021, pp. 207--214.

\bibitem{rosi2022sensing}
E.~Rosi, M.~Stolzle, F.~Solari, C.~Santina, Sensing soft robots' shape with
  cameras: an investigation on kinematics-aware slam, in: International
  Conference on Soft Robotics, IEEE, 2022, pp. 795--801.

\bibitem{coevoet2017optimization}
E.~Coevoet, A.~Escande, C.~Duriez, Optimization-based inverse model of soft
  robots with contact handling, IEEE Robotics and Automation Letters 2~(3)
  (2017) 1413--1419.

\bibitem{adagolodjo2021coupling}
Y.~Adagolodjo, F.~Renda, C.~Duriez, Coupling numerical deformable models in
  global and reduced coordinates for the simulation of the direct and the
  inverse kinematics of soft robots, IEEE Robotics and Automation Letters 6~(2)
  (2021) 3910--3917.

\bibitem{anor2011algorithms}
T.~Anor, J.~Madsen, P.~Dupont, Algorithms for design of continuum robots using
  the concentric tubes approach: A neurosurgical example, in: International
  Conference on Robotics and Automation, IEEE, 2011, pp. 667--673.

\bibitem{de2020topology}
E.~Souza, E.~Silva, Topology optimization applied to the design of actuators
  driven by pressure loads, Springer Structural and Multidisciplinary
  Optimization 61~(5) (2020) 1763--1786.

\bibitem{coello1998using}
C.~A.~C. Coello, A.~D. Christiansen, A.~H. Aguirre, Using a new ga-based
  multiobjective optimization technique for the design of robot arms, Cambridge
  University Press Robotica 16~(4) (1998) 401--414.

\bibitem{iwasaki2000evolutionary}
M.~Iwasaki, N.~Matsui, Evolutionary identification algorithm for unknown
  structured mechatronics systems using ga, in: 2000 26th Annual Conference of
  the IEEE Industrial Electronics Society. IECON 2000. 2000 International
  Conference on Industrial Electronics, Control and Instrumentation. 21st
  Century Technologies, Vol.~4, IEEE, 2000, pp. 2492--2496.

\bibitem{lim2005inverse}
D.~Lim, Y.-S. Ong, B.-S. Lee, Inverse multi-objective robust evolutionary
  design optimization in the presence of uncertainty, in: 7th annual workshop
  on genetic and evolutionary computation, 2005, pp. 55--62.

\bibitem{clark2013evolutionary}
A.~Clark, P.~McKinley, Evolutionary optimization of robotic fish control and
  morphology, in: Conference on Genetic and Evolutionary Computation, 2013, pp.
  21--22.

\bibitem{gao2014performance}
Z.~Gao, D.~Zhang, Performance analysis, mapping, and multiobjective
  optimization of a hybrid robotic machine tool, IEEE Transactions on
  Industrial Electronics 62~(1) (2014) 423--433.

\bibitem{wang2015multi}
C.~Wang, Y.~Fang, S.~Guo, Multi-objective optimization of a parallel ankle
  rehabilitation robot using modified differential evolution algorithm,
  Springer Chinese Journal of Mechanical Engineering 28~(4) (2015) 702--715.

\bibitem{filipiak2015infeasibility}
P.~Filipiak, K.~Michalak, P.~Lipinski, Infeasibility driven evolutionary
  algorithm with the anticipation mechanism for the reaching goal in dynamic
  constrained inverse kinematics, in: Companion Publication of the 2015 Annual
  Conference on Genetic and Evolutionary Computation, 2015, pp. 1389--1390.

\bibitem{rokbani2022beta}
N.~Rokbani, S.~Mirjalili, M.~Slim, A.~Alimi, A beta salp swarm algorithm
  meta-heuristic for inverse kinematics and optimization, Springer Applied
  Intelligence (2022) 1--26.

\bibitem{stroppa2023optimizing}
F.~Stroppa, A.~Soylemez, H.~T. Yuksel, B.~Akbas, M.~Sarac, Optimizing
  exoskeleton design with evolutionary computation: an intensive survey, MDPI
  Robotics 12~(4) (2023) 106.

\bibitem{deb1999multi}
K.~Deb, Multi-objective genetic algorithms: Problem difficulties and
  construction of test problems, MIT Press Evolutionary Computation 7~(3)
  (1999) 205--230.

\bibitem{bedell2011design}
C.~Bedell, J.~Lock, A.~Gosline, P.~Dupont, Design optimization of concentric
  tube robots based on task and anatomical constraints, in: 2011 International
  conference on robotics and automation, Ieee, 2011, pp. 398--403.

\bibitem{berger2015growing}
B.~Berger, A.~Andino, A.~Danise, J.~Rieffel, Growing and evolving vibrationally
  actuated soft robots, in: Conference on Genetic and Evolutionary Computation,
  2015, pp. 1221--1224.

\bibitem{cheney2015evolving}
N.~Cheney, J.~Bongard, H.~Lipson, Evolving soft robots in tight spaces, in:
  Conference on Genetic and Evolutionary Computation, 2015, pp. 935--942.

\bibitem{chen2019optimal}
Y.~Chen, Z.~Xia, Q.~Zhao, Optimal design of soft pneumatic bending actuators
  subjected to design-dependent pressure loads, IEEE Transactions on
  Mechatronics 24~(6) (2019) 2873--2884.

\bibitem{medvet2021biodiversity}
E.~Medvet, A.~Bartoli, F.~Pigozzi, M.~Rochelli, Biodiversity in evolved
  voxel-based soft robots, in: Genetic and Evolutionary Computation Conference,
  2021, pp. 129--137.

\bibitem{luo2015optimized}
M.~Luo, E.~Skorina, W.~Tao, F.~Chen, C.~Onal, Optimized design of a rigid
  kinematic module for antagonistic soft actuation, in: 2015 International
  Conference on Technologies for Practical Robot Applications (TePRA), IEEE,
  2015, pp. 1--6.

\bibitem{bodily2017multi}
D.~M. Bodily, T.~F. Allen, M.~D. Killpack, Multi-objective design optimization
  of a soft, pneumatic robot, in: International Conference on Robotics and
  Automation, IEEE, 2017, pp. 1864--1871.

\bibitem{tan2017simultaneous}
N.~Tan, X.~Gu, H.~Ren, Simultaneous robot-world, sensor-tip, and kinematics
  calibration of an underactuated robotic hand with soft fingers, IEEE Access 6
  (2017) 22705--22715.

\bibitem{liu2022simulation}
J.~Liu, J.~H. Low, Q.~Han, M.~Lim, D.~Lu, C.-H. Yeow, Z.~Liu, Simulation data
  driven design optimization for reconfigurable soft gripper system, IEEE
  Robotics and Automation Letters 7~(2) (2022) 5803--5810.

\bibitem{fitzgerald2021evolving}
S.~Fitzgerald, G.~Delaney, D.~Howard, F.~Maire, Evolving soft robotic jamming
  grippers, in: Genetic and Evolutionary Computation Conference, 2021, pp.
  102--110.

\bibitem{rieffel2009evolving}
J.~Rieffel, F.~Saunders, S.~Nadimpalli, H.~Zhou, S.~Hassoun, J.~Rife,
  B.~Trimmer, Evolving soft robotic locomotion in physx, in: 11th annual
  conference companion on genetic and evolutionary computation conference: Late
  breaking papers, 2009, pp. 2499--2504.

\bibitem{tesch2013expensive}
M.~Tesch, J.~Schneider, H.~Choset, Expensive multiobjective optimization for
  robotics, in: International Conference on Robotics and Automation, IEEE,
  2013, pp. 973--980.

\bibitem{methenitis2015novelty}
G.~Methenitis, D.~Hennes, D.~Izzo, A.~Visser, Novelty search for soft robotic
  space exploration, in: Conference on Genetic and Evolutionary Computation,
  2015, pp. 193--200.

\bibitem{kriegman2017minimal}
S.~Kriegman, N.~Cheney, F.~Corucci, J.~Bongard, A minimal developmental model
  can increase evolvability in soft robots, in: Genetic and Evolutionary
  Computation Conference, 2017, pp. 131--138.

\bibitem{sui2022task}
X.~Sui, T.~Zheng, J.~Qi, Z.~Yang, N.~Zhao, J.~Zhao, H.~Cai, Y.~Zhu,
  Task-oriented hierarchical control of modular soft robots with external
  vision guidance, Springer Journal of Bionic Engineering 19~(3) (2022)
  657--667.

\bibitem{marzougui2022comparative}
D.~Marzougui, M.~Biondina, F.~wyffels, A comparative analysis on genome
  pleiotropy for evolved soft robots, in: Genetic and Evolutionary Computation
  Conference Companion, 2022, pp. 136--139.

\bibitem{runge2017design}
G.~Runge, J.~Peters, A.~Raatz, Design optimization of soft pneumatic actuators
  using genetic algorithms, in: International Conference on Robotics and
  Biomimetics, IEEE, 2017, pp. 393--400.

\bibitem{ferigo2022optimizing}
A.~Ferigo, E.~Medvet, G.~Iacca, Optimizing the sensory apparatus of voxel-based
  soft robots through evolution and babbling, Springer Computer Science 3~(2)
  (2022) 1--17.

\bibitem{nguyen2007genetic}
V.~Nguyen, A.~Morris, Genetic algorithm tuned fuzzy logic controller for a
  robot arm with two-link flexibility and two-joint elasticity, Springer
  Journal of Intelligent and Robotic Systems 49~(1) (2007) 3--18.

\bibitem{pigozzi2022evolving}
F.~Pigozzi, Y.~Tang, E.~Medvet, D.~Ha, Evolving modular soft robots without
  explicit inter-module communication using local self-attention, arXiv
  preprint arXiv:2204.06481 (2022).

\bibitem{ghoul2022optimized}
A.~Ghoul, K.~Kara, M.~Benrabah, M.~Hadjili, Optimized nonlinear sliding mode
  control of a continuum robot manipulator, Springer Journal of Control,
  Automation and Electrical Systems (2022) 1--9.

\bibitem{chen2020obstacle}
L.~Chen, Y.~Ma, Y.~Zhang, J.~Liu, Obstacle avoidance and multitarget tracking
  of a super redundant modular manipulator based on bezier curve and particle
  swarm optimization, Springer Chinese Journal of Mechanical Engineering 33~(1)
  (2020) 1--19.

\bibitem{dakev1996evolutionary}
N.~Dakev, A.~Chipperfield, P.~Fleming, An evolutionary approach for path
  following optimal control of multibody systems, in: Proceedings of
  International Conference on Evolutionary Computation, IEEE, 1996, pp.
  512--516.

\bibitem{castillo2007multiple}
O.~Castillo, L.~Trujillo, P.~Melin, Multiple objective genetic algorithms for
  path-planning optimization in autonomous mobile robots, Springer Soft
  Computing 11~(3) (2007) 269--279.

\bibitem{chae2009trajectory}
K.~G. Chae, J.~H. Park, Trajectory optimization with ga and control for
  quadruped robots, Springer Journal of Mechanical Science and Technology
  23~(1) (2009) 114--123.

\bibitem{ayari2017new}
A.~Ayari, S.~Bouamama, A new multiple robot path planning algorithm: dynamic
  distributed particle swarm optimization, Springer Robotics and Biomimetics
  4~(1) (2017) 1--15.

\bibitem{miettinen2012nonlinear}
K.~Miettinen, Nonlinear multiobjective optimization, Vol.~12, Springer, 2012.

\bibitem{hwang2012multiple}
C.-L. Hwang, A.~S.~M. Masud, Multiple objective decision making—methods and
  applications: a state-of-the-art survey, Vol. 164, Springer, 2012.

\bibitem{goldberg1990real}
D.~E. Goldberg, Real-Coded Genetic Algorithms, Virtual Alphabets and Blocking,
  Vol.~5, 1991.

\bibitem{tsukagoshi2011tip}
H.~Tsukagoshi, N.~Arai, I.~Kiryu, A.~Kitagawa, Tip growing actuator with the
  hose-like structure aiming for inspection on narrow terrain., International
  Journal of Automation Technology 5~(4) (2011) 516--522.

\bibitem{sadeghi2017toward}
A.~Sadeghi, A.~Mondini, B.~Mazzolai, Toward self-growing soft robots inspired
  by plant roots and based on additive manufacturing technologies, Mary Ann
  Liebert, Inc. Soft Robotics 4~(3) (2017) 211--223.

\bibitem{jeong2020tip}
S.-G. Jeong, M.~M. Coad, L.~H. Blumenschein, M.~Luo, U.~Mehmood, J.~H. Kim,
  A.~M. Okamura, J.-H. Ryu, A tip mount for transporting sensors and tools
  using soft growing robots, in: International Conference on Intelligent Robots
  and Systems, 2020, pp. 8781--8788.

\bibitem{blumenschein2018tip}
L.~H. Blumenschein, L.~T. Gan, J.~A. Fan, A.~M. Okamura, E.~W. Hawkes, A
  tip-extending soft robot enables reconfigurable and deployable antennas, IEEE
  Robotics and Automation Letters 3~(2) (2018) 949--956.

\bibitem{kim2018origami}
S.-J. Kim, D.-Y. Lee, G.-P. Jung, K.-J. Cho, An origami-inspired, self-locking
  robotic arm that can be folded flat, Science Robotics 3~(16) (2018).

\bibitem{mishima2003development}
D.~Mishima, T.~Aoki, S.~Hirose, Development of pneumatically controlled
  expandable arm for search in the environment with tight access, in: Field and
  Service Robotics, Springer, 2003, pp. 509--518.

\bibitem{blumenschein2018helical}
L.~H. Blumenschein, N.~S. Usevitch, B.~H. Do, E.~W. Hawkes, A.~M. Okamura,
  Helical actuation on a soft inflated robot body, in: International Conference
  on Soft Robotics, 2018, pp. 245--252.

\bibitem{gan20203d}
L.~T. Gan, L.~H. Blumenschein, Z.~Huang, A.~M. Okamura, E.~W. Hawkes, J.~A.
  Fan, 3d electromagnetic reconfiguration enabled by soft continuum robots,
  IEEE Robotics and Automation Letters 5~(2) (2020) 1704--1711.

\bibitem{coad2020retraction}
M.~M. Coad, R.~P. Thomasson, L.~H. Blumenschein, N.~S. Usevitch, E.~W. Hawkes,
  A.~M. Okamura, Retraction of soft growing robots without buckling, IEEE
  Robotics and Automation Letters 5~(2) (2020) 2115--2122.

\bibitem{el2018development}
H.~El-Hussieny, U.~Mehmood, Z.~Mehdi, S.-G. Jeong, M.~Usman, E.~W. Hawkes,
  A.~M. Okamura, J.-H. Ryu, Development and evaluation of an intuitive flexible
  interface for teleoperating soft growing robots, in: International Conference
  on Intelligent Robots and Systems, 2018, pp. 4995--5002.

\bibitem{coello2002theoretical}
C.~A.~C. Coello, Theoretical and numerical constraint-handling techniques used
  with evolutionary algorithms: a survey of the state of the art, Elsevier
  Computer Methods in Applied Mechanics and Engineering 191~(11-12) (2002)
  1245--1287.

\bibitem{cohon1985multicriteria}
J.~L. Cohon, Multicriteria programming: Brief review and application, Elsevier
  Design Optimization (1985) 163--191.

\bibitem{stroppa2018convex}
F.~Stroppa, C.~Loconsole, A.~Frisoli, Convex polygon fitting in robot-based
  neurorehabilitation, Elsevier Applied Soft Computing 68 (2018) 609--625.

\bibitem{dong2015obstacle}
H.~Dong, Z.~Du, Obstacle avoidance path planning of planar redundant
  manipulators using workspace density, SAGE International Journal of Advanced
  Robotic Systems 12~(2) (2015) 9.

\bibitem{cheng1989historical}
H.~Cheng, K.~Gupta, An historical note on finite rotations, ASME Journal of
  Applied Mechanics 56 (1989) 139--145.

\bibitem{eshelman1993real}
L.~J. Eshelman, J.~D. Schaffer, Real-coded genetic algorithms and
  interval-schemata, in: Elsevier Foundations of Genetic Algorithms, Vol.~2,
  Elsevier, 1993, pp. 187--202.

\bibitem{michalewicz1992genetic}
Z.~Michalewicz, Genetic algorithms+ data structures= evolutionary programs,
  spingerverlag (1992).

\bibitem{goldberg1991comparative}
D.~E. Goldberg, K.~Deb, A comparative analysis of selection schemes used in
  genetic algorithms, in: Elsevier Foundations of Genetic Algorithms, Vol.~1,
  Elsevier, 1991, pp. 69--93.

\bibitem{beyer2002evolution}
H.-G. Beyer, H.-P. Schwefel, Evolution strategies--a comprehensive
  introduction, Springer Natural Computing 1 (2002) 3--52.

\bibitem{jh1975adaptation}
H.~Jh, Adaptation in natural and artificial systems, The University of Michigan
  Press Ann Arbor (1975).

\bibitem{erol2006new}
O.~K. Erol, I.~Eksin, A new optimization method: Big {B}ang--{B}ig {C}runch,
  Elsevier Advances in Engineering Software 37~(2) (2006) 106--111.

\end{thebibliography}

\end{document}